\pdfoutput=1
\documentclass[a4paper, 11pt]{article}

\usepackage{naacl2021}
\usepackage{graphicx}
\graphicspath{ {./images/} }
\usepackage{subfig}
\usepackage{hyperref}

\usepackage{algorithm}
\usepackage{algpseudocode}
\usepackage{pifont}

\usepackage{times}
\usepackage{latexsym}
\usepackage{amsmath}
\usepackage{bm}
\usepackage[T1]{fontenc}
\usepackage[utf8]{inputenc}
\usepackage{microtype}
\usepackage{amsfonts}
\usepackage{xcolor}
\title{Transformer visualization via dictionary learning:\\
contextualized embedding as a linear superposition of transformer factors}

\author{
\setlength{\tabcolsep}{10pt}
\begin{tabular}{@{}cccc@{}}
Zeyu Yun\thanks{\ equal contribution. Correspondence to: Zeyu Yun $<$chobitstian@berkeley.edu$>$, Yubei Chen $<$yubeic@\{fb.com, berkeley.edu\}$>$} $^{\ 2}$ & 
Yubei Chen$^{*\ 1,2}$ &
Bruno A Olshausen$^{2,4}$ & 
Yann LeCun$^{1,3}$\\
\end{tabular}\\[5pt]
\small\ $^{1}$ Facebook AI Research\\
\small\ $^{2}$ Berkeley AI Research (BAIR), UC Berkeley\\
\small\ $^{3}$ New York University\\
\small\ $^{4}$ Redwood Center for Theoretical Neuroscience, UC Berkeley\\
}

\usepackage{array}
\usepackage{dcolumn}
\usepackage{tabularx} 
\newcolumntype{P}[1]{>{\centering\arraybackslash}m{#1}}

\usepackage[a4paper]{geometry}
\usepackage{tabularx}
\usepackage{lipsum}
\begin{document}

\maketitle
\begin{abstract}
Transformer networks have revolutionized NLP representation learning since they were introduced. Though a great effort has been made to explain the representation in transformers, it is widely recognized that our understanding is not sufficient. One important reason is that there lack enough visualization tools for detailed analysis. In this paper, we propose to use dictionary learning to open up these `black boxes' as linear superpositions of transformer factors. Through visualization, we demonstrate the hierarchical semantic structures captured by the transformer factors, e.g., word-level polysemy disambiguation, sentence-level pattern formation, and long-range dependency. While some of these patterns confirm the conventional prior linguistic knowledge, the rest are relatively unexpected, which may provide new insights. We hope this visualization tool can bring further knowledge and a better understanding of how transformer networks work. The code is available at \url{https://github.com/zeyuyun1/TransformerVis}.
\end{abstract}

\section{Introduction}
Though the transformer networks \cite{vaswani2017attention, devlin2018BERT} have achieved great success, our understanding of how they work is still fairly limited. This has triggered increasing efforts to visualize and analyze these ``black boxes''. Besides a direct visualization of the attention weights, most of the current efforts to interpret transformer models involve ``probing tasks''. They are achieved by attaching a light-weighted auxiliary classifier at the output of the target transformer layer. Then only the auxiliary classifier is trained for well-known NLP tasks like part-of-speech (POS) Tagging, Named-entity recognition (NER) Tagging, Syntactic Dependency, etc. \citet{tenney2019you} and \citet{liu-etal-2019-linguistic} show transformer models have excellent performance in those probing tasks. These results indicate that transformer models have learned the language representation related to the probing tasks. Though the probing tasks are great tools for interpreting language models, their limitation is explained in \citet{Anna2020Bertology}. We summarize the limitation into three major points:
\begin{itemize}
    \item Most probing tasks, like POS and NER tagging, are too simple. A model that performs well in those probing tasks does not reflect the model’s true capacity.
    \item Probing tasks can only verify whether a certain prior structure is learned in a language model. They can not reveal the structures beyond our prior knowledge.
    \item It's hard to locate where exactly the related linguistic representation is learned in the transformer. 
\end{itemize}
Efforts are made to remove those limitations and make probing tasks more diverse. For instance, \citet{hewitt-manning-2019-structural} proposes ``structural probe'', which is a much more intricate probing task. \citet{Zhengbao2020Know} proposes to generate specific probing tasks automatically. Non-probing methods are also explored to relieve the last two limitations. For example, \citet{emily2019VisBert} visualizes embedding from BERT using UMAP and shows that the embeddings of the same word under different contexts are separated into different clusters. \citet{Kawin2019Contextual} analyzes the similarity between embeddings of the same word in different contexts. Both of these works show transformers provide a context-specific representation.

\citet{faruqui-etal-2015-sparse, arora2018linear, zhang2019word} demonstrate how to use dictionary learning to explain, improve, and visualize the uncontextualized word embedding representations. In this work, we propose to use dictionary learning to alleviate the limitations of the other transformer interpretation techniques. Our results show that dictionary learning provides a powerful visualization tool, leading to some surprising new knowledge.

\section{Method}
{\noindent \bf Hypothesis: contextualized word embedding as a sparse linear superposition of transformer factors.} It is shown that word embedding vectors can be factorized into a sparse linear combination of word factors \cite{arora2018linear, zhang2019word}, which correspond to elementary semantic meanings. An example is:
\begin{align*}
    \text{apple} =& 0.09``\text{dessert}" + 0.11``\text{organism}" + 0.16\\ ``\text{fruit}" 
    & + 0.22``\text{mobile\&IT}" + 0.42``\text{other}".
\end{align*}
We view the latent representation of words in a transformer as contextualized word embedding. Similarly, we hypothesize that a contextualized word embedding vector can also be factorized as a sparse linear superposition of a set of elementary elements, which we call \textit{transformer factors}. The exact definition will be presented later in this section.  

\begin{figure}[h]
\centering
\includegraphics[width=0.3\textwidth]{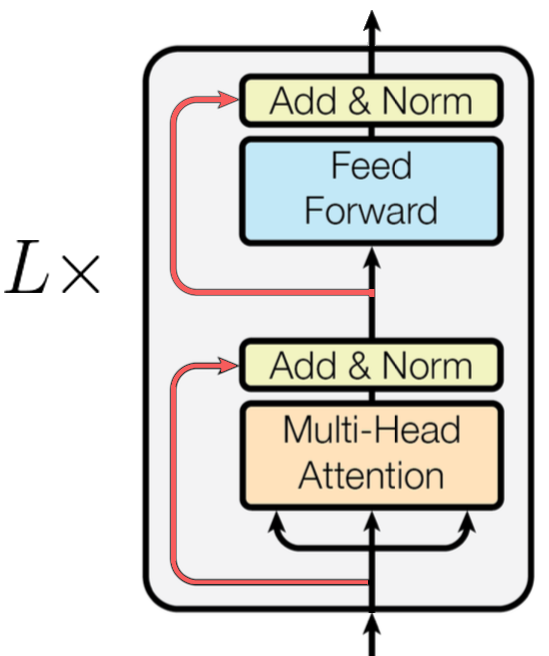}
\caption{Building block (layer) of transformer}
\end{figure}
Due to the skip connections in each of the transformer blocks, we hypothesize that the representation in any layer would be a superposition of the hierarchical representations in all of the lower layers. As a result, the output of a particular transformer block would be the sum of all of the modifications along the way. Indeed, we verify this intuition with the experiments. Based on the above observation, we propose to learn a single dictionary for the contextualized word vectors from different layers' output.

\vspace{0.1in}
{\noindent \bf To learn a dictionary of transformer factors with non-negative sparse coding.}

Given a set of tokenized text sequences, we collect the contextualized embedding of every word using a transformer model. We define the set of all word embedding vectors from $l$th layer of transformer model as $X^{(l)}$. Furthermore, we collect the embeddings across all layers into a single set $X = X^{(1)} \cup X^{(2)} \cup \cdots \cup X^{(L)}$.

By our hypothesis, we assume each embedding vector $x \in X$ is a sparse linear superposition of \textit{transformer factors}:

\begin{equation}\label{sparse}
    x = \Phi \alpha + \epsilon, \ s.t. \  \alpha \succeq 0,
\end{equation}
where $\Phi\in{\rm I\!R}^{d\times m}$ is a dictionary matrix with columns $\Phi_{:,c}\ $, $\bm{\alpha}\in{\rm I\!R}^m$ is a sparse vector of coefficients to be inferred and $\bm{\epsilon}$ is a vector containing independent Gaussian noise samples, which are assumed to be small relative to $\bm{x}$.  Typically $m>d$ so that the representation is {\em overcomplete}. This inverse problem can be efficiently solved by FISTA algorithm \cite{beck2009fast}. The dictionary matrix $\Phi$ can be learned in an iterative fashion by using non-negative sparse coding, which we leave to the appendix section \ref{sec:optimization}. Each column $\Phi_{:,c}\ $ of $\Phi$ is a {\it transformer factor} and its corresponding sparse coefficient $\bm{\alpha}_c$ is its activation level.

\vspace{0.1in}
{\noindent \bf Visualization by top activation and LIME interpretation.}
An important empirical method to visualize a feature in deep learning is to use the input samples, which trigger the top activation of the feature \cite{zeiler2014visualizing}. We adopt this convention. As a starting point, we try to visualize each of the dimensions of a particular layer, $X^{(l)}$. Unfortunately, the hidden dimensions of transformers are not semantically meaningful, which is similar to the uncontextualized word embeddings \cite{zhang2019word}.

\begin{figure*}[ht]%
    \centering
    \subfloat[\centering layer 0 \label{CWF 17 layer 0} ]{{\includegraphics[width=0.30\linewidth]{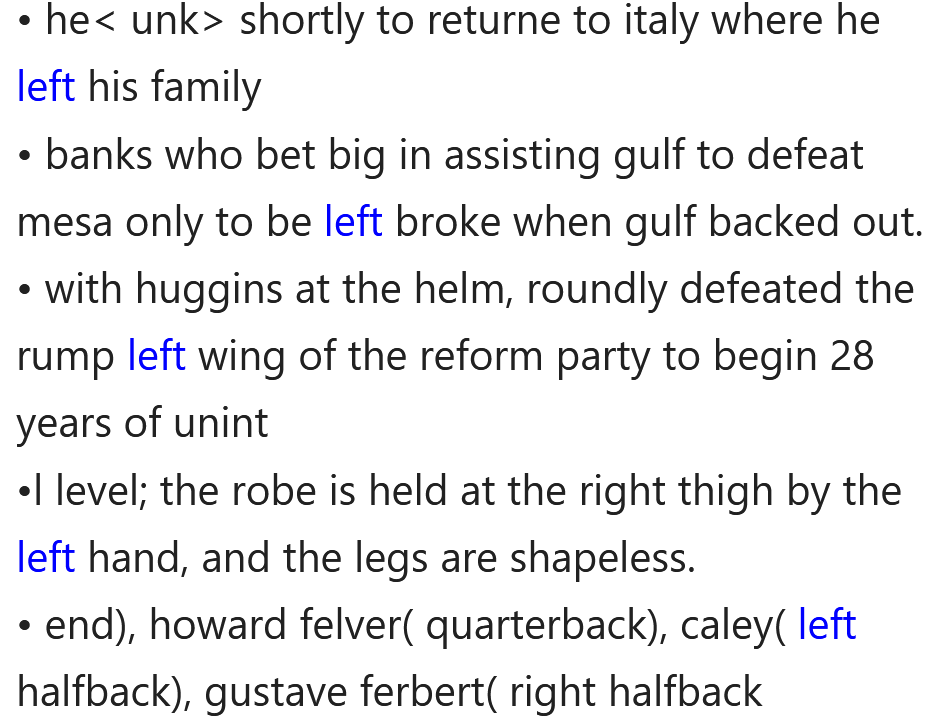} }}%
    \quad
    \subfloat[\centering layer 2]{{\includegraphics[width=0.30\linewidth]{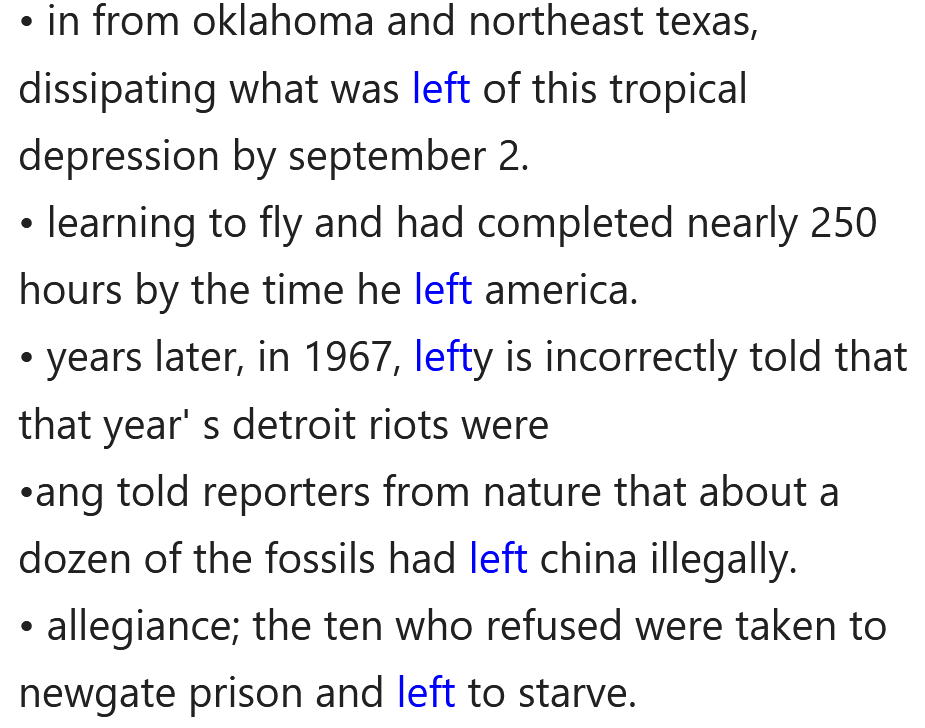} }}%
    \quad
    \subfloat[\centering layer 6]{{\includegraphics[width=0.33\linewidth]{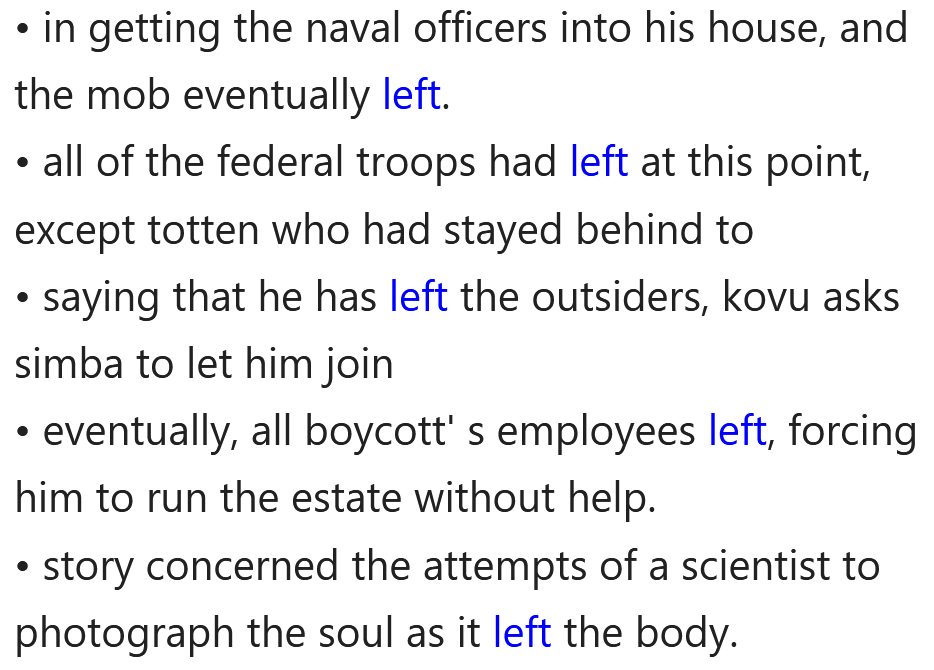} }}%

    \caption{Visualization of a low-level transformer factor, $\Phi_{:,30}$ at different layers.
    (a), (b) and (c) are the top-activated words and contexts for $\Phi_{:,30}$ in layer-$0$, $2$ and $4$ respectively. We can see that at layer-$0$, this transformer factor corresponds to word vectors that encode the word ``left'' with different senses. In layer-2, a majority of the top activated words ``left'' correspond to a single sense, "leaving, exiting."  In layer 4, all of the top-activated words ``left'' have corresponded to the same sense, "leaving, exiting." Due to space limitations, we invite the readers to use our \href{https://transformervis.github.io/transformervis/}{website} to see more of those disambiguation effects. 
    }%
    \label{CWF 17}%
\end{figure*}

Instead, we can try to visualize the transformer factors. For a transformer factor $\Phi_{:,c}$ and for a layer-$l$, we denote the 1000 contextualized word vectors with the largest sparse coefficients  $\alpha^{(l)}_c$ as $X^{(l)}_c \subset X^{(l)}$, which correspond to 1000 different sequences. For example, Figure~ \ref{CWF 17} shows the top 5 words that activated transformer factor-17 $\Phi_{:,17}$ at layer-$0$, layer-$2$, and layer-$6$ respectively. Since a contextualized word vector is generally affected by many tokens in the sequence, we can use LIME \cite{DBLP:journals/corr/RibeiroSG16} to assign a weight to each token in the sequence to identify their relative importance to $\alpha_c$. The detailed method is left to Section~\ref{sec:experiments}.

\begin{figure}%
    \centering
    \subfloat[\centering]{{\includegraphics[width=0.5\linewidth]{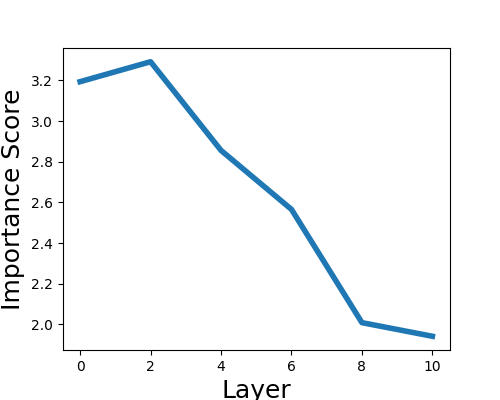}\label{low} }}%
    \subfloat[\centering]{{\includegraphics[width=0.5\linewidth]{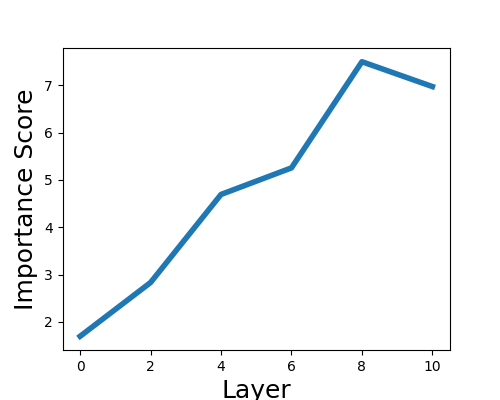} \label{mid}}}%
    \caption{Importance score (IS) across all layers for two different transformer factors. (a) This figure shows a typical IS curve of a transformer factor corresponding to low-level information. (b) This figure shows a typical IS curve of a transformer factor corresponds to mid-level information.}%
    \label{importance score}%
\end{figure}

\vspace{0.1in}
{\noindent \bf To determine low-, mid-, and high-level transformer factors with importance score.} As we build a single dictionary for all of the transformer layers, the semantic meaning of the transformer factors has different levels. While some of the factors appear in lower layers and continue to be used in the later stages, the rest of the factors may only be activated in the higher layers of the transformer network. A central question in representation learning is: ``where does the network learn certain information?'' To answer this question, we can compute an ``importance score'' for each transformer factor $\Phi_{:,c}$ at layer-$l$ as $I^{(l)}_c$. $I^{(l)}_c$ is the average of the largest 1000 sparse coefficients $\alpha^{(l)}_c$'s, which correspond to $X^{(l)}_c$. We plot the importance scores for each transformer factor as a curve is shown in Figure~\ref{importance score}. We then use these importance score (IS) curves to identify which layer a transformer factor emerges. Figure~\ref{low} shows an IS curve peak in the earlier layer. The corresponding transformer factor emerges in the earlier stage, which may capture lower-level semantic meanings. In contrast, Figure~\ref{mid} shows a peak in the higher layers, which indicates the transformer factor emerges much later and may correspond to mid- or high-level semantic structures. More subtleties are involved when distinguishing between mid-level and high-level factors, which will be discussed later.

\begin{figure*}%
    \centering

    \subfloat[\centering layer 4]{{\includegraphics[width=0.30\linewidth]{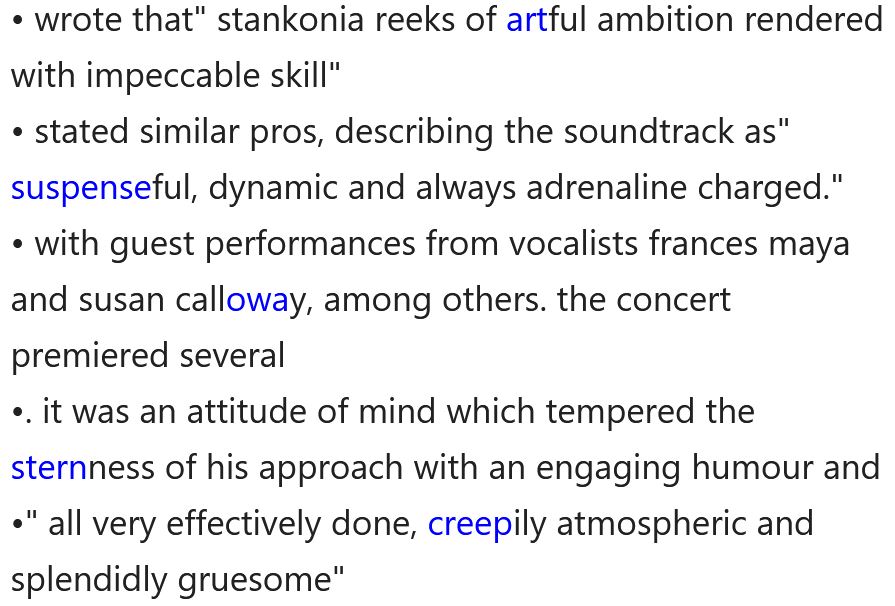} }}%
    \subfloat[\centering layer 6]{{\includegraphics[width=0.33\linewidth]{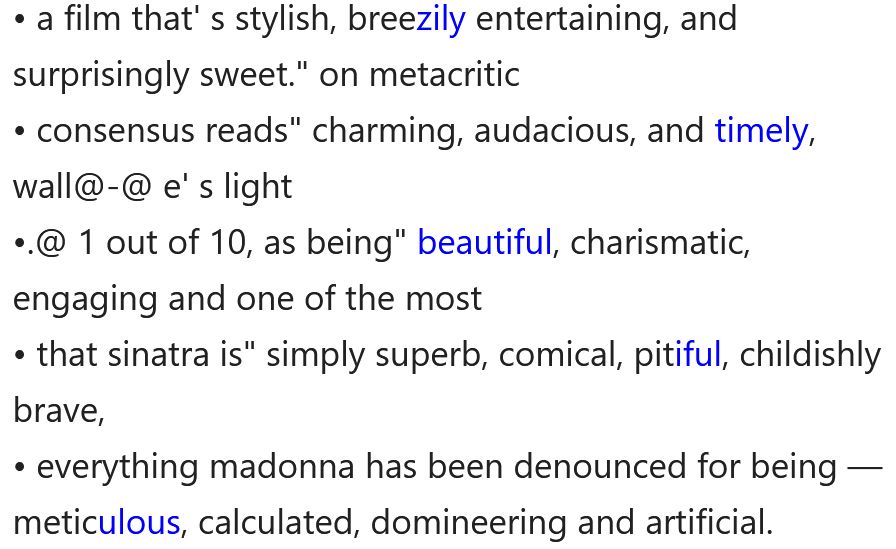} }}%
    \subfloat[\centering layer 8]{{\includegraphics[width=0.33\linewidth]{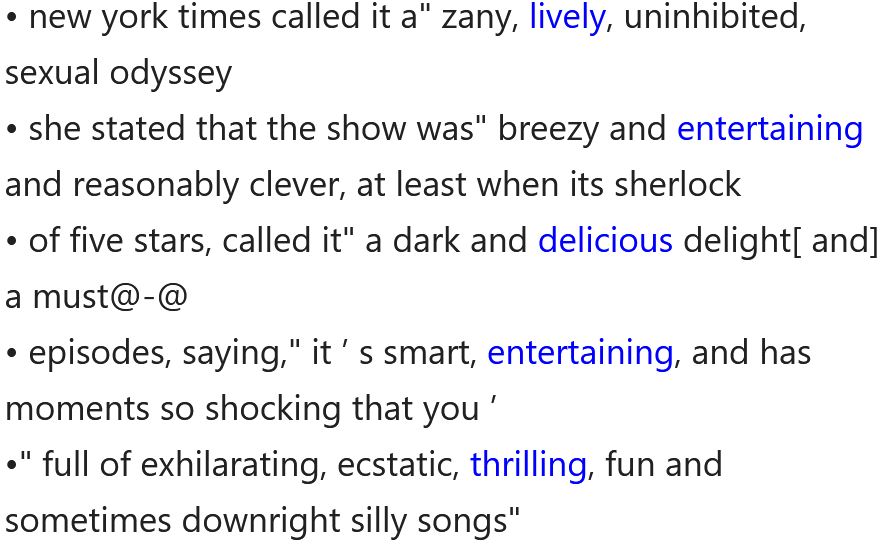} }}%
    
    \caption{Visualization of a mid-level transformer factor. (a), (b), (c) are the top 5 activated words and contexts for this transformer factor in layer-$4$, $6$, and $8$ respectively. Again, the position of the word vector is marked \textcolor{blue}{blue}. Please notice that sometimes only a part of a word is marked blue. This is due to that BERT uses word-piece tokenizer instead of whole word tokenizer. This transformer factor corresponds to the pattern of ``consecutive adjective''. As shown in the figure, this feature starts to develop at layer-$4$ and fully develops at layer-$8$. }%
    \label{CWF 35}%
\end{figure*}

An important characteristic is that the IS curve for each transformer factor is relatively smooth. This indicates if a vital feature is learned in the beginning layers, it won't disappear in later stages. Instead, it will be carried all the way to the end with gradually decayed weight since many more features would join along the way. Similarly, abstract information learned in higher layers is slowly developed from the early layers. Figure~\ref{CWF 17} and \ref{CWF 35} confirm this idea, which will be explained in the next section.

\begin{table*}[!h]
    \small
    \centering
    \begin{tabular}{|m{0.05\linewidth} | m{0.6\linewidth} | m{0.25\linewidth} |}
      \hline
         & Top 3 activated words and their contexts & Explanation \\
      \hline
        $\Phi_{:,2}$ & • that snare shot sounded like somebody' d kicked open the door to your \textcolor{blue}{mind}".\newline• i became very frustrated with that and finally made up my \textcolor{blue}{mind} to start getting back into things."\newline• when evita asked for more time so she could make up her \textcolor{blue}{mind}, the crowd demanded," ¡ ahora, evita,<&• Word ``mind''\newline • 
      Noun \newline
       • Definition: the element of a person that enables them to be aware of the world and their experiences.\\
      \hline
        $\Phi_{:,16}$ &•nington joined the five members xero and the band was renamed to linkin \textcolor{blue}{park}.\newline• times about his feelings about gordon, and the price family even sat away from \textcolor{blue}{park}' s supporters during the trial itself.\newline• on 25 january 2010, the morning of \textcolor{blue}{park}' s 66th birthday, he was found hanged and unconscious in his
        &  • Word ``park'' \newline • Noun \newline •  Definition: a common first and last name \\
      \hline
        $\Phi_{:,30}$ & • saying that he has \textcolor{blue}{left} the outsiders, kovu asks simba to let him join his pride\newline• eventually, all boycott' s employees \textcolor{blue}{left}, forcing him to run the estate without help.\newline• the story concerned the attempts of a scientist to photograph the soul as it \textcolor{blue}{left} the body. &  • Word ``left" \newline
        • Verb \newline
        • Definition: leaving, exiting\\
      \hline
        $\Phi_{:,33}$ &• forced to visit the sarajevo television station at night and to film with as little \textcolor{blue}{light} as possible to avoid the attention of snipers and bombers.\newline• by the modest, cream@-@ colored attire in the airy, \textcolor{blue}{light}@-@ filled clip.\newline• the man asked her to help him carry the case to his car, a \textcolor{blue}{light}@-@ brown volkswagen beetle. & • Word ``light'' \newline • Noun \newline
        • Definition: the natural agent that stimulates sight and makes things visible\\

      \hline

    \end{tabular}
    \caption{Several examples of low-level transformer factors. Their top-activated words in layer 4 are marked \textcolor{blue}{blue}, and the corresponding contexts are shown as examples for each transformer factor. As shown in the table, nearly all of the top-activated words are disambiguated into a single sense. Please note the last example of $\Phi_{:,33}$ is a rare exception, the reader may check the appendix to see a more complete list. More examples, top-activated words and contexts are provided in Appendix.  } 
    \label{low level table}
\end{table*}

\begin{figure*}[!h]%
    \centering
    \subfloat[\centering \label{compare} ]{{\includegraphics[width=0.45\linewidth]{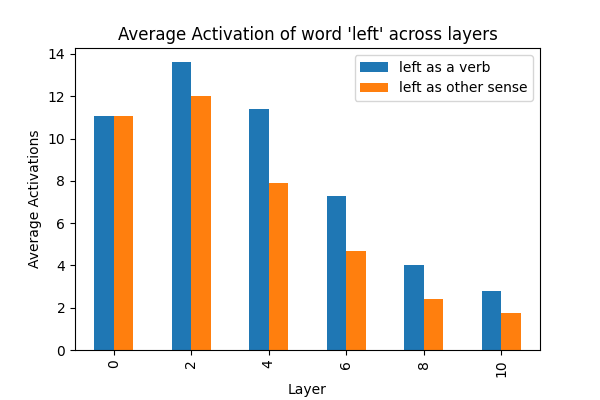} }}%
    \subfloat[\centering \label{linear}]{{\includegraphics[width=0.5\linewidth]{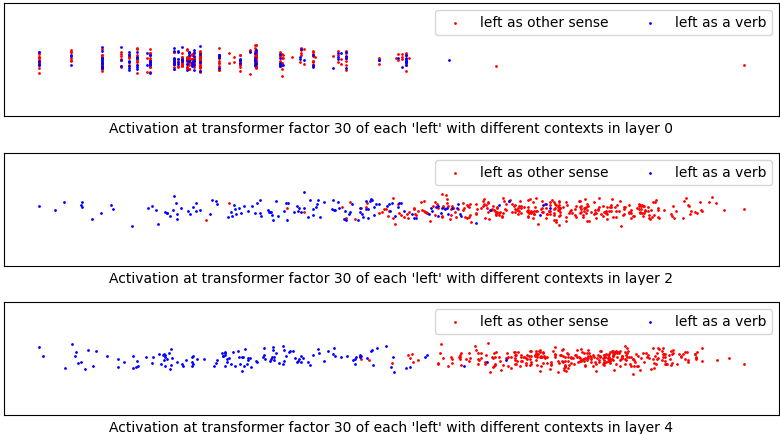} }}%

    \caption{(a) Average activation of $\Phi_{:,30}$ for word vector ``left'' across different layers. (b) Instead of averaging, we plot the activation of all ``left'' with different contexts in layer-$0$, $2$, and $4$. Random noise is added to the y-axis to prevent overplotting. The activation of $\Phi_{:,30}$ for two different word senses of ``left'' is blended together in layer-$0$. They disentangle to a great extent in layer-$2$ and nearly separable in layer-$4$ by this single dimension.}%
    \label{linear mul}%
\end{figure*}

 \begin{table}[ht]
    \small
    \centering
    \begin{tabular}{|p{0.4\linewidth} | P{0.13\linewidth} |P{0.10\linewidth} |P{0.15\linewidth} |}
      \hline
        & Precision (\%) & Recall (\%) & F1 score (\%) \\
      \hline
      Average perceptron POS tagger & \vspace{0.1in} 92.7 & \vspace{0.1in} 95.5 & \vspace{0.1in} 94.1 \\
      \hline
      Finetuned BERT base model for POS task   & \vspace{0.1in} 97.5 & \vspace{0.1in} 95.2 & \vspace{0.1in} 96.3 \\
      \hline
      Logistic regression classifier with activation of $\Phi_{:,30}$ at layer 4 & \vspace{0.18in} 97.2 & \vspace{0.18in} {\bf 95.8} & \vspace{0.18in} {\bf 96.5} \\
      \hline
    \end{tabular}
    \caption{Evaluation of binary POS tagging task: predict whether or not ``left'' in a given context is a verb.} 
    \label{evaluation}
    \vspace{-0.1in}
\end{table}

\begin{figure*}%
    \centering

    \subfloat[\centering layer 4]{{\includegraphics[width=0.33\linewidth]{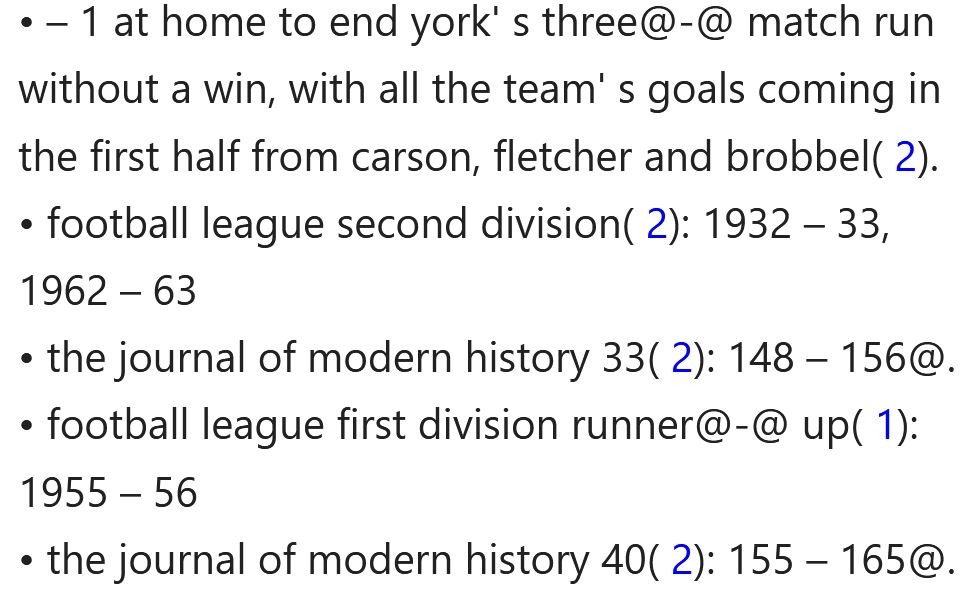} }}%
    \subfloat[\centering layer 6]{{\includegraphics[width=0.31\linewidth]{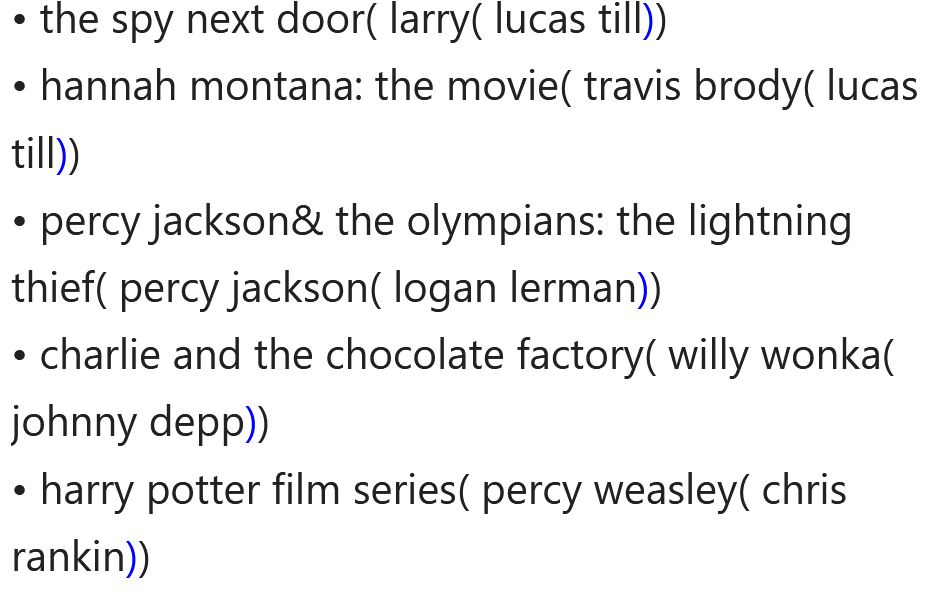} }}%
    \subfloat[\centering layer 8]{{\includegraphics[width=0.33\linewidth]{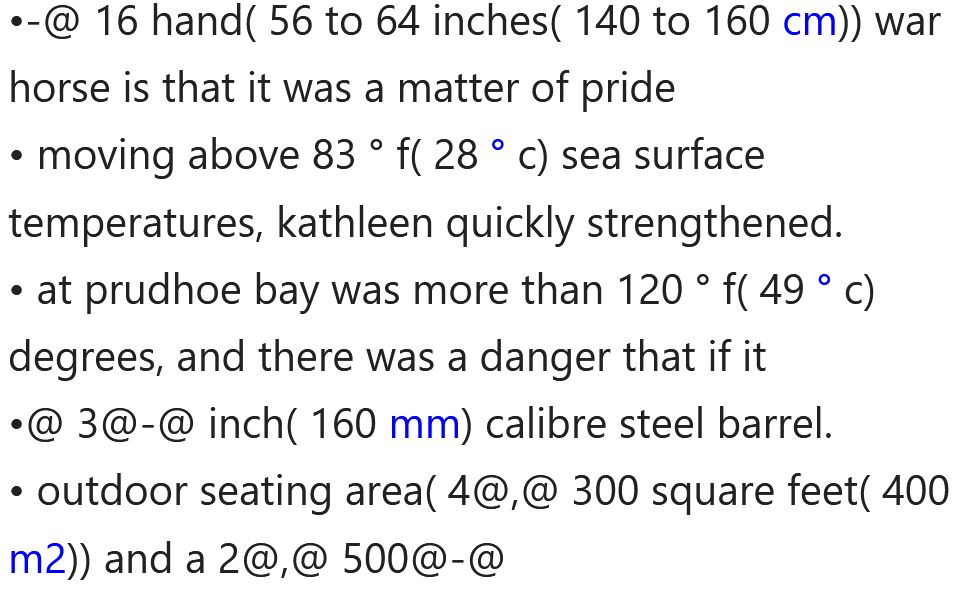} }}%
    
    \caption{Another example of a mid-level transformer factor visualized at layer-$4$, $6$, and $8$. The pattern that corresponds to this transformer factor is ``unit exchange''. Such a pattern is somewhat unexpected based on linguistic prior knowledge. }%
    \label{CWF 13}%
\end{figure*}

\section{Experiments and Discoveries}
\label{sec:experiments}

We use a 12-layer pre-trained BERT model \cite{PretrainedBERT,devlin2018BERT} and freeze the weights. Since we learn a single dictionary of transformer factors for all of the layers in the transformer, we show that these transformer factors correspond to different levels of semantic or syntactic patterns. The patterns can be roughly divided into three categories: word-level disambiguation, sentence-level pattern formation, and long-range dependency. In the following, we provide detailed visualization for each pattern category. Due to the space limit, only a small amount of the factors are demonstrated in the paper. To alleviate the ``cherry-picking'' 
bias, we also build a \href{https://transformervis.github.io/transformervis/}{website} for the interested readers to play with these results.

\vspace{0.1in}
{\noindent \bf Low-level: word-level polysemy disambiguation.}
While the input embedding of a token contains polysemy, we find transformer factors with early IS curve peaks usually correspond to a specific word-level meaning. By visualizing the top activation sequences, we can see how word-level disambiguation is gradually developed in a transformer. 

 We show how the disambiguation effect develops progressively through each layer in Figure~\ref{CWF 17}. In Figure~\ref{CWF 17},
 the top 5 activated words and their contexts for transformer factor $\Phi_{:,30}$ in different layers are listed. The top activated words in layer 0 contain the word ``left'' varying senses, which is being mostly disambiguated in layer 2 albeit not completely. In layer 4, the word ``left'' is fully disambiguated since the top-activated word contains only ``left'' with the word sense ``leaving, exiting.''
 We also show more examples of those types of transformer factors in Table~\ref{low level table}: for each transformer factor, we list out the top 3 activated words and their contexts in layer 4. As shown in the table, nearly all top-activated words are disambiguated into a single sense.

Further, we can quantify the quality of the disambiguation ability of the transformer model. In the example above, since the top 1000 activated words and contexts are ``left'' with only the word sense ``leave, exiting'', we can assume ``left'' when used as a verb, triggers higher activation in $\Phi_{:,30}$ than ``left'' used as other sense of speech. We can verify this hypothesis using a human-annotated corpus: Brown corpus \cite{francis79browncorpus}. In this corpus, each word is annotated with its corresponding part-of-speech. We collect all the sentences contains the word ``left'' annotated as a verb in one set and sentences contains ``left'' annotated as other part-of-speech. As shown in Figure~\ref{compare}, in layer 0, the average activation of $\Phi_{:,30}$ for the word ``left'' marked as a verb is no different from ``left'' as other senses. However, at layer 2, ``left'' marked as a verb triggers a higher activation of $\Phi_{:,30}$. In layer 4, this difference further increases, indicating disambiguation develops progressively across layers. In fact, we plot the activation of ``left'' marked as verb and the activation of other ``left'' in Figure~\ref{linear}. In layer 4, they are nearly linearly separable by this single feature. Since each word ``left'' corresponds to an activation value, we can perform a logistic regression classification to differentiate those two types of ``left''. From the result shown in Figure~\ref{compare}, it is pretty fascinating to see that the disambiguation ability of just $\Phi_{:,30}$ is better than the other two classifiers trained with supervised data. This result confirms that  disambiguation is indeed done in the early part of pre-trained transformer model and we are able to detect it via dictionary learning. 

\begin{table*}[!h]
    \small
    \centering
    \begin{tabular}{|P{0.05\linewidth} | m{0.45\linewidth} | m{0.2\linewidth} | P{0.03\linewidth} | P{0.03\linewidth} | P{0.03\linewidth} | P{0.03\linewidth} | }
      \hline
        & 2 example words and their contexts with high activation & Patterns & L4 (\%) & L6 (\%)& L8 (\%)& L10 (\%)\\
      \hline
       $\Phi_{:,13}$ & • the steel pipeline was about 20 ° f(- 7 \textcolor{blue}{°} c) degrees. 
       
       • hand( 56 to 64 inches( 140 to 160 \textcolor{blue}{cm})) war horse is that it was a & Unit exchange with parentheses &0 & 0 &64.5&95.5\\
      \hline
      $\Phi_{:,42}$ & • he died at the hospice of lancaster county from heart
      
      • holly' s drummer carl bunch suffered \textcolor{blue}{frost}bite to his toes( while aboard the
      \textcolor{blue}{ai}lments on 23 june 2007. & Something unfortunate happened &94.0&100&100&100\\
      \hline
      $\Phi_{:,50}$ & • hurricane pack 1 was a re\textcolor{blue}{vam}ped version of story mode; 
      
      • in 1998, the categories were \textcolor{blue}{re}titled best short form music video, and best & Doing something again, or making something new again &74.5&100&100&100 \\
      \hline
      $\Phi_{:,86}$ & • he finished the 2005 – \textcolor{blue}{06} season with 21 appearances and seven goals.
      
      • of an offensive game, finishing off the 2001 – \textcolor{blue}{02} season with 58 points in the 47 games
      & Consecutive years, used in foodball season naming &0&100&85.0&95.5\\
      \hline
      $\Phi_{:,102}$ & • the most prominent of which was bishop abel \textcolor{blue}{mu}zorewa' s united african national council 
      
      • ralambo' s father, and\textcolor{blue}{riam}anelo, had established rules of succession by
      & African names &99.0&100&100&100\\
      \hline
      $\Phi_{:,125}$ & • music writer \textcolor{blue}{jeff} weiss of pitchfork describes the" enduring image" 
      
      • club reviewer \textcolor{blue}{erik} adams wrote that the episode was a perfect mix & Describing someone in a paraphrasing style. Name, Career &15.5&99.0&100&98.5
      \\
      \hline
      $\Phi_{:,184}$ & • the world wide fund for nature( \textcolor{blue}{wwf}) announced in 2010 that a biodiversity study from\newline• fm) was halted by the federal communications commission( \textcolor{blue}{fcc}) due to a complaint that the company buying & Institution with abbreviation &0&15.5&39.0&63.0 \\ 
      \hline
      $\Phi_{:,193}$ &• 74, 22@,@ 500 vietnamese during 1979 \textcolor{blue}{–} 92, over 2@,@ 500 bosnian\newline •, the russo@-@ turkish war of 1877 \textcolor{blue}{–} 88 and the first balkan war in 1913.& Time span in years &97.0&95.5&96.5&95.5 \\ 
      \hline
      $\Phi_{:,195}$ & •s, hares, badgers, foxes, \textcolor{blue}{weasel}s, ground squirrels, mice, hamsters\newline•-@ watching, boxing, chess, cycling, \textcolor{blue}{drama}, languages, geography, jazz and other music& Consecutive of noun (Enumerating) &8.0&98.5&100&100 \\ 
      \hline
      $\Phi_{:,225}$ & • technologist at the united states marine hospital in key \textcolor{blue}{west}, florida who developed a morbid obsession for\newline• 00°,11'', w, near smith \textcolor{blue}{valley}, nevada. & Places in US, followings the convention ``city, state"&51.5&91.5&91.0&77.5\\
      \hline
    \end{tabular}
    \caption{A list of typical mid-level transformer factors. The top-activation words and their context sequences for each transformer factor at layer-$8$ are shown in the second column. We summarize the patterns of each transformer factor in the third column. The last 4 columns are the percentage of the top 200 activated words and sequences that contain the summarized patterns in layer-$4$,$6$,$8$, and $10$ respectively.} 
    \label{Mid Unexpected}
    \vspace{-0.1in}
\end{table*}

\vspace{0.1in}
\begin{table*}[!h]

    \small
    \centering
    \begin{tabular}{|P{0.05\linewidth} | m{0.45\linewidth} | m{0.30\linewidth} | P{0.05\linewidth} |}
      \hline
        & Adversarial Text & Explaination &  $\alpha_{35}$\\
      \hline
       (o) & album as "full of exhilarating, ecstatic, \textcolor{blue}{thrilling}, fun and sometimes downright silly songs"  & The original top-activated word and its context sentence for transformer factor $\Phi_{:,35}$ (not an adversarial text) & 9.5\\
      \hline
      (a) & album as "full of delightful, lively, \textcolor{blue}{exciting}, interesting and sometimes downright silly songs" & Replace the adjectives in sentence (o) with different adjectives. & 9.2 \\
      \hline
      (b) & album as "full of unfortunate, heartbroken, \textcolor{blue}{annoying}, boring and sometimes downright silly songs"  & Replace the adjectives in sentence (o) with negative adjectives. & 8.2\\
      \hline
      (c) & album as "full of [UNK], [UNK],  \textcolor{blue}{thrilling}, [UNK] and sometimes downright silly songs"  & Mask the adjectives in sentence (o) with unknown tokens. & 5.3\\
      \hline
      (d) & album as "full of \textcolor{blue}{thrilling} and sometimes downright silly songs"  & Remove the first three adjectives in sentence (o). & 7.8 \\
      \hline
      (e) & album as "full of \textcolor{blue}{natural}, smooth, rock, electronic and sometimes downright silly songs" & Replace the adjectives in sentence (o)  with neutral adjectives. & 6.2 \\
      \hline
      (f) & each participant starts the battle \textcolor{blue}{with} one balloon. these can be re@-@ inflated up to four & Use a random sentence.  & 0.0\\
      \hline
      (g) & The book is described as "innovative, \textcolor{blue}{beautiful} and brilliant". It receive the highest opinion from James Wood &  We create this sentence that contain the pattern of consecutive adjective.&  7.9 \\
      \hline
    \end{tabular}
    \caption{We construct adversarial texts similar but different to the pattern ``Consecutive adjective''. The last column shows the activation of $\Phi_{:,35}$, or $\alpha^{(8)}_{35}$, w.r.t. the blue-marked word in layer 8.} 
    \label{Adversarial Text}
\end{table*}
{\noindent \bf Mid level: sentence-level pattern formation.} We find most of the transformer factors, with an IS curve peak after layer 6, capture mid-level or high-level semantic meanings. In particular, the mid-level ones correspond to semantic patterns like phrases and sentences pattern. 

We first show two detailed examples of mid-level transformer factors. Figure~\ref{CWF 35} shows a transformer factor that detects the pattern of consecutive usage of adjectives. This pattern starts to emerge at layer 4, develops at layer 6, and becomes quite reliable at layer 8. Figure~\ref{CWF 13} shows a transformer factor, which corresponds to a pretty unexpected pattern: ``unit exchange'', e.g., 56 inches (140 cm). Although this exact pattern only starts to appear at layer 8, the sub-structures that make this pattern, e.g., parenthesis and numbers, appear to trigger this factor in layers 4 and 6. Thus this transformer factor is also gradually developed through several layers. 

While some mid-level transformer factors verify common semantic or syntactic patterns, there are also many surprising mid-level transformer factors. 
We list a few in Table~\ref{Mid Unexpected} with quantitative analysis. For each listed transformer factor, we analyze the top 200 activating words and their contexts in each layer. We record the percentage of those words and contexts that correspond to the factors' semantic pattern in Table~\ref{Mid Unexpected}. From the table, we see that large percentages of top-activated words and contexts do corresponds to the pattern we describe. It also shows most of these mid-level patterns start to develop at layer 4 or 6. More detailed examples are provided in the appendix section \ref{sec:mid}. Though it's still mysterious why the transformer network develops representations for these surprising patterns, we believe such a direct visualization can provide additional insights, which complements the ``probing tasks''.

To further confirm a transformer factor does correspond to a specific pattern, we can use constructed example words and context to probe their activation. In Table~\ref{Adversarial Text}, we construct several text sequences that are similar to the patterns corresponding to a particular transformer factor but with subtle differences. The result confirms that the context that strictly follows the pattern represented by that transformer factor triggers a high activation. On the other hand, the closer the adversarial example to this pattern, the higher activation it receives at this transformer factor. 

\vspace{0.1in}
{\noindent \bf High-level: long-range dependency.}
High-level transformer factors correspond to those linguistic patterns that span an extended range in the text. Since the IS curves of mid-level and high-level transformer factors are similar, it is difficult to distinguish those transformer factors based on their IS cures. Thus, we have to manually examine the top-activation words and contexts for each transformer factor to differentiate between mid-level and high-level transformer factors. To ease the process, we choose to use the black-box interpretation algorithm \emph{LIME} \cite{DBLP:journals/corr/RibeiroSG16} to identify the contribution of each token in a sequence. There also exist interpretation tools that specifically leverage the transformer architecture \citep{hila2021explainability,hila2021interpretability}. In the future, one could adapt those interpretation tools, which may potentially provide better visualization.

Given a sequence $s \in S$, we can treat $\alpha^{(l)}_{c,i}$, the activation of $\Phi_{:,c}$ in layer-$l$ at location $i$, as a scalar function of $s$, $f^{(l)}_{c,i}(s)$. Assume a sequence $s$ triggers a high activation $\alpha^{(l)}_{c,i}$, i.e. $f^{(l)}_{c,i}(s)$ is large. We want to know how much each token (or equivalently each position) in $s$ contributes to $f^{(l)}_{c,i}(s)$. To do so, we generated a sequence set $\mathcal{S}(s)$, where each $s'\in \mathcal{S}(s)$ is the same as $s$ except for that several random positions in $s'$ are masked by [`UNK'] (the unknown token). Then we learns a linear model $g_{w}(s')$ with weights $w \in \mathbb{R}^{T}$ to approximate $f(s')$, where $T$ is the length of sentence $s$. This can be solved as a ridge regression: $$\min_{w \in \mathbb{R}^T} \mathcal{L} (f,w,\mathcal{S}(s)) + \sigma \|w\|_2^2.$$

\begin{figure}[!h]
\centering
\includegraphics[width=0.49\textwidth]{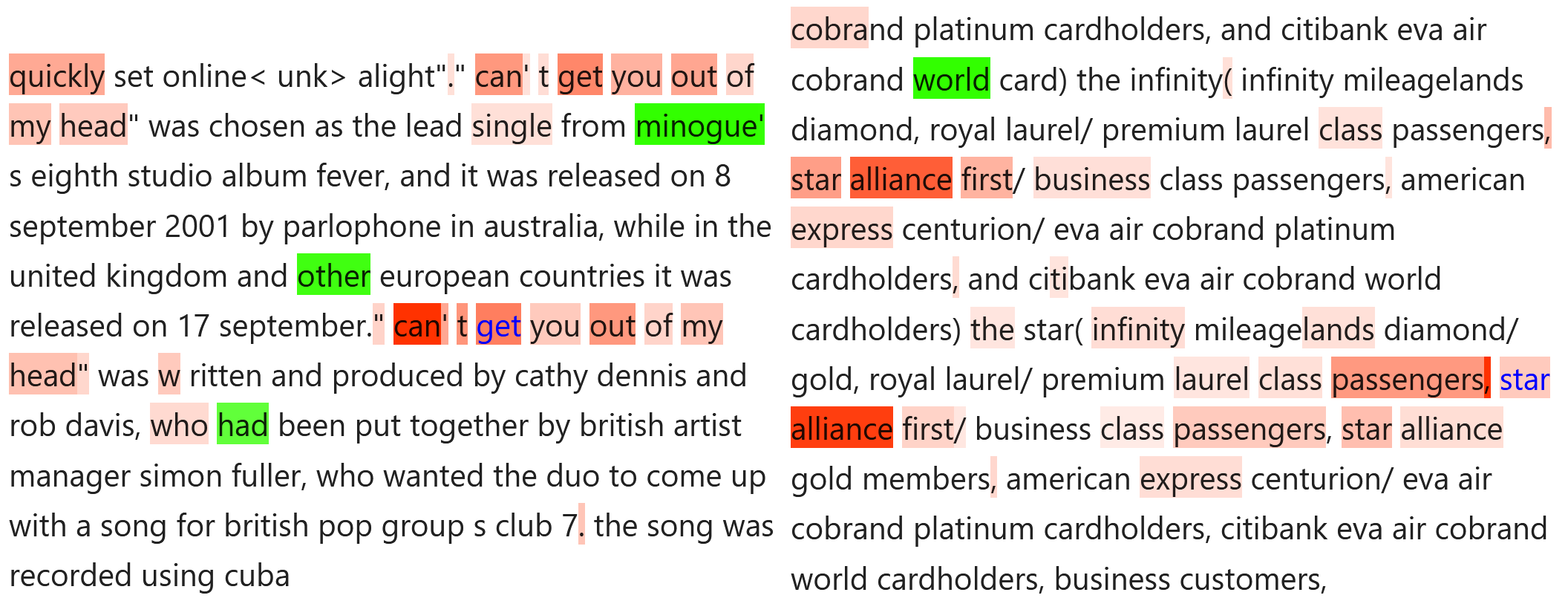}
\caption{Two examples of the high activated words and their contexts for transformer factor $\Phi_{:,297}$. We also provide the saliency map of the tokens generated using LIME. This transformer factor corresponds to the concept: ``repetitive pattern detector''. In other words, repetitive text sequences will trigger high activation of $\Phi_{:,297}$.}
\label{297}
\end{figure}

The learned weights $w$ can serve as a saliency map that reflects the ``contribution'' of each token in the sequence $s$. Like in Figure~\ref{297}, the color reflects the weights $w$ at each position. Red means the given position has positive weight and green means negative weight. The magnitude of weight is represented by the intensity. The redder a token is, the more it contributions to the activation of the transformer factor. We leave more implementation and mathematical formulation details of LIME algorithm in the appendix. 

\begin{figure}[!h]
\centering
\includegraphics[width=0.48\textwidth]{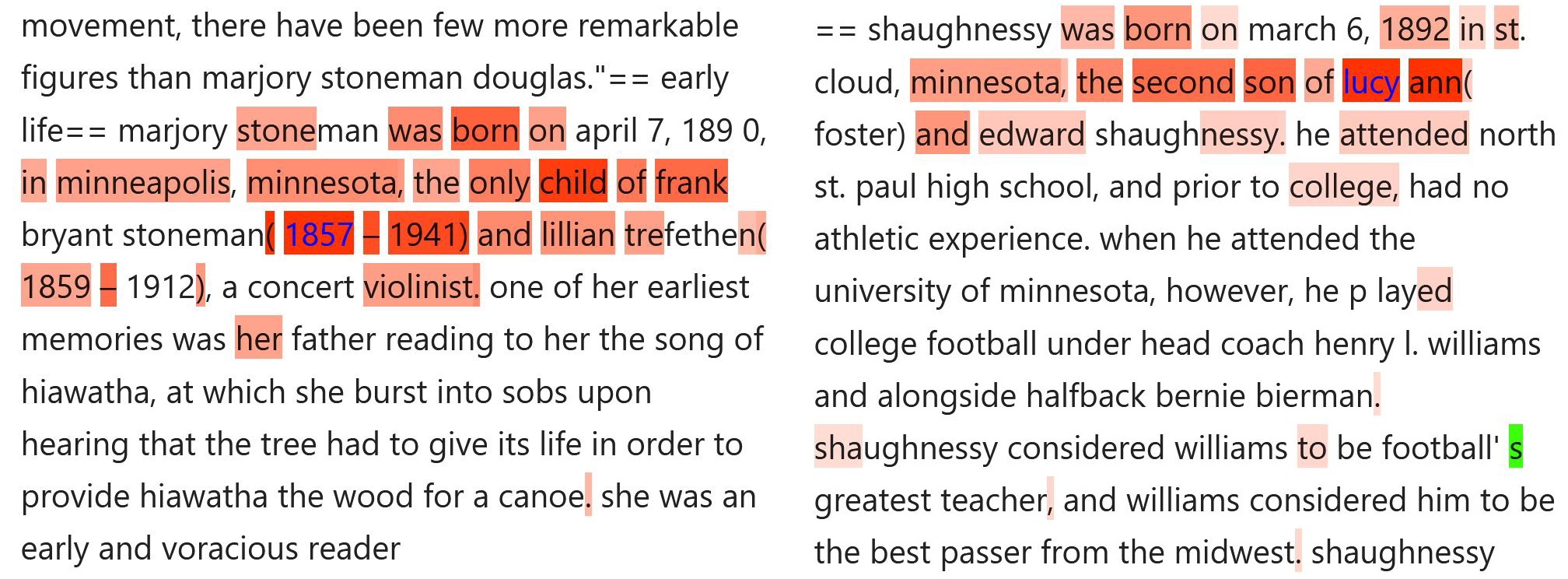}
\caption{Visualization of $\Phi_{:,322}$. This transformer factor corresponds to the concept: ``some born in some year'' in biography. All of the high-activation contexts contain the beginning of a biography. As shown in the figure, the attributes of someone, name, age, career, and familial relation all have high saliency weights.}
\vspace{-0.1in}
\label{322}
\end{figure}

We provide detailed visualization for two different transformer factors that show long-range dependency in Figure~\ref{297}, \ref{322}. Since visualization of high-level information requires more extended context, we only offer the top two activated words and their contexts for each such transformer factor. Many more will be provided in the appendix section \ref{sec:high}.

We name the pattern for transformer factor $\Phi_{:,297}$ in Figure~\ref{297} as ``repetitive pattern detector''. All top activated contexts for $\Phi_{:,297}$ contain an obvious repetitive structure. Specifically, the text snippet ``can't get you out of my head" appears twice in the first example, and the text snippet ``xxx class passenger, star alliance'' appears three times in the second example. Compared to the patterns we found in the mid-level [\ref{CWF 13}], the high-level patterns like ``repetitive pattern detector'' are much more abstract. In some sense, the transformer detects if there are two (or multiple) almost identical embedding vectors at layer-$10$ without caring what they are. Such behavior might be highly related to the concept proposed in the capsule networks \cite{sabour2017dynamic,hinton2021represent}. To further understand this behavior and study how the self-attention mechanism helps model the relationships between the features outlines an interesting future research direction.

Figure~\ref{322} shown another high-level factor, which detects text snippets related to ``the beginning of a biography''. The necessary components, day of birth as month and four-digit years, first name and last name, familial relation, and career, are all mid-level information. In Figure~\ref{322}, we see that all the information relates to biography has a high weight in the saliency map. Thus, they are all together combined to detect the high-level pattern.

\section{Discussion}
Dictionary learning has been successfully used to visualize the classical word embeddings \cite{arora2018linear, zhang2019word}. In this paper, we propose to use this simple method to visualize the representation learned in transformer networks to supplement the implicit ``probing-tasks'' methods. Our results show that the learned transformer factors are relatively reliable and can even provide many surprising insights into the linguistic structures. This simple tool can open up the transformer networks and show the hierarchical semantic or syntactic representation learned at different stages. In short, we find word-level disambiguation, sentence-level pattern formation, and long-range dependency. The idea of a neural network learns low-level features in early layers, and abstract concepts in the later stages are very similar to the visualization in CNN \cite{zeiler2014visualizing}. Dictionary learning can be a convenient tool to help visualize a broad category of neural networks with skip connections, like ResNet \cite{he2016deep}, ViT models \cite{dosovitskiy2020image}, etc. For more interested readers, we provide an interactive \href{https://transformervis.github.io/transformervis/}{website}\footnote{https://transformervis.github.io/transformervis/} for the readers to gain some further insights.

\section*{Acknowledgements}
We thank our reviewers for their detailed and insightful comments. We also thank Yuhao Zhang for his suggestions during the preparation of this paper.

\bibliography{naacl2021}

\begin{thebibliography}{24}
\expandafter\ifx\csname natexlab\endcsname\relax\def\natexlab#1{#1}\fi

\bibitem[{Pre()}]{PretrainedBERT}

\newblock Pretrained bert base model (12 layers).
\newblock \url{https://huggingface.co/bert-base-uncased}, last accessed on
  03/11/2021.

\bibitem[{Arora et~al.(2018)Arora, Li, Liang, Ma, and
  Risteski}]{arora2018linear}
Sanjeev Arora, Yuanzhi Li, Yingyu Liang, Tengyu Ma, and Andrej Risteski. 2018.
\newblock Linear algebraic structure of word senses, with applications to
  polysemy.
\newblock \emph{Transactions of the Association for Computational Linguistics},
  6:483--495.

\bibitem[{Beck and Teboulle(2009)}]{beck2009fast}
Amir Beck and Marc Teboulle. 2009.
\newblock A fast iterative shrinkage-thresholding algorithm for linear inverse
  problems.
\newblock \emph{SIAM journal on imaging sciences}, 2(1):183--202.

\bibitem[{Chefer et~al.(2020)Chefer, Gur, and Wolf}]{hila2021interpretability}
Hila Chefer, Shir Gur, and Lior Wolf. 2020.
\newblock \href {http://arxiv.org/abs/2012.09838} {Transformer interpretability
  beyond attention visualization}.
\newblock \emph{CoRR}, abs/2012.09838.

\bibitem[{Chefer et~al.(2021)Chefer, Gur, and Wolf}]{hila2021explainability}
Hila Chefer, Shir Gur, and Lior Wolf. 2021.
\newblock \href {http://arxiv.org/abs/2103.15679} {Generic attention-model
  explainability for interpreting bi-modal and encoder-decoder transformers}.
\newblock \emph{CoRR}, abs/2103.15679.

\bibitem[{Devlin et~al.(2018)Devlin, Chang, Lee, and
  Toutanova}]{devlin2018BERT}
Jacob Devlin, Ming{-}Wei Chang, Kenton Lee, and Kristina Toutanova. 2018.
\newblock \href {http://arxiv.org/abs/1810.04805} {{BERT:} pre-training of deep
  bidirectional transformers for language understanding}.
\newblock \emph{CoRR}, abs/1810.04805.

\bibitem[{Dosovitskiy et~al.(2020)Dosovitskiy, Beyer, Kolesnikov, Weissenborn,
  Zhai, Unterthiner, Dehghani, Minderer, Heigold, Gelly
  et~al.}]{dosovitskiy2020image}
Alexey Dosovitskiy, Lucas Beyer, Alexander Kolesnikov, Dirk Weissenborn,
  Xiaohua Zhai, Thomas Unterthiner, Mostafa Dehghani, Matthias Minderer, Georg
  Heigold, Sylvain Gelly, et~al. 2020.
\newblock An image is worth 16x16 words: Transformers for image recognition at
  scale.
\newblock \emph{arXiv preprint arXiv:2010.11929}.

\bibitem[{Duchi et~al.(2011)Duchi, Hazan, and Singer}]{duchi2011adaptive}
John Duchi, Elad Hazan, and Yoram Singer. 2011.
\newblock Adaptive subgradient methods for online learning and stochastic
  optimization.
\newblock \emph{Journal of Machine Learning Research}, 12(Jul):2121--2159.

\bibitem[{Ethayarajh(2019)}]{Kawin2019Contextual}
Kawin Ethayarajh. 2019.
\newblock How contextual are contextualized word representations? comparing the
  geometry of bert, elmo, and {GPT-2} embeddings.
\newblock In \emph{Proceedings of the 2019 Conference on Empirical Methods in
  Natural Language Processing and the 9th International Joint Conference on
  Natural Language Processing, {EMNLP-IJCNLP}}, pages 55--65. Association for
  Computational Linguistics.

\bibitem[{Faruqui et~al.(2015)Faruqui, Tsvetkov, Yogatama, Dyer, and
  Smith}]{faruqui-etal-2015-sparse}
Manaal Faruqui, Yulia Tsvetkov, Dani Yogatama, Chris Dyer, and Noah~A. Smith.
  2015.
\newblock Sparse overcomplete word vector representations.
\newblock In \emph{Proceedings of the 53rd Annual Meeting of the Association
  for Computational Linguistics and the 7th International Joint Conference on
  Natural Language Processing (Volume 1: Long Papers)}. Association for
  Computational Linguistics.

\bibitem[{Francis and Kucera(1979)}]{francis79browncorpus}
W.~N. Francis and H.~Kucera. 1979.
\newblock \href {http://icame.uib.no/brown/bcm.html} {Brown corpus manual}.
\newblock Technical report, Department of Linguistics, Brown University,
  Providence, Rhode Island, US.

\bibitem[{He et~al.(2016)He, Zhang, Ren, and Sun}]{he2016deep}
Kaiming He, Xiangyu Zhang, Shaoqing Ren, and Jian Sun. 2016.
\newblock Deep residual learning for image recognition.
\newblock In \emph{Proceedings of the IEEE conference on computer vision and
  pattern recognition}, pages 770--778.

\bibitem[{Hewitt and Manning(2019)}]{hewitt-manning-2019-structural}
John Hewitt and Christopher~D. Manning. 2019.
\newblock {A} structural probe for finding syntax in word representations.
\newblock In \emph{Proceedings of the 2019 Conference of the North {A}merican
  Chapter of the Association for Computational Linguistics: Human Language
  Technologies, Volume 1 (Long and Short Papers)}.

\bibitem[{Hinton(2021)}]{hinton2021represent}
Geoffrey Hinton. 2021.
\newblock How to represent part-whole hierarchies in a neural network.
\newblock \emph{arXiv preprint arXiv:2102.12627}.

\bibitem[{Jiang et~al.(2020)Jiang, Xu, Araki, and Neubig}]{Zhengbao2020Know}
Zhengbao Jiang, Frank~F. Xu, Jun Araki, and Graham Neubig. 2020.
\newblock \href {https://transacl.org/ojs/index.php/tacl/article/view/1983}
  {How can we know what language models know}.
\newblock \emph{Trans. Assoc. Comput. Linguistics}, 8:423--438.

\bibitem[{Liu et~al.(2019)Liu, Gardner, Belinkov, Peters, and
  Smith}]{liu-etal-2019-linguistic}
Nelson~F. Liu, Matt Gardner, Yonatan Belinkov, Matthew~E. Peters, and Noah~A.
  Smith. 2019.
\newblock Linguistic knowledge and transferability of contextual
  representations.
\newblock In \emph{Proceedings of the 2019 Conference of the North {A}merican
  Chapter of the Association for Computational Linguistics: Human Language
  Technologies, Volume 1 (Long and Short Papers)}. Association for
  Computational Linguistics.

\bibitem[{Reif et~al.(2019)Reif, Yuan, Wattenberg, Vi{\'{e}}gas, Coenen,
  Pearce, and Kim}]{emily2019VisBert}
Emily Reif, Ann Yuan, Martin Wattenberg, Fernanda~B. Vi{\'{e}}gas, Andy Coenen,
  Adam Pearce, and Been Kim. 2019.
\newblock Visualizing and measuring the geometry of {BERT}.
\newblock In \emph{Advances in Neural Information Processing Systems 32: Annual
  Conference on Neural Information Processing Systems, (NeurIPS)}, pages
  8592--8600.

\bibitem[{Ribeiro et~al.(2016)Ribeiro, Singh, and
  Guestrin}]{DBLP:journals/corr/RibeiroSG16}
Marco~T{\'{u}}lio Ribeiro, Sameer Singh, and Carlos Guestrin. 2016.
\newblock \href {http://arxiv.org/abs/1602.04938} {"why should {I} trust you?":
  Explaining the predictions of any classifier}.
\newblock \emph{CoRR}, abs/1602.04938.

\bibitem[{Rogers et~al.(2020)Rogers, Kovaleva, and
  Rumshisky}]{Anna2020Bertology}
Anna Rogers, Olga Kovaleva, and Anna Rumshisky. 2020.
\newblock \href {https://transacl.org/ojs/index.php/tacl/article/view/2257} {A
  primer in bertology: What we know about how {BERT} works}.
\newblock \emph{Trans. Assoc. Comput. Linguistics}, 8:842--866.

\bibitem[{Sabour et~al.(2017)Sabour, Frosst, and Hinton}]{sabour2017dynamic}
Sara Sabour, Nicholas Frosst, and Geoffrey~E Hinton. 2017.
\newblock Dynamic routing between capsules.
\newblock \emph{arXiv preprint arXiv:1710.09829}.

\bibitem[{Tenney et~al.(2019)Tenney, Xia, Chen, Wang, Poliak, McCoy, Kim,
  Van~Durme, Bowman, Das et~al.}]{tenney2019you}
Ian Tenney, Patrick Xia, Berlin Chen, Alex Wang, Adam Poliak, R~Thomas McCoy,
  Najoung Kim, Benjamin Van~Durme, Samuel~R Bowman, Dipanjan Das, et~al. 2019.
\newblock What do you learn from context? probing for sentence structure in
  contextualized word representations.
\newblock \emph{arXiv preprint arXiv:1905.06316}.

\bibitem[{Vaswani et~al.(2017)Vaswani, Shazeer, Parmar, Uszkoreit, Jones,
  Gomez, Kaiser, and Polosukhin}]{vaswani2017attention}
Ashish Vaswani, Noam Shazeer, Niki Parmar, Jakob Uszkoreit, Llion Jones,
  Aidan~N Gomez, Lukasz Kaiser, and Illia Polosukhin. 2017.
\newblock Attention is all you need.
\newblock \emph{arXiv preprint arXiv:1706.03762}.

\bibitem[{Zeiler and Fergus(2014)}]{zeiler2014visualizing}
Matthew~D Zeiler and Rob Fergus. 2014.
\newblock Visualizing and understanding convolutional networks.
\newblock In \emph{European conference on computer vision}, pages 818--833.
  Springer.

\bibitem[{Zhang et~al.(2019)Zhang, Chen, Cheung, and Olshausen}]{zhang2019word}
Juexiao Zhang, Yubei Chen, Brian Cheung, and Bruno~A Olshausen. 2019.
\newblock Word embedding visualization via dictionary learning.
\newblock \emph{arXiv preprint arXiv:1910.03833}.

\end{thebibliography}
\bibliographystyle{acl_natbib}

\clearpage
\appendix
\section*{Supplementary Materials}
\renewcommand{\thesubsection}{\Alph{subsection}}
\subsection{Importance Score (IS) Curves}
\label{sec 1}
\begin{figure}[h]%
    \centering
    \subfloat[\centering]{{\includegraphics[width=0.45\linewidth]{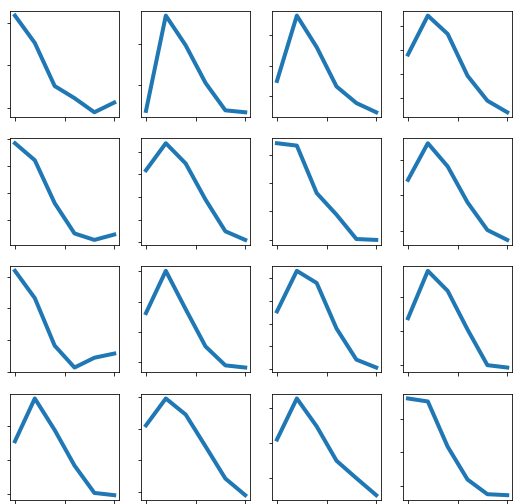}\label{low more} }}%
    \quad
    \subfloat[\centering]{{\includegraphics[width=0.44\linewidth]{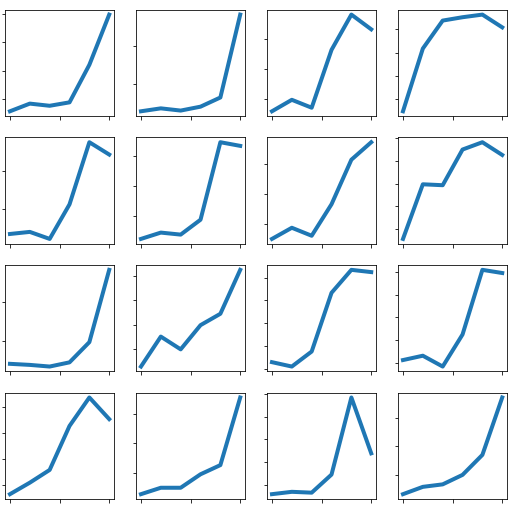} \label{high more}}}%
    \caption{(a) Importance score of 16 transformer factors corresponding to low level information. (b) Importance score of 16 transformer factors corresponds to mid level information respectively.}%
    \label{importance score more}%
\end{figure}

The importance score curve's characteristic has a strong correspondence to a transformer factor's categorization. Based on the location of the peak of an IS curve, we can classify a transformer factor as low-level, mid-level or high-level. The importance score for low-level transformer factors peak in early layers and slowly decrease across the rest of the layers. On the other hand, the importance score for mid-level and high-level transformers slowly increases and peaks at higher layers. In Figure~\ref{importance score more}, we show two sets of the examples to demonstrate the clear distinction between those two types of IS curves.

\begin{figure}[h]%
    \centering
    \subfloat[\centering]{{\includegraphics[width=0.45\linewidth]{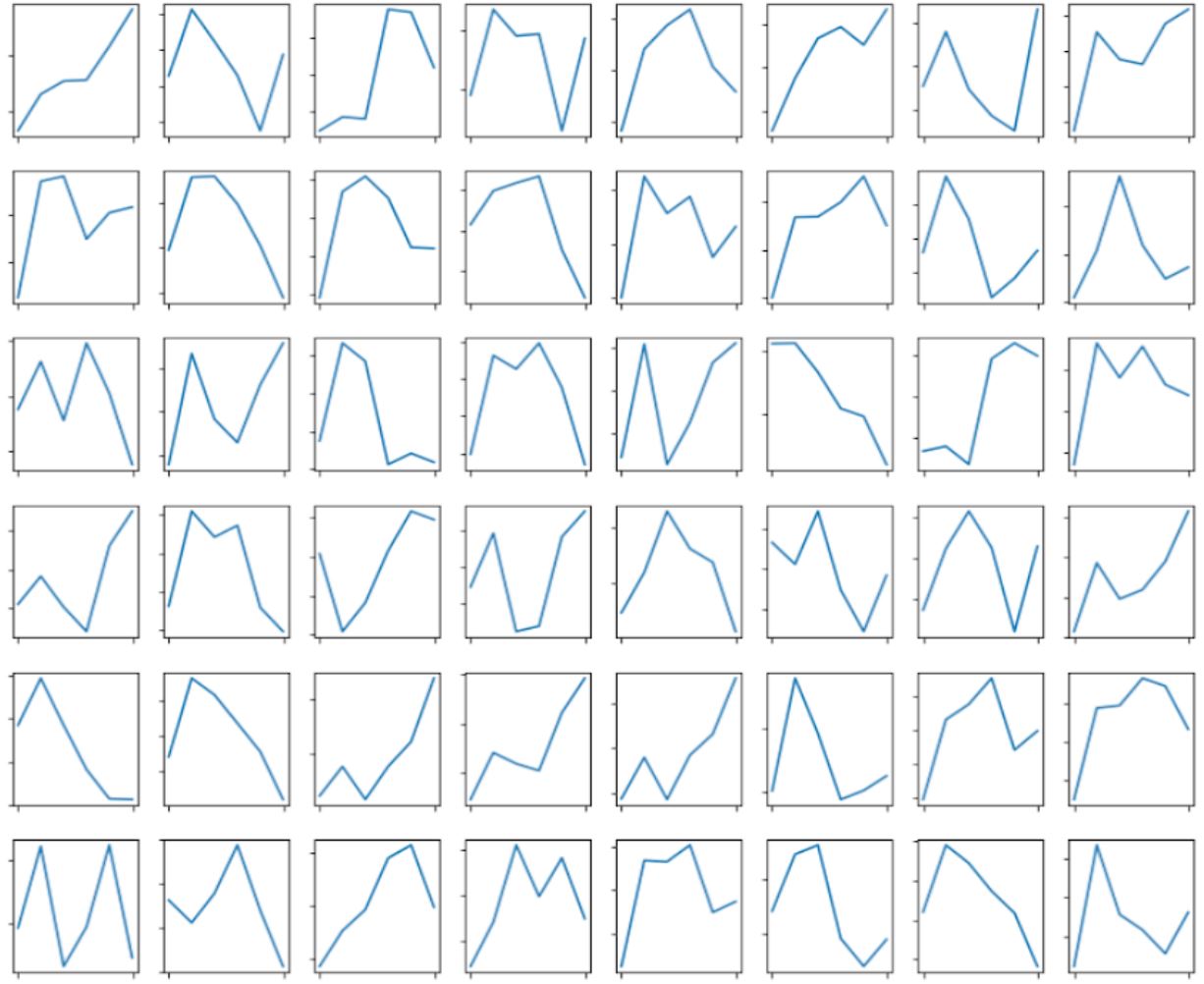}\label{no sparse} }}%
    \quad
    \subfloat[\centering]{{\includegraphics[width=0.45\linewidth]{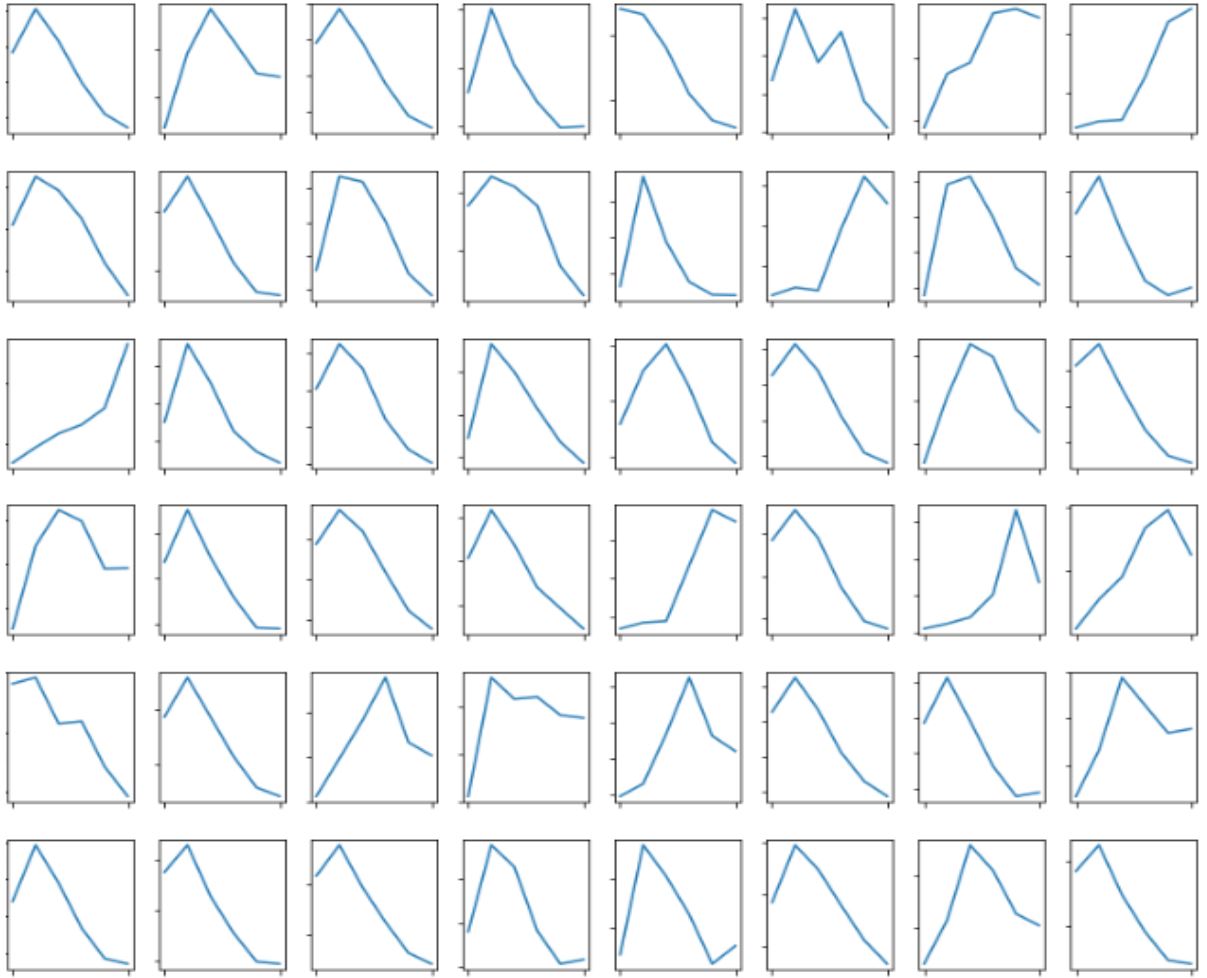} \label{sparse}}}%
    \caption{(a) Importance score calculated using certain dimension of word vectors without sparse coding.  (b) Importance score calculated using sparse coding of word vectors. }%
    \label{sparse or no sparse}%
\end{figure}

Taking a step back, we can also plot IS curve for each dimension of word vector (without sparse coding) at different layers. They do not show any specific patterns, as shown in Figure~\ref{sparse or no sparse}. This makes intuitive sense since we mentioned that each of the entries of a contextualized word embedding does not correspond to any clear semantic meaning.

\subsection{LIME: Local Interpretable Model-Agnostic Explanations}
\label{sec 2}

After we trained the dictionary $\Phi$ through non-negative sparse coding, the inference of the sparse code of a given input is $$\alpha(x) = \arg\min_{\alpha \in \mathbb{R}} || x - \Phi \alpha ||_2^2 + \lambda ||\alpha||_1 $$

For a given sentence and index pair $(s,i)$, the embedding of word $w = s[i]$ by layer $l$ of transformer is $x^{(l)} (x,i)$. Then we can abstract the inference of a specific entry of sparse code of the word vector as a black-box scalar-value function $f$: 

$$f((s,i)) = \alpha(x^{(l)}(s,i))$$

Let $RandomMask$ denotes the operation that generates perturbed version of our sentence $s$ by masking word at random location with ``[UNK]'' (unkown) tokens. For example, a masked sentence could be 

[Today is a [`UNK'],day]

Let $h$ denote a encoder for perturbed sentences compared to the unperturbed sentence $s$, such that 

\[ 
h(s)_t= \left\{
\begin{array}{ll}
      0 & \text{if } s[t] = \text{[`UNK']} \\
      1 & Otherwis \\
\end{array} 
\right. 
\]

The LIME algorithm we used to generated saliency map for each sentences is the following:

\begin{algorithm}
\caption{Explaining Sparse Coding Activation using LIME Algorithm}
\label{CHalgorithm}
\begin{algorithmic}[1]

\State $\mathcal{S} = \{h(s)\}$
\State $Y = \{f(s)\}$
\For{each $i$ in $\{1,2, ..., N\}$ }
\State $s_i' \leftarrow RandomMask(s)$
\State $\mathcal{S} \leftarrow \mathcal{S} \cup h(s_i') $
\State $Y \leftarrow Y \cup f(s_i') $
\EndFor
\State $w \leftarrow Ridge_w(\mathcal{S},Y)$
\end{algorithmic}
\end{algorithm}
\label{sec:appendix}

Where $Ridge_w$ is a weighted ridge regression defined as:  
$$w = \arg\min_{w \in R^t} ||\mathcal{S} \Omega w - Y||^2 + \lambda ||w||_2^2$$ $$\Omega = diag(d(h(\mathcal{S}_1),\vec{1}),d(h(\mathcal{S}_2),\vec{1}), \cdots, d(h(\mathcal{S}_n),\vec{1}))$$

$d(\cdot,\cdot)$ can be any metric that measures how much a perturbed sentence is different from the original sentence. If a sentence is perturbed such that every token is being masked, then the distance $h(h(s'),\vec{1})$ should be 0, if a sentence is not perturbed at all, then $h(h(s'),\vec{1})$ should be 1. We choose $d(\cdot,\cdot)$ to be cosine similarity in our implementation.

In practice, we also uses feature selection. This is done by running LIME twice. After we obtain the regression weight $w_1$ for the time, we use it to find the first $k$ indices corresponds to the entry in $w_1$ with highest absolute value. We use those $k$ index as location in the sentence and apply LIME for the second time with only those selected indices from step 1. 

Overall, the regression weight $w$ can be regarded as a saliency map. The higher the weight $w_k$ is, the more important the word $s[k]$ in the sentence since it contributes more to the activation of a specific transformer factor. 

We could also have negative weight in $w$. In general, negative weights are hard to interpret in the context of transformer factor. The activation will increase if they are removed those word correspond to negative weights. Since a transformer factor corresponds to a specific pattern, then word with negative weights are those word in a context that behaves ``opposite" of this pattern. 

\subsection{The Details of the Non-negative Sparse Coding Optimization}
\label{sec:optimization}

Let $S$ be the set of all sequences, recall how we defined word embedding using hidden state of transformer in the main section: $ X^{(l)} = \{x^{(l)}(s,i)| s \in S, i \in \left[0, len(s)\right] \}$ as the set of all word embedding at layer $l$, then the set of word embedding across all layers is defined as $$X = X^{(1)} \cup X^{(2)} \cup \cdots \cup X^{(L)}$$
In practice, we use BERT base model as our transformer model, each word embedding vector (hidden state of BERT) is dimension 768. To learn the transformer factors, we concatenate all word vector $x \in X$ into a data matrix $A$. We also defined $f(x)$ to be the frequency of the token that is embedded in word vector $x$. For example, if $x$ is the embedding of the word ``the'', it will have a much larger frequency i.e. $f(x)$ is high.

Using $f(x)$, we define the Inverse Frequency Matrix $\Omega$: $\Omega$ is a diagonal matrix where each entry on the diagonal is the square inverse frequency of each word, i.e.
$$\Omega = diag(\frac{1}{\sqrt{f(x_1)}},\frac{1}{\sqrt{f(x_2)}},...)$$

Then we use a typical iterative optimization procedure to learn the dictionary $\Phi$ described in the main section:
\begin{equation}
\min\limits_{A} \tfrac{1}{2} \| X - \Phi A\|_{F}^{2} + \lambda\sum_{i}{ \|\alpha_i\|_{1}},\ \text{s.t.}\ \alpha_i \succeq 0,
\label{appequ:sparse_coding}
\end{equation}
\begin{equation}
\min\limits_{\Phi} \tfrac{1}{2} \| X - \Phi \Omega A\|_{F}^{2},\ \|\Phi_{:,j}\|_2 \leq 1.
\label{appequ:dictionary_update}
\end{equation}

These two optimizations are both convex, we solve them iteratively to learn the transformer factors: In practice, we use minibatches contains 200 word vectors as $X$. The motivation of apply Inverse Frequency Matrix $\Omega$ is that we want to make sure all words in our vocabulary has the same contribution. When we sample our minibatch from $A$, frequent words like ``the'' and ``a'' are much likely to appear, which should receive lower weight during update.

Optimization~\ref{appequ:sparse_coding} can converge in 1000 steps using the FISTA algorithm\footnote{The FISTA algorithm can usually converge within 300 steps, we use 1000 steps nevertheless to avoid any potential numerical issue.}. We experimented with different $\lambda$ values from 0.03 to 3, and choose $\lambda=0.27$ to give results presented in this paper. Once the sparse coefficients have been inferred, we update our dictionary $\Phi$ based on Optimization~\ref{appequ:dictionary_update} by one step using an approximate second-order method, where the Hessian is approximated by its diagonal to achieve an efficient inverse \cite{duchi2011adaptive}. The second-order parameter update method usually leads to much faster convergence. Empirically, we train 200k steps and it takes roughly 2 days on a Nvidia 1080 Ti GPU.
\subsection{Hyperlinks for More Transformer Visualization}
In the following three sections, we provide visualization of more example transformer factor in low-level, mid-level, and high-level. Here's table of Contents that contain hyperlinks which direct to each level:
\begin{itemize}
    \item Low-Level: \ref{sec:low}
    \item Mid-Level: \ref{sec:mid}
    \item High-Level: \ref{sec:high}
\end{itemize}

\label{sec:hyper}

\clearpage

\subsection{Low-Level Transformer Factors}
\label{sec:low}

{\bf Transformer factor 2 in layer 4 \newline Explaination: Mind: noun, the element of a person that enables them to be aware of the world and their experiences. } \newline \newline • that snare shot sounded like somebody' d kicked open the door to your \textcolor{blue}{mind}".\newline
• i became very frustrated with that and finally made up my \textcolor{blue}{mind} to start getting back into things."\newline
• when evita asked for more time so she could make up her \textcolor{blue}{mind}, the crowd demanded," ¡ ahora, evita,<\newline
• song and watch it evolve in front of us... almost as a \textcolor{blue}{memory} in your head.\newline
• was to be objective and to let the viewer make up his or her own \textcolor{blue}{mind}."\newline
• managed to give me goosebumps, and those moments have remained on my \textcolor{blue}{mind} for weeks afterward."\newline
• rests the tir' d \textcolor{blue}{mind}, and waking loves to dream\newline
•, tracks like' halftime' and the laid back' one time 4 your \textcolor{blue}{mind}' demonstrated a[ high] level of technical precision and rhetorical dexter\newline
• so i went to bed with that on my \textcolor{blue}{mind}".\newline
•ment to a seed of doubt that had been playing on mulder' s \textcolor{blue}{mind} for the entire season".\newline
• my poor friend smart shewed the disturbance of his \textcolor{blue}{mind}, by falling upon his knees, and saying his prayers in the street\newline
• donoghue complained that lessing has not made up her \textcolor{blue}{mind} on whether her characters are" the salt of the earth or its sc\newline
• release of the new lanois@-@ produced album, time out of \textcolor{blue}{mind}.\newline
• sympathetic man to illegally" ghost@-@ hack" his wife' s \textcolor{blue}{mind} to find his daughter.\newline
• this album veered into" the corridors" of flying lotus'" own \textcolor{blue}{mind}", interpreting his guest vocalists as" disembodied phantom\newline
\newline \newline 
{\bf Transformer factor 16 in layer 4 \newline Explaination: Park: noun, 'park' as the name } \newline \newline • allmusic writer william ruhlmann said that" linkin \textcolor{blue}{park} sounds like a johnny@-@ come@-@ lately to an\newline
•nington joined the five members xero and the band was renamed to linkin \textcolor{blue}{park}.\newline
• times about his feelings about gordon, and the price family even sat away from \textcolor{blue}{park}' s supporters during the trial itself.\newline
• on 25 january 2010, the morning of \textcolor{blue}{park}' s 66th birthday, he was found hanged and unconscious in his\newline
• was her, and knew who had done it", expressing his conviction of \textcolor{blue}{park}' s guilt.\newline
• jeremy \textcolor{blue}{park} wrote to the north@-@ west evening mail to confirm that he\newline
• vanessa fisher, \textcolor{blue}{park}' s adoptive daughter, appeared as a witness for the prosecution at the\newline
• they played at< unk> for years before joining oldham athletic at boundary \textcolor{blue}{park} until 2010 when they moved to oldham borough' s previous ground,<\newline
• theme \textcolor{blue}{park} guests may use the hogwarts express to travel between hogsmead\newline
• s strength in both singing and rapping while comparing the sound to linkin \textcolor{blue}{park}.\newline
• in a statement shortly after \textcolor{blue}{park}' s guilty verdict, he said he had" no doubt" that\newline
• june 2013, which saw the band travel to rock am ring and rock im \textcolor{blue}{park} as headline act, the song was moved to the middle of the set\newline
• after spending the first decade of her life at the central \textcolor{blue}{park} zoo, pattycake moved permanently to the bronx zoo in 1982.\newline
• south park spoofed the show and its hosts in the episode" south \textcolor{blue}{park} is gay!"\newline
• harrison" sounds like he' s recorded his vocal track in one of the \textcolor{blue}{park}' s legendary caves".\newline
\newline \newline 
{\bf Transformer factor 30 in layer 4 \newline Explaination: left: verb, leaving, exiting } \newline \newline • did succeed in getting the naval officers into his house, and the mob eventually \textcolor{blue}{left}.\newline
• all of the federal troops had \textcolor{blue}{left} at this point, except totten who had stayed behind to listen to\newline
• saying that he has \textcolor{blue}{left} the outsiders, kovu asks simba to let him join his pride\newline
• eventually, all boycott' s employees \textcolor{blue}{left}, forcing him to run the estate without help.\newline
• the story concerned the attempts of a scientist to photograph the soul as it \textcolor{blue}{left} the body.\newline
• in time and will slowly improve until he returns to the point at which he \textcolor{blue}{left}.\newline
• peggy' s exit was a" non event", as" peggy just \textcolor{blue}{left}, nonsensically and at complete odds with everything we' ve\newline
• over the course of the group' s existence, several hundred people joined and \textcolor{blue}{left}.\newline
• no profit was made in six years, and the church \textcolor{blue}{left}, losing their investment.\newline
• on 7 november he \textcolor{blue}{left}, missing the bolshevik revolution, which began on that day.\newline
• he had not re@-@ written his will and when produced still \textcolor{blue}{left} everything to his son lunalilo.\newline
• they continued filming as normal, and when lynch yelled cut, the townspeople had \textcolor{blue}{left}.\newline
• with land of black gold( 1950), a story that he had previously \textcolor{blue}{left} unfinished, instead.\newline
• he was infuriated that the government had \textcolor{blue}{left} thousands unemployed by closing down casinos and brothels.\newline
• an impending marriage between her and albert interfered with their studies, the two brothers \textcolor{blue}{left} on 28 august 1837 at the close of the term to travel around europe\newline
\newline \newline 
{\bf Transformer factor 33 in layer 4 \newline Explaination: light: noun, the natural agent that stimulates sight and makes things visible:  } \newline \newline • forced to visit the sarajevo television station at night and to film with as little \textcolor{blue}{light} as possible to avoid the attention of snipers and bombers.\newline
• by the modest, cream@-@ colored attire in the airy, \textcolor{blue}{light}@-@ filled clip.\newline
• the man asked her to help him carry the case to his car, a \textcolor{blue}{light}@-@ brown volkswagen beetle.\newline
• they are portrayed in a particularly sympathetic \textcolor{blue}{light} when they are killed during the ending.\newline
• caught up" was directed by mr. x, who was behind the laser \textcolor{blue}{light} treatment of usher' s 2004 video" yeah!"\newline
• piracy in the indian ocean, and the script depicted the pirates in a sympathetic \textcolor{blue}{light}.\newline
• without the benefit of moon \textcolor{blue}{light}, the light horsemen had fired at the flashes of the enemy' s\newline
• second innings, voce repeated the tactic late in the day, in fading \textcolor{blue}{light} against woodfull and bill brown.\newline
•, and the workers were transferred on 7 july to another facility belonging to early \textcolor{blue}{light}, 30 km away in< unk> town.\newline
• unk> brooklyn avenue ne near the university of washington campus in a small \textcolor{blue}{light}@-@ industrial building leased from the university.\newline
• factory where the incident took place is the< unk>(" early \textcolor{blue}{light}") toy factory(< unk>), owned by hong\newline
•, a 1934 comedy in which samuel was portrayed in an unflattering \textcolor{blue}{light}, and mrs beeton, a 1937 documentary,< unk>\newline
• stage effects and blue@-@ red \textcolor{blue}{light} transitions give the video a surreal feel, while a stoic crowd make\newline
• set against the backdrop of mumbai' s red@-@ \textcolor{blue}{light} districts, it follows the travails of its personnel and principal,\newline
• themselves on the chinese flank in the foothills, before scaling the position at first \textcolor{blue}{light}.\newline
\newline \newline 
{\bf Transformer factor 47 in layer 4 \newline Explaination: plants: noun, vegetation } \newline \newline • the distinct feature of the main campus is the mall, which is a large \textcolor{blue}{tree} – laden grassy area where many students go to relax.\newline
• each school in the london borough of hillingdon was invited to plant a \textcolor{blue}{tree}, and the station commander of raf northolt, group captain tim o\newline
• its diet in summer contains a high proportion of insects, while more \textcolor{blue}{plant} items are eaten in autumn.\newline
• large fruitings of the fungus are often associated with damage to the host \textcolor{blue}{tree}, such as that which occurs with burning.\newline
• she nests on the ground under the cover of \textcolor{blue}{plants} or in cavities such as hollow tree trunks.\newline
• orchards, heaths and hedgerows, especially where there are some old \textcolor{blue}{trees}.\newline
• the scent of \textcolor{blue}{plants} such as yarrow acts as an olfactory attractant to females.\newline
• of its grasshopper host, causing it to climb to the top of a \textcolor{blue}{plant} and cling to the stem as it dies.\newline
• well@-@ drained or sandy soil, often in the partial shade of \textcolor{blue}{trees}.\newline
• food is taken from the ground, low@-@ growing \textcolor{blue}{plants} and from inside grass tussocks; the crake may search leaf\newline
• into his thought that the power of gravity( which brought an apple from a \textcolor{blue}{tree} to the ground) was not limited to a certain distance from earth,\newline
• they eat both seeds and green \textcolor{blue}{plant} parts and consume a variety of animals, including insects, crustaceans\newline
• fyne, argyll in the 1870s was named as the uk ’ s tallest \textcolor{blue}{tree} in 2011.\newline
•", or colourless enamel, as in the ground areas, rocks and \textcolor{blue}{trees}.\newline
• produced from 16 to 139 weeks after a forest fire in areas with coniferous \textcolor{blue}{trees}.\newline
\newline \newline

\noindent {\bf This is the end of visualization of low-level transformer factor. Click [\ref{sec:hyper}] to go back. }

\newpage\phantom{}
\newpage

\subsection{Mid-Level Transformer Factors}
\label{sec:mid}

{\bf Transformer factor 13 in layer 10 \newline Explaination: Unit exchange with parentheses: e.g. 10 m (1000cm) } \newline \newline • 14@-@ 16 hand( 56 to 64 inches( 140 to 160 \textcolor{blue}{cm})) war horse is that it was a matter of pride to a\newline
•, behind many successful developments, defaulted on the\$ 214 million(\$ 47 \textcolor{blue}{billion}) in bonds held by 60@,@ 000 investors; the van\newline
• straus, behind many successful developments, defaulted on the\$ 214 million(\textcolor{blue}{\$} 47 billion) in bonds held by 60@,@ 000 investors;\newline
• the niche is 4 m( 13 \textcolor{blue}{ft}) wide and 3@.\newline
• with a top speed of nearly 21 knots( 39 km/ h; 24 \textcolor{blue}{mph}).\newline
•@ 4 billion( us\$ 21 \textcolor{blue}{million}) — india' s highest@-@ earning film of the year\newline
•) at deep load as built, with a length of 310 ft( 94 \textcolor{blue}{m}), a beam of 73 feet 7 inches( 22@.\newline
•@ 3@-@ inch( 160 \textcolor{blue}{mm}) calibre steel barrel.\newline
• and gave a maximum speed of 23 knots( 43 km/ h; 26 \textcolor{blue}{mph}).\newline
• 2 km) in length, with a depth around 790 yards( 720 \textcolor{blue}{m}), and in places only a few yards separated the two sides.\newline
• hull provided a combined thickness of between 24 and 28 inches( 60 – 70 \textcolor{blue}{cm}), increasing to around 48 inches( 1@.\newline
• switzerland, austria and germany; and his mother, lynette federer( \textcolor{blue}{born} durand), from kempton park, gauteng, is\newline
•@ 2 in( 361 \textcolor{blue}{mm}) thick sides.\newline
•) and a top speed of 30 knots( 56 km/ h; 35 \textcolor{blue}{mph}).\newline
•, an outdoor seating area( 4@,@ 300 square feet( 400 \textcolor{blue}{m2})) and a 2@,@ 500@-@ square@\newline
\newline \newline 
{\bf Transformer factor 24 in layer 10 \newline Explaination: Male name } \newline \newline • divorcing doqui in 1978, michelle married robert h. tucker, \textcolor{blue}{jr}. the following year, changed her name to gee tucker, moved back\newline
• divorced doqui in 1978 and married new orleans politician robert h. tucker, \textcolor{blue}{jr}. the following year; she changed her name to gee tucker and became\newline
• including isabel sanford, when chuck and new orleans politician robert h. tucker, \textcolor{blue}{jr}. visited michelle at her hotel.\newline
• of 32 basidiomycete mushrooms showed that mutinus elegan\textcolor{blue}{s} was the only species to show antibiotic( both antibacterial\newline
• amphicoelias, it is probably synonymous with camarasaurus grand\textcolor{blue}{is} rather than c. supremus because it was found lower in the\newline
•[ her] for warmth and virtue" and mehul s. thakka\textcolor{blue}{r} of the deccan chronicle wrote that she was successful in" deliver[\newline
• em( queen latifah) and uncle henry( david alan gr\textcolor{blue}{ier}) own a diner, to which dorothy works for room and board.\newline
• in melbourne on 10 august 1895, presented by dion boucicault, \textcolor{blue}{jr}. and robert brough, and the play was an immediate success.\newline
• in the early 1980s, james r. tindall, \textcolor{blue}{sr}. purchased the building, the construction of which his father had originally financed\newline
• in 1937, when chakravarthi rajagopalacha\textcolor{blue}{ri} became the chief minister of madras presidency, he introduced hindi as a compulsory\newline
• in 1905 william lewis moody, \textcolor{blue}{jr}. and isaac h. kempner, members of two of galveston'\newline
• also, walter b. jones, \textcolor{blue}{jr}. of north carolina sent a letter to the republican conference chairwoman cathy\newline
• empire' s leading generals, nikephoros bryennios the \textcolor{blue}{elder}, the doux of dyrrhachium in the western balkans\newline
• in bengali as< unk>: the warrior by raj chakrabor\textcolor{blue}{ty} with dev and mimi chakraborty portraying the lead roles.\newline
• on 1 june 1989, erik g. braath\textcolor{blue}{en}, son of bjørn g., took over as ceo\newline
\newline \newline 
{\bf Transformer factor 25 in layer 10 \newline Explaination: Attributive Clauses } \newline \newline • which allows japan to mount an assault on the us; or kill him, \textcolor{blue}{which} lets the us discover japan' s role in rigging american elections —\newline
• certain stages of development, and constitutive heterochromatin \textcolor{blue}{that} consists of chromosome structural components such as telomeres and centromeres\newline
• to the mouth of the nueces river, and oso bay, \textcolor{blue}{which} extends south to the mouth of oso creek.\newline
•@,@ 082 metric tons, and argentina, which ranks 17th, \textcolor{blue}{with} 326@,@ 900 metric tons.\newline
• of\$ 42@,@ 693 and females had a median income of\textcolor{blue}{\$} 34@,@ 795.\newline
• ultimately scored 14 points with 70 per cent shooting, and crispin, who \textcolor{blue}{scored} twelve points with 67 per cent shooting.\newline
• and is operated by danish air transport, and one jetstream 32, which \textcolor{blue}{seats} 19 and is operated by helitrans.\newline
• acute stage, which occurs shortly after an initial infection, and a chronic stage \textcolor{blue}{that} develops over many years.\newline
•, earl of warwick and then william of lancaster, and ada de warenne \textcolor{blue}{who} married henry, earl of huntingdon.\newline
• who ultimately scored 14 points with 70 per cent shooting, and crispin, \textcolor{blue}{who} scored twelve points with 67 per cent shooting.\newline
• in america, while" halo/ walking on sunshine" charted at number 4 \textcolor{blue}{in} ireland, 9 in the uk, 10 in australia, 28 in canada\newline
• five events, heptathlon consisting of seven events, and decathlon \textcolor{blue}{consisting} of ten< unk> every multi event, athletes participate in a\newline
•@-@ life of 154@,@ 000 years, and 235np \textcolor{blue}{with} a half@-@ life of 396@.\newline
• comfort, and intended to function as the prison, and the second floor was \textcolor{blue}{better} finished, with a hall and a chamber, and probably operated as the\newline
•b, which serves the quonset freeway, and exit 7a, \textcolor{blue}{which} serves route 402( frenchtown road), another spur route connecting the\newline
\newline \newline 
{\bf Transformer factor 42 in layer 10 \newline Explaination: Some kind of disaster, something unfortunate happened } \newline \newline • after the first five games, all losses, jeff carter suffered \textcolor{blue}{a} broken foot that kept him out of the line@-@ up for\newline
• allingham died of natural \textcolor{blue}{causes} in his sleep at 3: 10 am on 18 july 2009 at his\newline
• upon reaching corfu, thousands of serb troops began showing symptoms of ty\textcolor{blue}{phus} and had to be quarantined on the island of< un\newline
• than a year after the senate general election, the september 11, 2001 terrorist \textcolor{blue}{attacks} took place, with giuliani still mayor.\newline
• the starting job because fourth@-@ year junior grady was under suspension related \textcolor{blue}{to} driving while intoxicated charges.\newline
• his majesty, but as soon as they were on board ship, they died \textcolor{blue}{of} melancholy, having refused to eat or drink.\newline
• on 16 september 1918, before she had even gone into action, she suffered \textcolor{blue}{a} large fire in one of her 6@-@ inch magazines, and\newline
• orange goalkeeper for long@-@ time starter john galloway who was sick with \textcolor{blue}{the} flu.\newline
• in 1666 his andover home was destroyed by \textcolor{blue}{fire}, supposedly because of" the carelessness of the maid".\newline
• the government, on 8 february, admitted that the \textcolor{blue}{outbreak} may have been caused by semi@-@ processed turkey meat imported directly\newline
•ikromo came under investigation by the justice office of the dutch east indies \textcolor{blue}{for} publishing several further anti@-@ dutch editorials.\newline
• that he could attend to the duties of his office, but fell ill with \textcolor{blue}{a} fever in august 1823 and died in office on september 1.\newline
•@ 2 billion initiative to combat \textcolor{blue}{cholera} and the construction of a\$ 17 million teaching hospital in< unk\newline
• he would not hear from his daughter until she was convicted \textcolor{blue}{of} stealing from playwright george axelrod in 1968, by which time rosaleen\newline
• relatively hidden location and proximity to piccadilly circus, the street suffers from \textcolor{blue}{crime}, which has led to westminster city council gating off the man in\newline
\newline \newline 
{\bf Transformer factor 50 in layer 10 \newline Explaination: Doing something again, or making something new again } \newline \newline • 2007 saw the show undergo a \textcolor{blue}{rev}amp, which included a switch to recording in hdtv, the introduction\newline
• during the ship' s 1930 reconstruction; the maximum elevation of the main guns \textcolor{blue}{was} increased to+ 43 degrees, increasing their maximum range from 25@,\newline
• hurricane pack 1 was a \textcolor{blue}{rev}amped version of story mode; team ninja tweaked the\newline
• she was fitted with \textcolor{blue}{new} engines and more powerful water@-@ tube boilers rated at 6@\newline
• from 1988 to 2000, the two western towers were substantially \textcolor{blue}{overhaul}ed with a viewing platform provided at the top of the north tower.\newline
• latest missoula downtown master plan in 2009, increased emphasis was directed toward \textcolor{blue}{red}eveloping the north side' s former rail yard and the area\newline
• 1896: the ribbon of the army version medal of honor \textcolor{blue}{was} redesigned with all stripes being vertical.\newline
• the \textcolor{blue}{new} badge includes a star to represent the european cup win in 1982, and\newline
• missoula downtown master plan in 2009, increased emphasis was directed toward red\textcolor{blue}{eve}loping the north side' s former rail yard and the area just\newline
• also assisted in comprehensive infrastructure renovations, restored a dependable supply of electricity, \textcolor{blue}{rev}amped the baggage handling facilities as well as the arrival and departure lounge\newline
• hurricane pack 1 was a rev\textcolor{blue}{amp}ed version of story mode; team ninja tweaked the encounters\newline
• 1896: the ribbon of the army version medal of honor was \textcolor{blue}{redesigned} with all stripes being vertical.\newline
• from 1988 to 2000, the two western towers were \textcolor{blue}{substantially} overhauled with a viewing platform provided at the top of the north tower\newline
• assisted in comprehensive infrastructure renovations, restored a dependable supply of electricity, rev\textcolor{blue}{amp}ed the baggage handling facilities as well as the arrival and departure lounges\newline
• bond series and the fourth to star roger moore as bond; the plot was \textcolor{blue}{significantly} changed from the novel to include excursions into space.\newline
\newline \newline 
{\bf Transformer factor 51 in layer 10 \newline Explaination: apostrophe s, possesive } \newline \newline • the irish times was critical of the book\textcolor{blue}{'} s text but wrote positively of the included photographs.\newline
• if it \textcolor{blue}{survived} long enough to become old@-@ fashioned it was likely to be\newline
• you by phil spector as his inspirations, which resulted to the album\textcolor{blue}{'} s wall of sound resonance.\newline
• the irish times was critical of the book' \textcolor{blue}{s} text but wrote positively of the included photographs.\newline
• album to the wu tang clan and nine inch nails, particularly comparing the album\textcolor{blue}{'} s production( which was done by various producers with executive producer don gilmore\newline
• to the wu tang clan and nine inch nails, particularly comparing the album' \textcolor{blue}{s} production( which was done by various producers with executive producer don gilmore)\newline
• toward the commoners and interested in easing their burden but suspicious about the letter\textcolor{blue}{'} s true purpose, reluctantly signed the document under intense pressure from the french\newline
• the novel\textcolor{blue}{'} s reception was even warmer than that of its predecessor; waugh was\newline
• first song selected for inclusion after her mother' s recommendation and the song' \textcolor{blue}{s} melancholic lyrics.\newline
• it divided critics at the time; although they praised the game\textcolor{blue}{'} s writing and scale of choice, they criticized its technical flaws.\newline
• mgm executive al lewin said that several years after the film' \textcolor{blue}{s} release stroheim asked him for the cut footage.\newline
• the game' \textcolor{blue}{s} production was turbulent, as the design' s scope exceeded the available resources\newline
• nicki escudero from the phoenix new times noted the song\textcolor{blue}{'} s superficial themes which included lyrics about" sex, money and cheating"\newline
• mgm executive al lewin said that several years after the film\textcolor{blue}{'} s release stroheim asked him for the cut footage.\newline
• labrie said that there was" a lot of discussion" about the song\textcolor{blue}{'} s wording and how direct it should be.\newline
\newline \newline 
{\bf Transformer factor 86 in layer 10 \newline Explaination: Pattern: Consecutive years, this is convention to name foodball/rugby game season } \newline \newline • with york the previous season, signed a contract until the end of 2013 – \textcolor{blue}{14} and sheffield united midfielder elliott whitehouse signed on a one@-@\newline
• as of the end of the 2014 – \textcolor{blue}{15} season, aston villa have spent 104 seasons in the top tier of english\newline
• won 13 and drew two of their opening 15 league matches of the 1985 – \textcolor{blue}{86} campaign, and seemed destined to win the first division title.\newline
• mcallister, still without a goal in 2009 – \textcolor{blue}{10}, couldn' t get on the scoresheet in the three games\newline
• john bentley led united to a fourth@-@ place finish in 1912 – \textcolor{blue}{13}.\newline
• he made 46 appearances, scoring three goals, in the 2001 – \textcolor{blue}{02} season before spending the close season with the kalamazoo kingdom in the\newline
• he moved to basingstoke town towards the end of 2001 – \textcolor{blue}{02}, making his debut in march 2002.\newline
• 7[ note 1] was the worst record in the nhl for 2011 – \textcolor{blue}{12} and the first time in franchise history they finished in last place.\newline
• side, who withdrew from the football league at the end of the 1893 – \textcolor{blue}{94} season after finishing bottom of the second division.\newline
• spent a year as a physics instructor at the university of minnesota in 1916 – \textcolor{blue}{17}, then two years as a research engineer with the westinghouse lamp\newline
• defeat was a 7 – 2 loss to witton albion in the 2001 – \textcolor{blue}{02} season.\newline
• york achieved three successive wins for the first time in 2013 – \textcolor{blue}{14} after beating northampton 2 – 0 away, with bowman and fletcher scoring in\newline
• he started to develop more of an offensive game, finishing off the 2001 – \textcolor{blue}{02} season with 58 points in the 47 games he played in seattle.\newline
• suart limited matthews to 19 league appearances in 1958 – \textcolor{blue}{59}.\newline
• jaw warriors of the western hockey league( whl) during the 2000 – \textcolor{blue}{01} season.\newline
\newline \newline 
{\bf Transformer factor 99 in layer 10 \newline Explaination: past tense } \newline \newline • r. in their review of rihanna' s top 20 songs, time out \textcolor{blue}{ranked}" man down" as their tenth best track, writing that it is\newline
• rolling stone \textcolor{blue}{ranked}" imagine" number three on its list of" the 500 greatest songs\newline
• japan' s computer entertainment rating organization( cero) \textcolor{blue}{rated} ninja gaiden and black, on their release, as 18+\newline
• ultimate classic rock \textcolor{blue}{ranked}" lola" as the kinks' third best song, saying"\newline
• adrien begrand of popmatters \textcolor{blue}{described}" south of heaven" as" an unorthodox set opener\newline
• columbia records \textcolor{blue}{released} it as the album' s fourth and final single on june 14,\newline
• rolling stone \textcolor{blue}{ranked} it the best song of 2009 and the 36th@-@ best song\newline
• indielondon' s jack foley \textcolor{blue}{noted}" wind it up" as a highlight of the sweet escape and called\newline
• premiere magazine \textcolor{blue}{listed} frank booth, played by dennis hopper, as the fifty@-@\newline
• columbia records \textcolor{blue}{released}" crazy in love" on may 18, 2003, as the lead\newline
• the times \textcolor{blue}{considered} the production the best since the original, and praised it for its fidelity\newline
• the good food guide \textcolor{blue}{ranked} hibiscus as the eighth@-@ best restaurant in the uk\newline
• viz media later began \textcolor{blue}{releasing} the manga as simply" ral grad" in february 2008.\newline
• entertainment weekly magazine \textcolor{blue}{ranked}" crazy in love" forty@-@ seven in its list of\newline
• the japanese publisher nihon bungeisha \textcolor{blue}{released} the series in collected volumes from january 2000 to september 2009.\newline
\newline \newline 
{\bf Transformer factor 102 in layer 10 \newline Explaination: African name } \newline \newline • s 1966 to 1971 live performances in paris, prepared to press the album once \textcolor{blue}{mw}anga provided the label with the record< unk>.\newline
• of america" with the nhk symphony orchestra, but cancelled both deals upon \textcolor{blue}{mw}anga' s return from japan.\newline
• 1966 to 1971 live performances in paris, prepared to press the album once mw\textcolor{blue}{anga} provided the label with the record< unk>.\newline
• and langston hughes, and by modern african poets and folk artists such as \textcolor{blue}{kw}esi brew and efua sutherland, which also influenced her auto\newline
• america" with the nhk symphony orchestra, but cancelled both deals upon mw\textcolor{blue}{anga}' s return from japan.\newline
• du bois was buried in \textcolor{blue}{acc}ra near his home, which is now the du bois memorial centre.\newline
• du bois returned to africa in late 1960 to attend the inauguration of \textcolor{blue}{n}namdi azikiwe as the first african governor of nigeria.\newline
• david mcgurk, lanre o\textcolor{blue}{ye}banjo, danny parslow, tom platt and chris smith signed new\newline
• and moderate nationalist parties, the most prominent of which was bishop abel muzo\textcolor{blue}{rew}a' s united african national council( uanc).\newline
• a few weeks after of human feelings was recorded, \textcolor{blue}{mw}anga went to japan to negotiate a deal with trio records to have the\newline
• and was part of two large campaigns, one to witu and another to \textcolor{blue}{mw}ele.\newline
• returned to africa in late 1960 to attend the inauguration of nnamdi az\textcolor{blue}{iki}we as the first african governor of nigeria.\newline
• in april, \textcolor{blue}{mw}anga arranged another session at cbs studios in new york city, and coleman\newline
• the government and moderate nationalist parties, the most prominent of which was bishop abel \textcolor{blue}{mu}zorewa' s united african national council( uanc).\newline
• ralambo' s father, and\textcolor{blue}{riam}anelo, had established rules of succession by which ralambo'\newline
\newline \newline 
{\bf Transformer factor 125 in layer 10 \newline Explaination: Describing someone in a paraphrasing style. Name, Career } \newline \newline • journalist tim \textcolor{blue}{judah} suggests that the move may have been motivated by a desire to control a\newline
• the historian \textcolor{blue}{nora} berend says that the latter measure" may have adversely affected\newline
• from the pyx that were not assayed, and numismatic \textcolor{blue}{historian} roger burdette speculates that ashbrook, generally well@-\newline
• the pyx that were not assayed, and numismatic historian \textcolor{blue}{roger} burdette speculates that ashbrook, generally well@-@\newline
• the cricket historian \textcolor{blue}{derek} birley notes that many of these bowlers imitated the methods of\newline
• critic \textcolor{blue}{roberta} reeder notes that the early poems always attracted large numbers of admirers\newline
•; the figures for the last two years are not available, but sf historian \textcolor{blue}{mike} ashley estimates that fantastic paid circulation may have been as low as 13@\newline
• aesthetically, ign' s \textcolor{blue}{tal} blevins noted that the game had" a very distinct 40s\newline
• sf historian \textcolor{blue}{everett} bleiler notes that hersey did not mention the venture in his\newline
• similarly, duke university professor, \textcolor{blue}{mark} anthony neal, writes, “ nas was at the forefront of a renaissance\newline
• club reviewer \textcolor{blue}{erik} adams wrote that the episode was a perfect mix, between the more subtle\newline
• commenting on the album and its use of samples, pitchfork' s \textcolor{blue}{jeff} weiss claims that both nas and his producers found inspiration for the album'\newline
• the \textcolor{blue}{historian} stanley karnow said of ky and thi:" both fl\newline
• that were not assayed, and numismatic historian roger burde\textcolor{blue}{tte} speculates that ashbrook, generally well@-@ treated by the\newline
• irataba was described as an eloquent speaker, and linguist lean\textcolor{blue}{ne} hinton suggests that he was among the first mohave people to\newline
\newline \newline 
{\bf Transformer factor 134 in layer 10 \newline Explaination: Transition sentence  } \newline \newline • fanny workman have tended to slight or belittle her achievements, but \textcolor{blue}{contemporaries}, unaware of the far greater accomplishments to come, held the workmans\newline
• scheduled to air in its regular half@-@ hour time slot, but \textcolor{blue}{nbc} later announced it would be expanded to fill an hour time slot beginning a\newline
• wine and savoy cabbage with a red wine and smoked chocolate sauce, but he \textcolor{blue}{otherwise} felt that the food was" over@-@ worked" and the\newline
• lap melee when he was hit by romain grosjean; webber was \textcolor{blue}{forced} to pit straight away, while grosjean was given a ten@\newline
• ra. one was initially scheduled to release on 3 june 2011, but \textcolor{blue}{delays} due to a lengthy post@-@ production process and escalating\newline
• yamina nomads who were centered at tuttul, and the \textcolor{blue}{rebels} were supported by yamhad' s king sumu@-@\newline
• the item was intended simply as a piece of news, but \textcolor{blue}{telegraph} lines quickly spread the news throughout the state, fueling procession sentiment.\newline
• both twc and comcast began trials of services based on the system; \textcolor{blue}{turner} broadcasting was an early supporter of the system, providing access to tbs and\newline
•k> have claimed that he proposed a dictatorship for robespierre, but \textcolor{blue}{nonetheless} some of them considered him to be redeemable, or at least\newline
•@ 2 style with superfiring pairs of turrets fore and aft; the \textcolor{blue}{middle} turrets were not superfiring, and had a funnel between them.\newline
• romani being ordered to move out with supplies for the advancing troops, but 150 \textcolor{blue}{men}, most of whom were past the end of their contracts and entitled to\newline
•' s boats to enter the creek into which the schooner had fled, the \textcolor{blue}{small} craft entering the waterway in the hope of storming and capturing the vessel\newline
•-@ person shooter elements and a unique on rails control scheme, but the \textcolor{blue}{core} adventure@-@ style gameplay has been compared to myst and snatch\newline
• stanza 6; movement 4 incorporates ideas from stanzas 7 – 14, and \textcolor{blue}{movement} 5 relies on stanzas 15 and< unk> movement 2,\newline
• as corps troops that were usually allocated at a rate of one per division; \textcolor{blue}{several} of the militia units were also later designated australian imperial force units, after\newline
\newline \newline 
{\bf Transformer factor 152 in layer 10 \newline Explaination: in some locations } \newline \newline • while most breeding stallions and racehorses of the era had \textcolor{blue}{stable} companions, waxy reportedly was fond of rabbits in his later years and\newline
• planet and the helter skelter music bookshop have also been based \textcolor{blue}{on} the street.\newline
• the central bank of somalia, the national monetary authority, also has its headquarters \textcolor{blue}{in} mogadishu.\newline
• some allotropes of the other actinides also exhibit \textcolor{blue}{similar} behaviour, though to a lesser degree.\newline
•; fortune 1000 technology company< unk>, for instance, is headquartered \textcolor{blue}{in} the area.\newline
• musical@-@ comedy television series maid marian and her merry men were filmed \textcolor{blue}{in} cleeve abbey.\newline
• ireland,< unk> and donegal bay in particular, have popular surfing \textcolor{blue}{beaches}, being fully exposed to the atlantic ocean.\newline
•lstoy' s war and peace and chekhov' s peasants both feature \textcolor{blue}{scenes} in which wolves are hunted with hounds and< unk>.\newline
• while most breeding stallions and racehorses of the era had stable \textcolor{blue}{companions}, waxy reportedly was fond of rabbits in his later years and"\newline
• the lancashire and england test cricketer paul allott was born \textcolor{blue}{in} altrincham.\newline
•asura, the demon devotee of shiva, are both credited with building \textcolor{blue}{temples} or cut caves to live.\newline
• forbidden planet and the helter skelter music bookshop have also been \textcolor{blue}{based} on the street.\newline
•thopedic shriners hospitals in the u. s. is also located \textcolor{blue}{in} spokane.\newline
• dykes to watch out for and fun home, was born in lock haven \textcolor{blue}{in} 1960.\newline
•< unk>, and alessandra ambrosio have each worn \textcolor{blue}{two} fantasy bras.\newline
\newline \newline

\noindent {\bf This is the end of visualization of mid level transformer factor. Click [\ref{sec:hyper}] to go back. }
\newpage\phantom{}
\newpage

\subsection{High-Level Transformer Factors}
\label{sec:high}

 {\bf Transformer factor 297 in layer 10 with saliency map \newline Explaination: repetitive structure detector }  \newline \newline 
 \colorbox{red!56}{frontier} works, and an original soundtrack by avex group were created based on the game. drama \colorbox{green!95}{cd}: tales of grace\colorbox{green!100}{s} 1 to 4 are side stories that take place during the game' s plot. they were released between may 26, 2010 and august 25, 2010\colorbox{red!42}{.} anthology drama cd: \colorbox{red!32}{tales} of graces f \colorbox{red!19}{2010} winter\colorbox{red!25}{,} \colorbox{red!23}{anthology} drama cd\colorbox{red!15}{:} \colorbox{red!44}{tales} of graces f \colorbox{red!25}{2011} summer, anthology drama cd\colorbox{red!16}{:} \colorbox{red!29}{tales} of graces f 2012 winter, anthology drama cd\colorbox{red!21}{:} \colorbox{red!19}{tales} of graces f 2012 summer\colorbox{red!41}{,} \colorbox{red!35}{anthology} \colorbox{red!42}{drama} \colorbox{red!65}{cd}\colorbox{red!77}{:} \colorbox{red!100}{\textcolor{blue}{tales}} \colorbox{red!95}{of} \colorbox{red!100}{grace}\colorbox{red!58}{s} \colorbox{red!30}{f} \colorbox{red!45}{2013} \colorbox{red!33}{win}\colorbox{red!22}{te} r, and anthology drama \colorbox{red!25}{cd}: tales of graces \colorbox{red!22}{f} 2013\\\\
 \colorbox{red!28}{cobra}nd platinum cardholders, and citibank eva air \colorbox{green!77}{cobra}\colorbox{green!100}{nd} world card) the infinity( infinity mileagelands diamond, royal laurel\colorbox{red!17}{/} premium laurel class \colorbox{red!16}{passengers}\colorbox{red!27}{,} \colorbox{red!42}{star} \colorbox{red!66}{alliance} \colorbox{red!39}{first}/ \colorbox{red!17}{business} class passengers, american \colorbox{red!14}{express} centurion/ eva air cobrand platinum cardholders\colorbox{red!14}{,} and ci\colorbox{red!13}{ti}\colorbox{red!15}{bank} eva air cobrand world card\colorbox{red!13}{holders}) the star( infinity mileagelands diamond/ gold, royal laurel/ premium \colorbox{red!16}{laurel} class \colorbox{red!37}{passengers}\colorbox{red!89}{,} \colorbox{red!23}{\textcolor{blue}{star}} \colorbox{red!100}{alliance} \colorbox{red!21}{first}\colorbox{red!15}{/} \colorbox{red!12}{business} class \colorbox{red!18}{passengers}, \colorbox{red!33}{star} \colorbox{red!19}{alliance} gold members\colorbox{red!19}{,} american \colorbox{red!16}{express} centurion/ eva air cobrand platinum cardholders\colorbox{red!14}{,} citibank eva air cobrand world cardholders, business customers\colorbox{red!14}{,}\\\\
 \colorbox{red!28}{quickly} set online< unk> alight"\colorbox{red!17}{.}" \colorbox{red!54}{can}' \colorbox{red!28}{t} \colorbox{red!55}{get} \colorbox{red!32}{you} \colorbox{red!43}{out} of \colorbox{red!19}{my} \colorbox{red!41}{head}" was chosen as the lead single from minogue' \colorbox{red!11}{s} eighth studio album fever, and it was released on 8 september 2001 by \colorbox{red!9}{par}lophone in australia, \colorbox{green!100}{while} \colorbox{red!21}{in} the united kingdom \colorbox{green!84}{and} other european countries it was released on 17 september\colorbox{red!12}{.}\colorbox{red!20}{"} \colorbox{red!100}{can}\colorbox{red!39}{'} \colorbox{red!55}{t} \colorbox{red!57}{\textcolor{blue}{get}} \colorbox{red!38}{you} \colorbox{red!38}{out} \colorbox{red!19}{of} \colorbox{red!19}{my} \colorbox{red!39}{head}" was w ritten and produced by \colorbox{red!15}{cathy} dennis \colorbox{red!9}{and} rob davis, who had been \colorbox{red!14}{put} together by british artist manager simon fuller\colorbox{red!17}{,} who wanted the duo to come up with \colorbox{red!13}{a} song for british pop group s club 7. the song was recorded using cuba\\\\
 \colorbox{red!61}{typhoon} status with two\colorbox{red!17}{@}\colorbox{red!31}{-}\colorbox{red!19}{@} minute sustained winds estimated at 125 km/ h( 78 mph). around 1700 utc \colorbox{red!18}{on} may 31, the storm tracked approximately 65 km( 40 mi) west of iwo \colorbox{red!34}{jim}a\colorbox{red!22}{.} roughly five hours later\colorbox{red!17}{,} it \colorbox{red!17}{moved} within 15 km( 10 mi) of \colorbox{red!22}{chi}\colorbox{red!23}{chi}\colorbox{red!18}{@}\colorbox{red!24}{-}\colorbox{red!30}{@} \colorbox{red!51}{jim}a where \colorbox{red!25}{a} \colorbox{red!12}{pressure} of 992 mb( hpa; 29@.@ 30 inhg) \colorbox{red!21}{was} measured\colorbox{red!36}{.} \colorbox{red!21}{sustained} \colorbox{red!35}{winds} \colorbox{red!28}{on} \colorbox{red!20}{chi}\colorbox{red!24}{chi}\colorbox{red!37}{@}\colorbox{red!100}{-}\colorbox{green!0}{\textcolor{blue}{@}} \colorbox{red!61}{jim}\colorbox{red!95}{a} reached 95 km/ h( \colorbox{red!19}{60} mph); \colorbox{red!21}{however}, these were determined to be unrepresentative of lucille' s actual intensity due\\\\
 \colorbox{red!43}{first} book \colorbox{red!13}{in} vocal music. the \colorbox{red!25}{modern} \colorbox{red!18}{music} series. book 1. new york, new york: silver burdette and company. smith, eleanor( 1901). a second book in vocal music. \colorbox{red!13}{the} \colorbox{red!29}{modern} \colorbox{red!23}{music} series. book 2. new york, new york: silver burdette and company. smith, eleanor( \colorbox{green!58}{1901}\colorbox{green!81}{)}. a third book in vocal music. \colorbox{red!15}{the} \colorbox{red!41}{modern} music \colorbox{red!19}{series}. book 3. \colorbox{green!74}{new} york, new york: silver burdette and company. smith, \colorbox{red!19}{eleanor}( 1905)\colorbox{red!17}{.} a \colorbox{red!20}{fourth} \colorbox{red!12}{book} \colorbox{red!17}{in} vocal \colorbox{red!17}{music}\colorbox{red!22}{.} \colorbox{red!100}{the} \colorbox{red!53}{\textcolor{blue}{modern}} \colorbox{red!85}{music} \colorbox{red!66}{series}\colorbox{red!14}{.} \colorbox{red!17}{book} \colorbox{red!13}{4}\colorbox{red!13}{.} new york\colorbox{red!11}{,} new york: \colorbox{green!100}{silver} burde\\\\
\colorbox{red!41}{@} breaking eight weeks at number one on the airplay chart of \colorbox{green!78}{the} country and \colorbox{green!100}{became} the first to \colorbox{red!9}{garner} 3000 radio plays in a single week. subsequently\colorbox{red!13}{,} it became \colorbox{red!9}{the} most@-@ played song of 2001 in the region\colorbox{red!13}{.}\colorbox{red!24}{"} \colorbox{red!56}{can}\colorbox{red!25}{'} \colorbox{red!22}{t} \colorbox{red!55}{get} \colorbox{red!29}{you} \colorbox{red!43}{out} of \colorbox{red!28}{my} \colorbox{red!29}{head}" was certified platinum \colorbox{red!15}{by} the british phonographic industry for shipments of 600@,@ 000 units in 2001. the certification \colorbox{red!14}{was} upgraded to double@-@ platinum in 2015, denoting shipments \colorbox{green!77}{of} 1@,@ 200@,@ 000 units. in the united states,\colorbox{red!18}{"} \colorbox{red!100}{can}\colorbox{red!44}{'} \colorbox{red!45}{t} \colorbox{red!34}{\textcolor{blue}{get}} \colorbox{red!33}{you} \colorbox{red!48}{out} \colorbox{red!40}{of} \colorbox{red!33}{my} \colorbox{red!32}{head}\colorbox{red!21}{"} peaked at number seven on the\\\\
 chart. in mid@-@ august 2015,\colorbox{red!17}{"} \colorbox{red!46}{la} \colorbox{red!55}{mor}\colorbox{red!48}{di}\colorbox{red!33}{dit}a" earned martin his twenty@-@ sixth top ten hit on hot latin songs. he became the fourth artist \colorbox{red!13}{with} the \colorbox{green!56}{most} top tens in \colorbox{green!100}{the} 29@-@ year history of the chart\colorbox{red!21}{.} in late august 2015\colorbox{red!17}{,} \colorbox{red!21}{martin} \colorbox{red!28}{earned} \colorbox{red!18}{with}\colorbox{red!36}{"} \colorbox{red!69}{la} \colorbox{red!45}{\textcolor{blue}{mor}}\colorbox{red!100}{di}\colorbox{red!47}{dit}\colorbox{red!53}{a}\colorbox{red!30}{"} \colorbox{red!21}{his} \colorbox{red!15}{fifteenth} number@-@ one on the latin airplay chart\colorbox{red!16}{(} up 58 percent, to 11@.@ 8 million audience impressions)\colorbox{green!86}{.} \colorbox{green!63}{eventually}," \colorbox{red!42}{la} \colorbox{red!57}{mor}\colorbox{red!50}{di}\colorbox{red!36}{dit}a" \colorbox{red!18}{peaked} at number six on the us hot latin songs chart, number one on latin airpla y and\\\\
\colorbox{red!20}{,} was delivered to sukhoi' s experimental workshop to be outfitted with exclusive systems. built by knaapo, its structure has increased carbon@-@ fibre and \colorbox{red!30}{al}@-@ li content. installed was \colorbox{red!17}{the} 2d thrust@-@ \colorbox{red!19}{vector}ing \colorbox{red!25}{l}\colorbox{red!25}{yu}\colorbox{red!42}{lka} \colorbox{red!37}{al}\colorbox{red!14}{@}-@ 31fp\colorbox{red!20}{,} an \colorbox{red!27}{interim} measure pending the availability of the \colorbox{red!38}{al}@-@ 37fu(< unk>< unk>,\colorbox{red!23}{"} after\colorbox{red!28}{burn}er@-@ controlled"\colorbox{red!19}{)}. \colorbox{red!23}{the} 3d thrust@\colorbox{red!17}{-}@ vector\colorbox{red!28}{ing} \colorbox{red!25}{l}\colorbox{red!25}{yu}\colorbox{red!52}{lka} \colorbox{red!22}{\textcolor{blue}{al}}\colorbox{red!100}{@}\colorbox{red!23}{-}@ \colorbox{red!21}{37}fu \colorbox{red!23}{was} still \colorbox{red!17}{in} \colorbox{red!24}{development}. the \colorbox{red!19}{al}@-@ 31fp\colorbox{red!17}{,} in\\\\
\colorbox{red!15}{} ke' s former band, though \colorbox{green!78}{escape} the fate only charted at number 25, seven spots lower than \colorbox{red!18}{the} \colorbox{red!57}{drug} \colorbox{red!21}{in} \colorbox{red!29}{me} is you\colorbox{red!13}{,} despite equal sales\colorbox{red!22}{.} in \colorbox{red!19}{its} \colorbox{red!19}{second} week on sales\colorbox{red!36}{,} \colorbox{red!100}{the} \colorbox{red!49}{\textcolor{blue}{drug}} \colorbox{red!80}{in} \colorbox{red!77}{me} \colorbox{red!22}{is} you \colorbox{red!15}{dropped} about 70\% in the united states\colorbox{red!11}{,} selling \colorbox{green!100}{5}\colorbox{red!18}{@},\colorbox{green!78}{@} 870 copies. this dropped the album 60 spots to number 79 on the billboard 200, and brought total us sales for the album to around 24\colorbox{green!86}{@},@ 000 copies. on the billboard charts, the \colorbox{red!45}{drug} \colorbox{red!22}{in} \colorbox{red!29}{me} \colorbox{red!16}{is} you charted at number two \colorbox{red!17}{on} the \colorbox{red!13}{top} hard rock albums chart, number \colorbox{green!49}{three} on \colorbox{red!12}{the} top alternative albums and top rock albums charts,\\\\
 \colorbox{red!13}{no}, \colorbox{red!34}{no}\colorbox{red!26}{,} \colorbox{red!18}{no}", \colorbox{green!86}{reached} number one on the billboard \colorbox{red!14}{hot} \colorbox{green!100}{r}\& b/ hip@-@ hop singles\& tracks and number three on the billboard hot 100. its follow@-@ up \colorbox{red!20}{single}," with me \colorbox{red!12}{part} 1" failed to reproduce \colorbox{red!20}{the} \colorbox{red!19}{success} \colorbox{red!33}{of}\colorbox{red!90}{"} \colorbox{red!89}{no}\colorbox{red!79}{\textcolor{blue}{,}} \colorbox{red!66}{no}\colorbox{red!100}{,} \colorbox{red!19}{no}\colorbox{red!56}{"}. \colorbox{red!15}{meanwhile}\colorbox{red!24}{,} the group featured on a song from \colorbox{red!15}{the} soundtrack album of the romantic drama why do fools fall in love and" get on the \colorbox{red!20}{bus}" had a limited release in europe and other markets\colorbox{red!14}{.} in 1998, destiny' s child garnered three soul train lady of soul awards including best new artist \colorbox{red!18}{for}\colorbox{red!25}{"} \colorbox{red!18}{no}\colorbox{red!15}{,} \colorbox{red!25}{no}, \colorbox{red!16}{no}\\\\
\colorbox{red!41}{} oistic warm\colorbox{green!42}{ong}ers. alexander \colorbox{red!12}{k}rivenko( \colorbox{green!71}{jonathan} adams) \colorbox{green!55}{finally}, introduced in trivial games and paranoid pursuits, is russian alexander \colorbox{red!7}{k}\colorbox{red!8}{ri}venko, the commander of the moonbase where the ispf have their headquarters. a winner of the nobel prize for medicine, it is \colorbox{red!10}{k}ri\colorbox{green!100}{ven}\colorbox{green!62}{ko}' s research into bone damage that has contributed to enabling humanity to access space easily. although the star cops are independent, \colorbox{red!4}{spring}' s \colorbox{red!9}{relationship} with \colorbox{red!13}{k}rivenko is often def\colorbox{red!6}{ere}ntial and he \colorbox{red!5}{frequently} \colorbox{red!14}{seems} \colorbox{red!7}{to} cap\colorbox{red!10}{it}\colorbox{red!23}{ulate} \colorbox{red!33}{to} \colorbox{red!42}{\textcolor{blue}{k}}\colorbox{red!100}{ri}\colorbox{red!43}{ven}\colorbox{red!16}{ko}\colorbox{red!10}{'} \colorbox{red!5}{s} \colorbox{red!10}{wishes}\colorbox{red!18}{.}== production history===\colorbox{red!7}{=}\colorbox{red!5}{=} origins=\\\\
 \colorbox{green!100}{that} build faith: from \colorbox{red!16}{the} life and ministry of thomas s. monson, \colorbox{green!58}{salt} lake city\colorbox{red!16}{,} utah\colorbox{red!18}{:} \colorbox{red!57}{des}\colorbox{red!46}{ere}\colorbox{red!27}{t} \colorbox{red!23}{book}, isbn 978\colorbox{red!17}{@}-@ \colorbox{green!64}{0}@-@ 87579@-@ 901@-@ 8 — \colorbox{green!91}{—}( 1996), faith rewarded: a personal account of prophetic promises to the east german saints, salt lake city, \colorbox{red!23}{utah}: \colorbox{red!50}{des}\colorbox{red!50}{ere}\colorbox{red!25}{t} \colorbox{red!19}{book}, isbn 978\colorbox{green!55}{@}-@ 1@-@ 57345@-@ 186@-@ \colorbox{green!38}{4} — —( 1997\colorbox{red!25}{)}\colorbox{red!25}{,} invitation to exal\colorbox{red!16}{tation}\colorbox{red!26}{,} salt lake city\colorbox{red!68}{,} \colorbox{red!39}{utah}\colorbox{red!75}{:} \colorbox{red!32}{\textcolor{blue}{des}}\colorbox{red!100}{ere}\colorbox{red!89}{t} \colorbox{red!42}{book}, isbn 978@-\\\\
\colorbox{red!31}{} \colorbox{red!17}{dell}, tom( 2015). gunnerkrigg court volume 5:< unk>\colorbox{green!99}{.} gunnerkrigg court. arch\colorbox{green!53}{aia} studios press. isbn 978@-@< unk\colorbox{red!12}{>}.=== side comics=== siddell, tom( 2013\colorbox{green!60}{)}. annie in the forest part \colorbox{green!41}{one}. beyond the \colorbox{red!13}{walls}\colorbox{red!14}{.} \colorbox{red!39}{robot} \colorbox{red!39}{voice} \colorbox{red!27}{comics}. siddell, tom\colorbox{green!59}{(} 2013). annie in the \colorbox{green!100}{forest} part two. beyond the walls. \colorbox{red!52}{robot} \colorbox{red!38}{voice} \colorbox{red!24}{comics}\colorbox{red!19}{.} siddell, tom( 2015). \colorbox{red!21}{traveller}\colorbox{red!22}{.} \colorbox{red!16}{beyond} the \colorbox{red!31}{walls}\colorbox{red!16}{.} \colorbox{red!17}{\textcolor{blue}{robot}} \colorbox{red!100}{voice} \colorbox{red!37}{comics}\colorbox{red!33}{.}=\colorbox{red!14}{=}\colorbox{red!13}{=} ex\colorbox{red!15}{pl}anatory footnotes======\\\\
 \colorbox{red!75}{95}@.@ 4 kn\colorbox{red!19}{)} f\colorbox{red!18}{40}2@-@ \colorbox{red!60}{rr}@\colorbox{red!15}{-}@\colorbox{green!83}{<} unk> engine, while later examples were fitted with the 23@,@ 000 lbf( 105@.@ 8 kn\colorbox{red!57}{)} f\colorbox{red!31}{40}\colorbox{red!27}{2}\colorbox{red!49}{@}\colorbox{red!100}{-}\colorbox{green!0}{\textcolor{blue}{@}} \colorbox{red!41}{rr}\colorbox{red!26}{@}-\colorbox{red!24}{@} \colorbox{red!32}{40}\colorbox{red!20}{8}\colorbox{red!20}{a}\colorbox{red!30}{.} in \colorbox{red!15}{the} early 2000s\colorbox{red!28}{,} 17 tav@-@ \colorbox{red!13}{8}bs were upgraded to include \colorbox{red!14}{a} night@-@ attack capability\colorbox{red!18}{,} \colorbox{red!17}{the} \colorbox{red!21}{f}402@-@ \colorbox{red!32}{rr}@\colorbox{red!22}{-}@ 408 \colorbox{red!19}{engine}, and software and structural changes.< unk> in 1991\colorbox{red!17}{,} the night attack harrier was the first upgrade of the \colorbox{green!100}{av}@-@ 8\\\\
\colorbox{red!26}{,} aitrus' meeting with ti' ana, and the birth of their son gehn\colorbox{red!12}{.} the book also explains the destruction of \colorbox{red!19}{the} \colorbox{red!31}{d}\colorbox{red!24}{'} \colorbox{red!39}{ni} civilization. \colorbox{red!13}{two} \colorbox{red!28}{d}\colorbox{red!19}{'} \colorbox{red!17}{ni}, veovis \colorbox{red!14}{and} a' gaeris, plot to destroy \colorbox{red!12}{their} civilization, which they \colorbox{green!47}{believe} has been \colorbox{red!12}{corrupted}. veov\colorbox{red!15}{is} \colorbox{red!13}{and} a' gaeris create a plague which wipes out \colorbox{green!100}{many} \colorbox{red!35}{of} \colorbox{red!55}{the} \colorbox{red!88}{d}\colorbox{red!47}{\textcolor{blue}{'}} \colorbox{red!100}{ni} \colorbox{red!21}{and} \colorbox{red!20}{follows} \colorbox{red!15}{them} \colorbox{red!14}{through} the ages\colorbox{red!13}{.} veovis is murdered by \colorbox{green!32}{a}' gaeris \colorbox{red!10}{for} refusing to \colorbox{red!10}{write} an age where the two of them would have been worshipped as gods, and aitrus sacrifices himself in order to\\\\
\colorbox{red!22}{} inants"\colorbox{green!77}{.} a gb\colorbox{red!27}{rm}\colorbox{red!14}{pa} \colorbox{green!57}{briefing} stated the \colorbox{green!100}{company} had" threatened \colorbox{red!5}{a} compensation claim of\$< unk> \colorbox{red!15}{should} \colorbox{red!35}{the} \colorbox{red!39}{gb}\colorbox{red!55}{\textcolor{blue}{rm}}\colorbox{red!100}{pa} \colorbox{red!11}{intend} to \colorbox{red!8}{ex}ert authority over the company' s operations". in response \colorbox{red!6}{to} the< unk> of the dumping incidents, the \colorbox{red!9}{gb}\colorbox{red!14}{rm}\colorbox{red!9}{pa} stated: we \colorbox{red!5}{have} \colorbox{red!6}{strongly} encouraged the company to investigate \colorbox{red!5}{options} that \colorbox{green!73}{don}' t \colorbox{red!5}{en}tail releasing the \colorbox{red!7}{material} to the environment and to develop a management plan to eliminate this potential hazard; \colorbox{red!6}{however}, \colorbox{red!10}{gb}\colorbox{red!16}{rm}\colorbox{red!19}{pa} does not have legislative control over how the< un\colorbox{green!79}{k}> tailings dam is managed.\colorbox{green!63}{=}==< unk>=== following a\\\\
 \colorbox{red!32}{of} warped \colorbox{green!73}{tour}. following this, a lesson in romantics was \colorbox{green!93}{released} on july 10 through fearless records. in august, the band went on tour with olympia \colorbox{red!13}{and} sound \colorbox{green!81}{the} alarm. the music video \colorbox{red!13}{for}\colorbox{red!17}{"} \colorbox{red!71}{when} \colorbox{red!53}{i} \colorbox{red!41}{\textcolor{blue}{get}} \colorbox{red!100}{home}\colorbox{red!23}{,} you\colorbox{red!10}{'} re so dead\colorbox{green!70}{"}, directed by marco de la torre, was filmed in september. \colorbox{red!11}{in} late \colorbox{red!14}{september} 2007, the band supported paramore \colorbox{red!11}{in} japan an d australia\colorbox{red!14}{.} the band went \colorbox{red!9}{on} \colorbox{green!100}{a} co@\colorbox{red!12}{-}@ headlining tour with madina lake in october and november. the" \colorbox{red!23}{when} \colorbox{red!43}{i} \colorbox{red!41}{get} \colorbox{red!71}{home}\colorbox{red!21}{,} you' re so \colorbox{red!17}{dead}" music \colorbox{red!13}{video} \colorbox{red!18}{was} released \colorbox{green!74}{on} november 14, and the single was released on\\\\
 \colorbox{red!49}{of} the english football league including promo tion and relegation. the player' s \colorbox{red!22}{team} begins with a \colorbox{red!20}{low} rating in an \colorbox{red!68}{8}\colorbox{red!30}{@}\colorbox{red!23}{-}\colorbox{red!31}{@} \colorbox{red!35}{team} \colorbox{red!26}{league}. by winning games, the player earns credits, which can be used to purchase the contracts of free agents\colorbox{red!34}{.} \colorbox{red!33}{by} finishing \colorbox{red!32}{high} \colorbox{red!45}{in} \colorbox{red!71}{the} \colorbox{red!44}{8}\colorbox{red!46}{@}\colorbox{red!92}{-}\colorbox{red!36}{\textcolor{blue}{@}} \colorbox{red!52}{team} \colorbox{red!100}{league}\colorbox{red!57}{,} \colorbox{red!21}{the} player' \colorbox{red!22}{s} \colorbox{red!27}{team} \colorbox{red!37}{advances} to a 16@-@ team league and eventually a 32@-@ team league\colorbox{red!29}{.} the player improves their team by periodically signing free agents\colorbox{red!22}{,} as the \colorbox{green!100}{competition} is tougher in each league. the player wins the \colorbox{red!28}{mode} after winning a playoff tournament in the 32@-@ team \colorbox{red!25}{league}\\\\
\colorbox{red!31}{} o\colorbox{red!9}{da} ministra kultury \colorbox{red!21}{i} \colorbox{red!38}{s}\colorbox{red!57}{z}\colorbox{red!100}{tu}\colorbox{red!21}{ki} ii< unk\colorbox{red!7}{>}) 1972 – member of commission" poland 2000" of the polish \colorbox{green!39}{academy} of sciences 1973 prize of the minister of \colorbox{green!46}{foreign} affairs for popular\colorbox{red!12}{ization} of polish culture \colorbox{red!10}{abroad}(< unk> ministra< unk>< unk> za< unk>< unk> kultury za< \colorbox{red!9}{un}k>\colorbox{red!12}{)} literary prize of the minister of culture and art(< un\colorbox{red!12}{k}>< unk> mini\colorbox{red!16}{stra} ku\colorbox{red!20}{lt}\colorbox{red!34}{ury} \colorbox{red!22}{i} \colorbox{red!57}{s}\colorbox{red!41}{\textcolor{blue}{z}}\colorbox{red!74}{tu}\colorbox{red!40}{ki}\colorbox{red!26}{)} and honorary member of science \colorbox{red!14}{fiction} \colorbox{red!10}{writers} of america \colorbox{red!11}{1976} \colorbox{green!67}{–} \colorbox{green!100}{state} \colorbox{red!11}{prize} 1st level in the area of\\\\
 \colorbox{red!30}{to} power the antarctic outpost. above earth, ba' \colorbox{red!15}{al}' s armada arrives. to \colorbox{red!11}{the} displeasure of his subordinates, the other system lords\colorbox{red!12}{,} ba\colorbox{red!16}{'} al \colorbox{green!82}{announces} that he will treat the \colorbox{red!12}{tau}< \colorbox{green!44}{un}k> leniently. suspicious about ba\colorbox{red!12}{'} \colorbox{red!14}{al}' s thorough knowledge \colorbox{red!13}{of} earth, qetesh betrays \colorbox{red!18}{him} and forces him to tell her everything. she orders the \colorbox{red!15}{destruction} of mcmurdo and the ancient outpost in ba\colorbox{red!15}{'} \colorbox{red!16}{al}\colorbox{red!16}{'} s name\colorbox{red!15}{,} \colorbox{red!16}{but} \colorbox{red!13}{she} \colorbox{red!76}{kills} \colorbox{red!100}{ba}\colorbox{red!60}{\textcolor{blue}{'}} \colorbox{red!100}{al} \colorbox{red!37}{when} \colorbox{green!94}{tea}l' c discovers what she \colorbox{red!12}{is} doing. as teal' \colorbox{green!30}{c} escapes to \colorbox{green!100}{an} \colorbox{red!16}{al}< unk>, q\colorbox{red!13}{ete}sh\\\\
\colorbox{red!100}{(} 156+ kn) each fuel capacity: 18@,@ 000 lb( 8@,@ 200 kg) internally, or 26@\colorbox{red!10}{,}@ 000 lb( 12@,@ 000 kg) with two external fuel tanks performance maximum speed: at altitude: mach 2@\colorbox{red!17}{.}@ 25( \colorbox{red!19}{1}@\colorbox{red!15}{,}\colorbox{red!11}{@} 500 \colorbox{red!33}{mph}\colorbox{red!18}{,} 2@\colorbox{red!18}{,}@ 410 \colorbox{red!14}{km}/ \colorbox{red!18}{h}\colorbox{red!17}{)}[ \colorbox{red!14}{estimated}] supercruise: mach 1@\colorbox{red!15}{.}@ 82\colorbox{green!100}{(} \colorbox{red!12}{1}@\colorbox{red!18}{,}@ \colorbox{red!14}{220} \colorbox{red!61}{mph}\colorbox{red!97}{,} \colorbox{red!56}{1}\colorbox{red!26}{\textcolor{blue}{@}}\colorbox{red!41}{,}@ 960 \colorbox{red!29}{km}\colorbox{red!12}{/} \colorbox{red!31}{h}\colorbox{red!24}{)} range\colorbox{red!21}{:}> 1@,@ 600 nmi( 1@,\colorbox{green!50}{@} 840 mi, 2@,@ \colorbox{red!13}{960}\\\\
\newline \newline
 {\bf Transformer factor 322 in layer 10 with saliency map \newline Explaination: biography, someone born in some year... }  \newline \newline 
\colorbox{red!5}{.} \colorbox{red!6}{only} three \colorbox{red!7}{pitchers} threw more complete games in major \colorbox{red!8}{league} careers shorter than getzein' s nine@-@ year career. getzein had his most extensive playing time with \colorbox{red!8}{the} detroit wolverines, compiling records of 30@-@ 11 and 29@-@ 13 in 1886 and 1887\colorbox{red!7}{.} in the 1887 world series( which detroit won, 10 games to 5), getzein pitched six \colorbox{red!7}{complete} games and \colorbox{red!6}{compiled} a 4\colorbox{red!5}{@}-@ 2 record with a 2@.@ 48 era\colorbox{red!6}{.} \colorbox{red!6}{he} \colorbox{red!8}{also} won 23 games \colorbox{red!9}{for} \colorbox{red!5}{the} \colorbox{red!9}{boston} bean\colorbox{red!8}{ea}\colorbox{red!11}{ters} \colorbox{red!11}{in} \colorbox{red!20}{1890}\colorbox{red!11}{.}\colorbox{red!6}{=}\colorbox{red!10}{=} early years\colorbox{red!13}{=}\colorbox{red!7}{=} get\colorbox{red!6}{ze}\colorbox{green!100}{in} \colorbox{red!17}{was} \colorbox{red!100}{born} \colorbox{red!17}{in} \colorbox{red!100}{\textcolor{blue}{1864}}\\\\
 \colorbox{green!100}{and} telegraph lines and networks. the west construction company, based in chattanooga, tennessee, was a general contracting and construction firm also involved in the operation and maintenance of railway, telephone, and telegraph lines.== personal life===== marriage and \colorbox{red!19}{children}=== on april \colorbox{red!21}{10}\colorbox{red!20}{,} \colorbox{red!18}{1875}\colorbox{red!22}{,} \colorbox{red!14}{in} hampshire county, flournoy \colorbox{red!55}{married} \colorbox{red!23}{frances}" fannie" ann armstrong white\colorbox{red!100}{(} \colorbox{red!60}{april} \colorbox{red!30}{10}\colorbox{red!99}{,} \colorbox{red!100}{\textcolor{blue}{1844}} \colorbox{red!23}{–} february \colorbox{red!20}{25}\colorbox{red!26}{,} \colorbox{red!41}{1922}\colorbox{red!31}{)}\colorbox{red!26}{,} \colorbox{red!20}{the} \colorbox{red!29}{daughter} of hampshire \colorbox{red!23}{county} \colorbox{red!19}{clerk} \colorbox{red!21}{of} \colorbox{red!25}{court} \colorbox{red!20}{john} baker white \colorbox{red!27}{and} his wife frances ann streit white\colorbox{red!25}{.} frances white' s brother, robert white, served as west virginia \colorbox{red!23}{attorney} general, and her\\\\
 \colorbox{red!7}{buffalo}, new york businessman who made his fortune in \colorbox{red!9}{five}@-@ \colorbox{red!7}{and}@-@ dime stores. he \colorbox{green!94}{merged} \colorbox{red!6}{his} more than \colorbox{red!8}{100} stores with those of his first cousins, frank winfield woolworth and charles woolworth, to form \colorbox{red!7}{the} f. \colorbox{green!65}{w}. \colorbox{red!7}{wool}worth company. he went on to hold prominent positions in the merged company as well as marine trust co. he was the \colorbox{green!100}{father} \colorbox{red!7}{of} seymour h. knox ii and grandfather of seymour h\colorbox{red!9}{.} knox iii and northrup knox, the co@-@ founders of the buffalo sabres in the \colorbox{red!7}{national} \colorbox{red!16}{hockey} \colorbox{red!12}{league}\colorbox{red!12}{.}=\colorbox{red!10}{=} \colorbox{red!9}{biography}\colorbox{red!7}{=}= \colorbox{red!16}{he} \colorbox{red!36}{was} \colorbox{red!100}{born} \colorbox{red!16}{in} \colorbox{red!11}{april} \colorbox{red!100}{\textcolor{blue}{1861}} \colorbox{red!17}{in} \colorbox{red!12}{russell}\colorbox{red!25}{,} \colorbox{red!18}{saint} \colorbox{red!15}{lawrence}\\\\
 stars for eighteen years. the american film institute( afi) ranked cooper eleventh on its list of the twenty five greatest male stars of classic hollywood \colorbox{red!25}{cinema}.== early life== frank james \colorbox{red!27}{cooper} was born on may 7, \colorbox{red!21}{1901}, \colorbox{red!22}{at} 730 eleventh avenue in helena, montana \colorbox{red!36}{to} english \colorbox{red!28}{immigrants} alice( \colorbox{red!35}{nee} brazier, 1873 \colorbox{red!26}{–} \colorbox{red!35}{1967}) \colorbox{red!71}{and} \colorbox{red!48}{charles} \colorbox{red!27}{henry} cooper\colorbox{red!100}{(} \colorbox{red!100}{\textcolor{blue}{1865}} \colorbox{red!37}{–} \colorbox{red!71}{1946}\colorbox{red!23}{)}\colorbox{red!63}{.} \colorbox{red!21}{his} \colorbox{red!49}{father} \colorbox{red!26}{emigrated} from \colorbox{red!21}{houghton} regis\colorbox{red!21}{,} bedfordshire \colorbox{red!20}{and} became a prominent \colorbox{red!34}{lawyer}, rancher, and eventually a \colorbox{red!23}{montana} supreme \colorbox{red!23}{court} justice\colorbox{red!45}{.} his mother emigrated from gillingham, kent and married charles in montana\colorbox{red!21}{.} in 1906, charles purchased the 600@-@ \colorbox{red!25}{acre}\\\\
 \colorbox{red!44}{orange}( \colorbox{red!21}{1971}), which ku\colorbox{red!15}{brick} pulled from circulation in the \colorbox{red!17}{uk} following a mass media frenzy — most of his films were nominated for oscars, golden globes, or bafta awards. his last film, eyes wide shut, was completed shortly before his death in 1999.\colorbox{red!17}{=}= early life\colorbox{green!100}{=}= stanley kubrick was born on july 26, 1928, in the \colorbox{red!16}{lying}@-@ in hospital at 307 second avenue in manhattan, new \colorbox{red!13}{york} city. he \colorbox{red!18}{was} the first \colorbox{red!16}{of} \colorbox{red!17}{two} \colorbox{red!67}{children} \colorbox{red!24}{of} \colorbox{red!37}{jacob} \colorbox{red!20}{leonard} kubrick\colorbox{red!52}{(} \colorbox{red!39}{may} \colorbox{red!47}{21}\colorbox{red!100}{,} \colorbox{red!100}{\textcolor{blue}{1902}} \colorbox{red!53}{–} october 19\colorbox{red!18}{,} \colorbox{red!30}{1985}\colorbox{red!34}{)}\colorbox{red!34}{,} \colorbox{red!25}{known} \colorbox{red!26}{as} jack or jacques\colorbox{red!17}{,} \colorbox{red!25}{and} his \colorbox{red!16}{wife} sadie gertrude kubrick\\\\
 managed with a catch and release regulation. trophy trout and wild brook trout enhancement regulations apply to the remainder. a \colorbox{green!13}{total} of 31 class \colorbox{green!66}{a} wild trout waters have been \colorbox{green!47}{designated} as wilderness trout streams. fishing in class a wild trout waters is permitted year@-@ round, \colorbox{red!14}{although} the killing of fish is forbidden from labor day to the beginning \colorbox{red!13}{of} the following year' s trout season\colorbox{red!26}{.}== gallery=== henry bell gil\colorbox{red!12}{kes}\colorbox{red!13}{on}= \colorbox{red!20}{henry} \colorbox{green!100}{bell} gilkes\colorbox{red!27}{on}\colorbox{red!100}{(} \colorbox{red!59}{june} \colorbox{red!48}{6}\colorbox{red!86}{,} \colorbox{red!100}{\textcolor{blue}{1850}} \colorbox{red!30}{–} september 29\colorbox{red!42}{,} \colorbox{red!40}{1921}\colorbox{red!49}{)} \colorbox{red!28}{was} \colorbox{red!15}{an} \colorbox{red!20}{american} \colorbox{red!27}{lawyer}, \colorbox{red!21}{politician}, \colorbox{red!13}{school} administrator, and \colorbox{red!31}{banker} in west \colorbox{red!23}{virginia}\colorbox{red!28}{.} gilkeson was \colorbox{red!23}{born} in moorefield,\\\\
 movement, there have been few more remarkable figures than \colorbox{red!22}{mar}jory stoneman douglas."== early life== marjory stone\colorbox{red!48}{man} \colorbox{red!51}{was} \colorbox{red!44}{born} \colorbox{red!30}{on} april 7, 189 0, \colorbox{red!34}{in} minneapolis, \colorbox{red!47}{minnesota}\colorbox{red!36}{,} \colorbox{red!43}{the} \colorbox{red!60}{only} \colorbox{red!58}{child} \colorbox{red!57}{of} \colorbox{red!66}{frank} \colorbox{red!34}{bryant} stoneman\colorbox{red!79}{(} \colorbox{red!100}{\textcolor{blue}{1857}} \colorbox{red!55}{–} \colorbox{red!100}{1941}\colorbox{red!67}{)} \colorbox{red!46}{and} \colorbox{red!36}{lillian} trefethen\colorbox{red!33}{(} \colorbox{red!51}{1859} \colorbox{red!55}{–} 1912\colorbox{red!41}{)}\colorbox{red!32}{,} a concert \colorbox{red!47}{violinist}\colorbox{red!39}{.} one \colorbox{red!37}{of} \colorbox{red!34}{her} earliest memories was her father reading to her the song of hiawatha, at which she burst into sobs upon hearing that the tree had to give its life in order to provide hiawatha the wood for a canoe. she was an early and voracious reader\\\\
 \colorbox{red!29}{amazon}\colorbox{red!14}{.} com.=== dvd release==== \colorbox{red!20}{johann} mickl= \colorbox{red!18}{johann} \colorbox{red!17}{mick}\colorbox{red!29}{l}\colorbox{red!100}{(} \colorbox{red!34}{18} \colorbox{red!63}{april} \colorbox{red!100}{\textcolor{blue}{1893}} \colorbox{red!38}{–} 10 \colorbox{red!34}{april} \colorbox{red!39}{1945}\colorbox{red!43}{)} \colorbox{red!50}{was} an \colorbox{red!18}{austrian}@-@ \colorbox{red!14}{born} \colorbox{red!17}{general}le\colorbox{red!15}{utnant} and division commander \colorbox{red!16}{in} the \colorbox{red!20}{german} army \colorbox{red!13}{during} world \colorbox{red!19}{war} ii\colorbox{red!14}{,} \colorbox{red!23}{and} was one of only 88\colorbox{red!17}{2} recipients of the knight' s cross of the iron cross with oak leaves\colorbox{red!20}{.} he was commissioned shortly before the outbreak of world war i, and served with austro@-@ hungarian forces on the eastern and italian fronts as company commander in the imperial@-@ royal mountain troops\colorbox{red!16}{.} during world war i he was decorated several \colorbox{red!15}{times} for bravery and leadership, \colorbox{red!18}{and}\\\\
 \colorbox{green!100}{very} unusual properties, such as a quantum critical point behavior, exotic supercondu\colorbox{red!18}{ct}ivity, \colorbox{red!17}{and} high@-@ temperature ferromagnetism\colorbox{red!26}{.}= babe ruth= \colorbox{red!21}{george} herman \colorbox{red!18}{ruth} \colorbox{red!58}{jr}\colorbox{red!54}{.}\colorbox{red!100}{(} \colorbox{red!63}{february} \colorbox{red!28}{6}\colorbox{red!67}{,} \colorbox{red!100}{\textcolor{blue}{1895}} \colorbox{red!25}{–} august 16\colorbox{red!26}{,} \colorbox{red!44}{1948}\colorbox{red!46}{)}\colorbox{red!36}{,} \colorbox{red!26}{better} \colorbox{red!35}{known} as babe ruth\colorbox{red!23}{,} \colorbox{red!21}{was} an \colorbox{red!23}{american} \colorbox{red!20}{professional} baseball \colorbox{red!21}{player} \colorbox{red!20}{whose} career in \colorbox{red!21}{major} league baseball( mlb) spanned 22 seasons, from 1914 \colorbox{red!16}{through} 1935\colorbox{red!20}{.} nicknamed" the bambino" and" the sultan of swat", he began his mlb career as a stellar left@-@ handed pitcher for the boston red sox, \colorbox{red!20}{but} achieved his greatest fame as a slugging outfielder for the\\\\
 \colorbox{green!100}{air} in regular scheduled services. it includes the city, country, airport and the period in which the \colorbox{red!15}{airline} served the airport. hubs are \colorbox{red!15}{denoted} with a dagger()\colorbox{red!11}{.}= william s. taylor= william sylvester \colorbox{red!17}{taylor}\colorbox{red!100}{(} \colorbox{red!25}{october} \colorbox{red!14}{10}\colorbox{red!69}{,} \colorbox{red!100}{\textcolor{blue}{1853}} \colorbox{red!34}{–} \colorbox{red!24}{august} 2\colorbox{red!28}{,} \colorbox{red!42}{1928}\colorbox{red!25}{)} \colorbox{red!38}{was} the \colorbox{red!18}{33rd} governor \colorbox{red!22}{of} \colorbox{red!12}{kentucky}\colorbox{red!32}{.} \colorbox{red!13}{he} \colorbox{red!15}{was} \colorbox{red!13}{initially} declared the winner of the disputed gubernatorial \colorbox{red!16}{election} \colorbox{red!13}{of} \colorbox{red!21}{1899}, but the kentucky general assembly, dominated by the \colorbox{red!22}{democrats}, reversed the election results, giving the victory to his democratic party( united states) opponent\colorbox{green!29}{,} william goebel. taylor served only 50 \colorbox{red!12}{days} as governor. a poorly educated but politically as\colorbox{red!15}{tute} lawyer, taylor\\\\
 \colorbox{red!33}{woods} hole, massachusetts, where he studied marine bioluminescence. he also worked at the woods hole oceanographic institution\colorbox{red!31}{.}\colorbox{red!24}{=}\colorbox{red!24}{=} early life\colorbox{red!17}{=}= \colorbox{red!77}{george} \colorbox{red!100}{\textcolor{blue}{thomas}} \colorbox{red!100}{reynolds} \colorbox{green!100}{was} \colorbox{red!70}{born} \colorbox{red!30}{in} \colorbox{red!33}{trenton}\colorbox{red!27}{,} new \colorbox{red!18}{jersey} \colorbox{red!19}{on} may 27\colorbox{red!20}{,} \colorbox{red!36}{1917}\colorbox{red!23}{,} \colorbox{red!20}{the} \colorbox{red!43}{son} of george w. \colorbox{red!16}{reynolds}, a< unk> for \colorbox{red!19}{the} \colorbox{red!19}{pennsylvania} \colorbox{red!20}{railroad}\colorbox{red!19}{,} \colorbox{red!24}{and} his wife laura, a secretary with the new jersey department of geology\colorbox{red!25}{.} he attended franklin junior high school in highland park, new jersey, until year 10, and then new brunswick high school. he received a bachelor' s degree in physics from rutgers university in \colorbox{red!25}{1939}. he then entered princeton university, \colorbox{red!18}{where} \colorbox{green!20}{was} awarded\\\\
\colorbox{red!22}{=}= shaughnessy \colorbox{red!32}{was} \colorbox{red!47}{born} \colorbox{red!27}{on} march 6, \colorbox{red!23}{1892} \colorbox{red!33}{in} \colorbox{red!34}{st}. cloud\colorbox{red!28}{,} \colorbox{red!41}{minnesota}\colorbox{red!24}{,} \colorbox{red!72}{the} \colorbox{red!70}{second} \colorbox{red!64}{son} \colorbox{red!41}{of} \colorbox{red!100}{\textcolor{blue}{lucy}} \colorbox{red!100}{ann}\colorbox{red!25}{(} foster) \colorbox{red!32}{and} \colorbox{red!31}{edward} shaugh\colorbox{red!24}{nessy}. he attended \colorbox{red!16}{north} st. paul high school, and prior to college, had no athletic \colorbox{red!24}{experience}. when he attended the university of minnesota\colorbox{red!19}{,} however, he p layed college football under head coach henry l. williams and alongside halfback bernie bierman\colorbox{red!21}{.} shaughnessy considered williams to be \colorbox{red!16}{football}' s greatest teacher, and \colorbox{green!100}{williams} considered him to be the best passer from \colorbox{red!18}{the} \colorbox{red!21}{midwest}. shaughnessy handled both \colorbox{red!23}{the} passing and kicking duties for the team\colorbox{red!22}{.} he played on\\\\
\colorbox{green!100}{} \colorbox{red!20}{s} gregoras likewise \colorbox{red!23}{avoids} negative comments, as do most modern historians\colorbox{red!25}{.}= george \colorbox{red!27}{nico}l\colorbox{red!21}{(} baseball\colorbox{green!34}{)}= george \colorbox{red!35}{edward} \colorbox{red!38}{nico}\colorbox{red!21}{l}\colorbox{red!100}{(} \colorbox{red!91}{october} \colorbox{red!40}{17}\colorbox{red!74}{,} \colorbox{red!100}{\textcolor{blue}{1870}} \colorbox{red!57}{–} \colorbox{red!45}{august} 4\colorbox{red!43}{,} \colorbox{red!56}{1924}\colorbox{red!48}{)} \colorbox{red!55}{was} \colorbox{red!20}{an} \colorbox{red!38}{american} baseball pitcher and outfielder who \colorbox{red!27}{played} three seasons in major league baseball( mlb)\colorbox{red!32}{.} he \colorbox{red!25}{played} for the st. louis browns, chicago colts, pittsburgh pirates and louisville colonels from 1890 to 1894\colorbox{red!25}{.} possessing the rare combination of batting right@-@ handed and throwing left@-@ handed, he served primarily as a right fielder when he did not pitch\colorbox{red!22}{.} signed by \colorbox{red!22}{the} browns without having previously played any minor league baseball, nico\colorbox{red!24}{l} made his\\\\
 \colorbox{red!18}{dispatched} powell and major benjamin mcculloch to utah to ease tensions with brigham \colorbox{green!63}{young} and the mormons. powell assumed his \colorbox{red!8}{senate} seat on his return from utah, just prior to the election of abraham lincoln as president. powell became an outspoken critic of lincoln\colorbox{red!11}{'} s administration, so much so that the kentucky general assembly asked for his resignation and some of his fellow senators tried to have him expelled from the body\colorbox{red!15}{.} both \colorbox{green!98}{groups} later renounced their actions\colorbox{red!15}{.} \colorbox{red!9}{powell} died at his home \colorbox{red!11}{near} \colorbox{red!12}{henderson}\colorbox{red!12}{,} \colorbox{red!8}{kentucky} shortly following a failed \colorbox{red!13}{bid} to return to \colorbox{red!9}{the} \colorbox{red!12}{senate} \colorbox{red!11}{in} \colorbox{red!22}{1867}.\colorbox{red!10}{=}\colorbox{red!11}{=} early \colorbox{green!100}{life}== \colorbox{red!27}{powell} \colorbox{red!42}{was} \colorbox{red!100}{born} \colorbox{red!42}{on} \colorbox{red!41}{october} \colorbox{red!37}{6}\colorbox{red!65}{,} \colorbox{red!100}{\textcolor{blue}{1812}} near \colorbox{red!46}{henderson}\colorbox{red!25}{,}\\\\
 \colorbox{red!44}{the} \colorbox{red!18}{army} in 1948. \colorbox{red!19}{he} was promoted to lieutenant general just before his retirement on 29 february 1948 in recognition of his leadership of the bomb program. by a special act of congress, his date of rank was backdated to 16 july 1945, the date of the trinity nuclear test\colorbox{red!22}{.} \colorbox{red!16}{groves} went on to become a vice@-@ president at sperry rand.== early life== leslie \colorbox{red!21}{richard} \colorbox{red!17}{groves} jr\colorbox{red!20}{.} \colorbox{red!27}{was} \colorbox{red!38}{born} \colorbox{red!14}{in} albany, new \colorbox{red!22}{york}, \colorbox{red!28}{on} 17 august \colorbox{red!20}{1896}\colorbox{red!17}{,} \colorbox{red!23}{the} \colorbox{red!21}{third} son \colorbox{red!19}{of} four \colorbox{red!31}{children} \colorbox{red!40}{of} a pastor\colorbox{red!21}{,} \colorbox{red!100}{leslie} \colorbox{red!100}{\textcolor{blue}{richard}} \colorbox{red!91}{groves} \colorbox{red!20}{sr}\colorbox{green!100}{.}, \colorbox{red!20}{and} his wife gwen \colorbox{red!26}{nee} griffith\colorbox{red!41}{.} a descendant \colorbox{red!21}{of} french huguenots who\\\\
\colorbox{red!16}{,} burns died on november 11\colorbox{red!21}{,} \colorbox{red!29}{1928} in brooklyn, new york\colorbox{red!24}{.}== biography== thomas \colorbox{red!22}{p}. \colorbox{red!23}{burns} \colorbox{red!52}{was} \colorbox{red!100}{born} \colorbox{red!38}{on} \colorbox{red!47}{september} \colorbox{red!43}{6}\colorbox{red!59}{,} \colorbox{red!100}{\textcolor{blue}{1864}}\colorbox{red!29}{,} \colorbox{red!24}{in} \colorbox{red!76}{philadelphia}\colorbox{red!26}{.} his \colorbox{red!22}{parents}, \colorbox{red!18}{patrick} \colorbox{red!19}{and} \colorbox{red!25}{mary} burns, were both irish immigrants. in 1883\colorbox{red!18}{,} burns began his professional baseball career as a pitcher with harrisburg of the minor@-@ league interstate association. \colorbox{red!18}{on} the year, burns \colorbox{red!16}{posted} an earned run average( era) of 2@\colorbox{red!17}{.}@ 30 over 20 games pitched, 15 \colorbox{red!12}{of} which were starts\colorbox{red!16}{.} \colorbox{red!19}{when} he wasn' t pitching, burns played second and third base. burns began the 1884 season \colorbox{red!16}{playing} for the \colorbox{red!17}{wilmington} quicksteps,\\\\
@ beats".== \colorbox{red!18}{credits} and personnel== lady gaga – vocals, songwriter and producer redone – songwriter, \colorbox{red!14}{producer}, vocal editing, vocal arrangement, audio engineering, instrumentation, programming, and recording at tour bus in europe trevor mu\colorbox{red!16}{zzy} \colorbox{red!12}{–} recording, vocal editing, audio engineering, \colorbox{red!16}{and} audio mixing at larrabee, north holly wood, los angeles, california gene grimaldi – audio mastering at oasis mastering, burbank, california \colorbox{red!20}{credits} adapted from born this \colorbox{red!17}{way} album liner \colorbox{red!14}{notes}\colorbox{red!13}{.}== \colorbox{green!100}{charts}==\colorbox{red!15}{=} \colorbox{red!14}{travis} \colorbox{red!20}{jackson}\colorbox{red!16}{=} travis calvin jackson\colorbox{red!54}{(} \colorbox{red!50}{november} \colorbox{red!42}{2}\colorbox{red!100}{,} \colorbox{red!100}{\textcolor{blue}{1903}} \colorbox{red!42}{–} july \colorbox{red!15}{27}\colorbox{red!39}{,} \colorbox{red!43}{1987}\colorbox{red!49}{)} \colorbox{red!53}{was} \colorbox{red!18}{an} \colorbox{red!44}{american} \colorbox{red!21}{baseball} \colorbox{red!48}{shortstop}\colorbox{red!34}{.}\\\\
\colorbox{red!21}{=} mons\colorbox{red!25}{on} \colorbox{red!47}{was} \colorbox{red!65}{born} \colorbox{red!18}{on} august 21, 1927\colorbox{red!25}{,} \colorbox{red!23}{in} salt lake \colorbox{red!20}{city}\colorbox{red!19}{,} \colorbox{red!17}{utah} \colorbox{red!100}{to} \colorbox{red!84}{g}. \colorbox{red!100}{\textcolor{blue}{spencer}} \colorbox{red!53}{mons}\colorbox{red!29}{on}\colorbox{red!19}{(} \colorbox{red!41}{1901} – 1979\colorbox{red!28}{)} \colorbox{red!43}{and} \colorbox{red!16}{gladys}< unk> monson( \colorbox{red!19}{1902} – 1973)\colorbox{red!28}{.} \colorbox{red!24}{the} \colorbox{red!22}{second} of \colorbox{red!19}{six} \colorbox{red!36}{children}\colorbox{red!24}{,} he grew up in a" tight@-@ knit" family — many of his mother' s \colorbox{red!18}{relatives} living on \colorbox{red!15}{the} same street and the extended family frequently going on trips together. the family' s neighborhood included several residents of mexican descent, an environment in which he \colorbox{green!100}{says} he developed a love for the mexican people and culture. monso n often spent weekends with relatives on their farms in granger(\\\\
 \colorbox{green!73}{it}. \colorbox{red!9}{anderson} \colorbox{green!93}{was} a professional accordion player and wrote poetry \colorbox{green!70}{for} various american pagan \colorbox{red!10}{magazines}\colorbox{red!10}{.} in 1970, he published his first book of poetry, thorns of \colorbox{red!8}{the} blood rose, which contained devotional religious poetry dedicated to the goddess; it won the clover international poetry competition award in 1975. anderson continued to promote the feri tradition until his death, at which point april \colorbox{red!11}{ni}ino was \colorbox{red!7}{appointed} as \colorbox{red!7}{the} new grandmaster of the tradition\colorbox{red!15}{.}== early life===== childhood: \colorbox{red!8}{1917} \colorbox{green!100}{–} 1931=== anderson \colorbox{green!56}{was} born on may 21\colorbox{red!8}{,} \colorbox{red!12}{1917} \colorbox{red!8}{at} \colorbox{red!12}{the} buffalo horn ranch in \colorbox{red!13}{clayton}\colorbox{red!14}{,} new \colorbox{red!11}{mexico}\colorbox{red!24}{.} \colorbox{red!36}{his} \colorbox{red!100}{parents} \colorbox{red!44}{were} \colorbox{red!33}{hi}\colorbox{red!55}{lb}\colorbox{red!27}{art} \colorbox{red!100}{\textcolor{blue}{alexander}} \colorbox{red!66}{anderson}\\\\
 \colorbox{green!100}{was} elsewhere. he had recently become engaged and bought his first house in hillsborough. franklin and benjamin pierce were among the prominent citizens who welcomed president jackson to the state on his visit in mid@-@ 1833.=== marriage and children=== \colorbox{red!16}{on} november 19\colorbox{red!16}{,} 1834\colorbox{red!15}{,} \colorbox{red!15}{pierce} \colorbox{red!64}{married} \colorbox{red!33}{jane} \colorbox{green!72}{means} \colorbox{red!27}{appleton}\colorbox{red!69}{(} \colorbox{red!49}{march} \colorbox{red!37}{12}\colorbox{red!100}{,} \colorbox{red!100}{\textcolor{blue}{1806}} \colorbox{red!18}{–} \colorbox{red!30}{december} 2\colorbox{red!41}{,} \colorbox{red!58}{1863}\colorbox{red!21}{)}\colorbox{red!20}{,} the \colorbox{red!20}{daughter} of jesse appleton, a \colorbox{red!17}{congregational} \colorbox{red!20}{minister} and former president \colorbox{red!17}{of} bowdo\colorbox{red!12}{in} \colorbox{red!20}{college}, \colorbox{red!24}{and} \colorbox{red!20}{elizabeth} means\colorbox{red!29}{.} the appletons were \colorbox{red!18}{prominent} whigs, in contrast with the pierces' democratic affiliation. jane was shy, devoutly religious, and pro@-@ temperance\\\\
 which took delivery of its eight and last globemaster in november 2015; no. 38 squadron, operating king airs; and the australian army' s 68 ground liaison section\colorbox{red!26}{.} all units are based at amberley\colorbox{red!22}{,} with the exception of no. 38 squadron, \colorbox{green!33}{located} at townsville\colorbox{red!25}{.}= clark shaughnessy= clark \colorbox{red!32}{daniel} sha\colorbox{red!19}{ugh}nessy\colorbox{red!30}{(} \colorbox{red!18}{originally} o' shaughnessy\colorbox{green!100}{)}\colorbox{red!99}{(} \colorbox{red!55}{march} \colorbox{red!46}{6}\colorbox{red!100}{,} \colorbox{red!100}{\textcolor{blue}{1892}} \colorbox{red!61}{–} \colorbox{red!22}{may} 15\colorbox{red!43}{,} \colorbox{red!67}{1970}\colorbox{red!67}{)} \colorbox{red!57}{was} \colorbox{red!31}{an} \colorbox{red!38}{american} football \colorbox{red!24}{coach} and \colorbox{red!25}{inn}\colorbox{red!22}{ova}\colorbox{red!26}{tor}\colorbox{red!52}{.} \colorbox{red!28}{he} \colorbox{red!25}{is} sometimes called the" father of the t formation" and the original founder of the \colorbox{red!28}{forward} pass, although that system had previously been used as early as the 1880s\\\\
\newline \newline
 {\bf Transformer factor 386 in layer 10 with saliency map \newline Explaination: topic: war }  \newline \newline
 \colorbox{red!14}{he} was \colorbox{red!11}{awarded} a companion of the \colorbox{red!11}{order} of st michael and st george \colorbox{red!6}{for} \colorbox{red!12}{his} \colorbox{red!8}{command} \colorbox{red!9}{of} \colorbox{green!100}{the} \colorbox{red!37}{4th} machine \colorbox{red!100}{gun} \colorbox{red!100}{\textcolor{blue}{battalion}}\colorbox{red!27}{,} \colorbox{red!8}{the} \colorbox{red!8}{recommendation} \colorbox{red!7}{of} \colorbox{red!13}{which} particularly citing his success during attacks \colorbox{red!7}{on} the hindenburg line\colorbox{red!12}{.} murray' s final honour came on 11 july \colorbox{red!6}{1919}, when \colorbox{red!7}{he} was mentioned in despa\colorbox{green!23}{tch}es for the fourth time, having received his third mention on 31 december 1918. from june to september 1919\colorbox{red!7}{,} \colorbox{red!11}{murray} — along with fellow australian victoria cross recipient william donovan joynt — led parties of aif members on \colorbox{red!7}{a} tour \colorbox{red!6}{of} the farming districts of britain and denmark to study agricultural methods \colorbox{red!6}{under} the \colorbox{red!6}{education} schemes. \colorbox{red!9}{after} touring through france \colorbox{red!7}{and} belgium\colorbox{red!9}{,}\\\\
 \colorbox{red!55}{from} large@-@ calibre shells\colorbox{red!13}{;} one of them, allegedly a 14@-\colorbox{green!77}{@} inch( 356 mm) round, blew a large hole in \colorbox{green!53}{her} quarterdeck \colorbox{red!19}{and} wrecked the wardroom \colorbox{red!11}{and} the gunroom\colorbox{red!14}{.} she also took several hits by light shells that day, and, although she suffered damage to \colorbox{red!13}{her} superstructure, her fighting and \colorbox{red!16}{steaming} capabilities were not seriously impaired. the \colorbox{red!14}{ship} \colorbox{red!20}{also} participated in the main attack on the dardan\colorbox{red!17}{elles} \colorbox{red!14}{forts} on 18 \colorbox{red!23}{march}. \colorbox{red!22}{this} time \colorbox{red!13}{a} 6@\colorbox{red!14}{-}@ inch\colorbox{green!100}{(} 152 \colorbox{red!14}{mm}\colorbox{red!32}{)} \colorbox{red!61}{how}\colorbox{red!100}{\textcolor{blue}{itzer}} \colorbox{red!100}{battery} \colorbox{red!28}{opened} \colorbox{red!29}{fire} \colorbox{red!15}{on} agamemnon \colorbox{red!15}{and} \colorbox{red!13}{hit} her 12 times \colorbox{red!13}{in} 25 minutes\colorbox{red!15}{;} five of the\\\\
. lt. riefkohl, \colorbox{red!13}{who} was also the first \colorbox{green!32}{puerto} rican to graduate from the united states \colorbox{green!100}{naval} academy\colorbox{red!11}{,} served as a rear \colorbox{green!58}{admiral} in world war ii. frederick l. riefkohl' s \colorbox{red!10}{brother}, rudolph william riefkohl also served\colorbox{red!15}{.} riefkohl \colorbox{red!18}{was} \colorbox{red!17}{commissioned} a second \colorbox{red!9}{lieutenant} \colorbox{red!10}{and} \colorbox{red!26}{assigned} \colorbox{red!16}{to} \colorbox{red!18}{the} \colorbox{red!28}{63}\colorbox{red!32}{rd} \colorbox{red!100}{heavy} \colorbox{red!100}{\textcolor{blue}{artillery}} \colorbox{red!73}{regiment} \colorbox{red!31}{in} \colorbox{red!39}{france} where he actively participated \colorbox{green!63}{in} the meuse\colorbox{green!44}{@}-@ argonne \colorbox{red!28}{offensive}\colorbox{red!21}{.} \colorbox{red!16}{according} \colorbox{red!13}{to} the united states war department\colorbox{red!13}{,} \colorbox{red!14}{after} the war he served as captain of coastal artillery \colorbox{red!10}{at} the letterman army medical center in presidio of san francisco, in california( 1918).\\\\
 \colorbox{red!21}{washington} times@-@ herald, which ran the headline" hardy wild@\colorbox{red!24}{-}@ eyed aus\colorbox{green!79}{sies} called world' s finest troops\colorbox{green!100}{"}. an article in the chicago daily \colorbox{red!23}{news} told its \colorbox{red!26}{readers} that \colorbox{red!27}{australians}" in their realistic attitude \colorbox{green!65}{towards} power politics, prefer \colorbox{red!19}{to} \colorbox{green!64}{send} their boys to fight far overseas rather than fighting a battle in the suburbs of sydney"\colorbox{red!37}{.} \colorbox{red!45}{during} the \colorbox{red!28}{battle}\colorbox{red!18}{,} wave\colorbox{red!26}{ll} \colorbox{red!22}{had} \colorbox{red!41}{received} a \colorbox{red!21}{cable} \colorbox{red!90}{from} \colorbox{red!100}{\textcolor{blue}{general}} \colorbox{red!100}{sir} \colorbox{red!98}{john} dil\colorbox{red!28}{l} \colorbox{red!44}{stress}\colorbox{red!21}{ing} the \colorbox{red!19}{political} importance of such victories \colorbox{red!25}{in} the united states, where \colorbox{red!28}{president} franklin d. roosevelt \colorbox{red!17}{was} attempting to get the lend@-@ lease act passed. it was finally enacted in march \colorbox{red!23}{1941}. mackay wrote\\\\
\colorbox{red!75}{.} he also showed respect for occupied populations \colorbox{red!21}{and} never tolerated pillaging \colorbox{red!23}{nor} violence from his men. as \colorbox{red!20}{a} sign a gratitude, he was \colorbox{red!25}{offered} gifts several times but \colorbox{red!21}{he} was often seen refusing and sending \colorbox{red!15}{them} back. while on \colorbox{red!19}{campaign} in \colorbox{red!20}{tyrol}, he was recorded \colorbox{red!25}{to} \colorbox{red!22}{have} accepted a large sum of money but \colorbox{red!23}{he} immediately distributed it to the local hospitals. further evidence of his humanity was the ca re that \colorbox{red!22}{he} displayed for the lives and \colorbox{green!78}{well}@-@ being \colorbox{red!21}{of} his men, whom he was always reluctant to \colorbox{green!97}{sacrifice} for the sake of glory\colorbox{red!44}{.} \colorbox{red!28}{overall} \colorbox{red!69}{as} \colorbox{red!83}{a} \colorbox{green!100}{heavy} cavalry \colorbox{red!100}{\textcolor{blue}{commander}}\colorbox{red!100}{,} nanso\colorbox{green!100}{ut}y \colorbox{red!37}{was} \colorbox{red!12}{one} of the \colorbox{red!16}{best} men \colorbox{red!19}{available} \colorbox{red!43}{during} the \colorbox{red!46}{napoleonic}\\\\
@ 000 troops on 11 february. in march \colorbox{red!16}{1919}, princess matoika and rijndam raced each other from saint@-@ \colorbox{green!10}{na}zaire to \colorbox{red!16}{newport} news in a friendly competition that received national press coverage in the united states. rijndam, the slower ship\colorbox{red!18}{,} was just able to edge out the \colorbox{red!19}{princess} — \colorbox{red!16}{and} cut two days \colorbox{red!13}{from} her previous \colorbox{green!100}{fastest} crossing time — by appealing to the \colorbox{red!24}{honor} of \colorbox{green!34}{the} \colorbox{red!48}{soldiers} \colorbox{red!50}{of} the \colorbox{red!37}{133}\colorbox{red!62}{rd} \colorbox{red!100}{field} \colorbox{red!100}{\textcolor{blue}{artillery}}\colorbox{red!49}{(} \colorbox{red!39}{returning} \colorbox{red!28}{home} \colorbox{green!47}{aboard} the former holland \colorbox{green!26}{america} \colorbox{red!25}{liner}) \colorbox{red!25}{and} employing them as extra stokers for \colorbox{green!52}{her} \colorbox{red!21}{boilers}\colorbox{red!22}{.} \colorbox{red!22}{on} her next trip, \colorbox{red!18}{the} \colorbox{red!15}{veteran} transport loaded troops at saint\colorbox{green!68}{@}-@ nazaire\\\\
@ july, met his wife in new york\colorbox{red!25}{,} and \colorbox{red!20}{together} \colorbox{red!16}{they} traveled to \colorbox{red!24}{columbus}\colorbox{green!78}{,} georgia by way of washington, d. \colorbox{red!17}{c}. and atlanta\colorbox{green!65}{.}=\colorbox{red!22}{=} military schools== for the ten years following world war \colorbox{green!100}{i}, troy middleton would be either an instructor or a \colorbox{red!19}{student} in the succession \colorbox{green!60}{of} military \colorbox{red!25}{schools} that army officers attend during their careers. middleton \colorbox{green!32}{arrived} in columbus, \colorbox{red!23}{georgia} with strong praise \colorbox{red!21}{from} his superiors, and would soon get his efficiency report, in \colorbox{red!40}{which} \colorbox{red!100}{\textcolor{blue}{brigadier}} \colorbox{red!100}{general} \colorbox{red!35}{benjamin} \colorbox{red!27}{poor}e \colorbox{red!36}{of} the \colorbox{red!29}{4th} \colorbox{red!34}{division} \colorbox{red!36}{wrote} \colorbox{red!25}{of} \colorbox{red!24}{him}\colorbox{red!25}{,}\colorbox{red!29}{"} the best all@-@ around officer \colorbox{red!18}{i} have yet seen\colorbox{red!24}{.}< unk> by his rapid promotion from\\\\
 \colorbox{red!100}{coal} and 700 long \colorbox{red!15}{tons}( \colorbox{red!21}{710} t) of fuel oil and \colorbox{red!16}{that} provided her a \colorbox{red!21}{range} of 3@\colorbox{red!14}{,}@ 500 nautical miles( 6@,@ 500 km) at a speed of 10 knots( 19 km/ h)\colorbox{red!17}{.} her \colorbox{red!19}{main} \colorbox{red!32}{armament} consisted \colorbox{red!23}{of} a dozen obukhovskii 12@-@ \colorbox{red!29}{inch}\colorbox{red!16}{(} 305 \colorbox{red!15}{mm}\colorbox{red!24}{)} \colorbox{red!26}{pattern} 1907 \colorbox{red!24}{52}\colorbox{green!31}{@}-\colorbox{green!100}{@} \colorbox{red!100}{\textcolor{blue}{cal}}\colorbox{red!79}{ib}\colorbox{red!18}{re} \colorbox{red!71}{guns} mounted in four triple \colorbox{red!22}{turrets} distributed the length \colorbox{red!18}{of} the ship\colorbox{red!40}{.} the \colorbox{red!16}{russians} \colorbox{red!20}{did} not believe that super firing turrets offered any \colorbox{red!18}{advantage} \colorbox{red!12}{as} \colorbox{red!21}{they} discounted the value of axial fire and believed that super firing turrets could not fire while over the lower turret because of\\\\
\colorbox{red!30}{'} ll \colorbox{red!13}{still} be playing from 2007" and awarded it" \colorbox{green!40}{playstation} 3@-@ exclusive game of the year".\colorbox{green!25}{=} 11th battalion( australia)= the 11th battalion was an australian army battalion that was among the first infantry units raised during world war i for \colorbox{red!11}{the} \colorbox{red!15}{first} \colorbox{red!12}{australian} imperial force. it was the first battalion recruited in western australia, \colorbox{red!11}{and} following a brief \colorbox{green!20}{training} period in perth\colorbox{red!11}{,} the battalion sailed to egypt where it undertook four months of intensive training\colorbox{red!14}{.} in \colorbox{red!16}{april} 1915 it took part \colorbox{green!30}{in} the invasion of \colorbox{red!13}{the} gallip\colorbox{green!20}{oli} peninsula\colorbox{red!15}{,} landing at anzac cove\colorbox{red!14}{.} in \colorbox{red!15}{august} \colorbox{red!29}{1915} the \colorbox{red!22}{battalion} \colorbox{green!72}{was} \colorbox{red!100}{in} \colorbox{red!100}{\textcolor{blue}{action}} \colorbox{red!77}{in} \colorbox{red!24}{the} \colorbox{red!93}{battle} \colorbox{green!100}{of} \colorbox{red!18}{lone} \colorbox{red!17}{pine}\colorbox{red!18}{.} following\\\\
 \colorbox{red!25}{was} transferred to western \colorbox{red!10}{australia}\colorbox{red!10}{,} being attached to the 6th brigade, which was based around geraldton\colorbox{red!12}{.} \colorbox{red!10}{in} september 1942, as part \colorbox{red!9}{of} an army@-\colorbox{green!95}{@} wide reduction that came about because of over@-@ mobilisation\colorbox{red!15}{,} \colorbox{red!8}{the} battalion \colorbox{red!12}{was} amalgamated with the 14th battalion to become the 14th/ 32nd battalion( \colorbox{green!100}{pr}ahran/ footscray regiment)\colorbox{red!23}{.} \colorbox{red!24}{in} \colorbox{red!9}{early} \colorbox{red!24}{1943}\colorbox{red!17}{,} the 14th\colorbox{red!7}{/} 32nd battalion \colorbox{red!29}{carried} out \colorbox{red!100}{amphibious} \colorbox{red!100}{\textcolor{blue}{warfare}} \colorbox{red!89}{training} \colorbox{red!31}{in} \colorbox{red!21}{queensland} before being deployed to the bun\colorbox{red!9}{a} – gona area in new guinea in july\colorbox{red!14}{.} \colorbox{red!11}{the} \colorbox{red!9}{battalion} \colorbox{red!9}{would} remain in mainland new guinea \colorbox{red!9}{and} new britain for the next two years\colorbox{red!12}{,} under the command of lieutenant\\\\
 \colorbox{red!30}{in} an allied air raid on 10 december 1941\colorbox{red!8}{,} mickl \colorbox{red!12}{was} appointed to \colorbox{red!7}{temporarily} command the division\colorbox{red!8}{.} \colorbox{red!7}{during} december, mickl was wounded \colorbox{red!9}{in} \colorbox{red!9}{the} head and hand\colorbox{red!10}{,} but remained at his post\colorbox{red!10}{.} rom\colorbox{red!9}{mel} recommended \colorbox{red!7}{mick}l for the knight' s cross of the iron cross, for \colorbox{red!9}{his} leadership at sidi rezegh, and it was duly awarded on 13 december 1941\colorbox{red!12}{.} \colorbox{red!13}{the} harsh \colorbox{red!9}{conditions} of \colorbox{red!8}{desert} warfare had begun to affect mickl' s health, so at the end of december he was sent home on convalescent leave.=\colorbox{red!9}{=}\colorbox{red!8}{=} \colorbox{red!13}{eastern} \colorbox{red!23}{front}\colorbox{red!9}{=}\colorbox{red!8}{=}=\colorbox{red!9}{=}=== \colorbox{red!46}{12th} \colorbox{red!100}{rifle} \colorbox{red!100}{\textcolor{blue}{brigade}}===\colorbox{red!9}{=} \colorbox{green!85}{on} \colorbox{green!100}{25}\\\\
 \colorbox{red!21}{on} to bijeljina which was taken against light partisan resistance late on 16 march\colorbox{red!10}{.} \colorbox{red!9}{the} 27th \colorbox{green!100}{regiment} then consolidated its position in bi\colorbox{red!9}{je}ljina while \colorbox{red!11}{the} 28th regiment and \colorbox{red!8}{the} divisional reconnaissance battalion( german:< un\colorbox{red!11}{k}>) bore the brunt of the fighting as they advanced through< unk>, celic and koraj at the foot of the majevica mountains\colorbox{red!8}{.} sauberzweig later recorded \colorbox{red!15}{that} \colorbox{red!14}{the} \colorbox{red!28}{2nd} \colorbox{red!41}{battalion} \colorbox{red!100}{of} \colorbox{red!100}{\textcolor{blue}{the}} \colorbox{red!30}{28th} \colorbox{red!51}{regiment}\colorbox{red!33}{(} \colorbox{red!29}{ii}\colorbox{red!22}{/} \colorbox{red!9}{28}\colorbox{red!12}{)}\colorbox{red!23}{"} \colorbox{red!17}{at} ce\colorbox{red!9}{lic} \colorbox{red!20}{stormed} the partisan \colorbox{red!10}{defenses} with( new) battalion commander hans hanke at the poin t" \colorbox{red!9}{and} that \colorbox{red!8}{enemy} forces \colorbox{red!8}{withdrew} after\\\\
 \colorbox{red!13}{of} matthews\colorbox{green!100}{,} the company 2ic\colorbox{red!5}{,} who had taken \colorbox{red!6}{command} almost immediately after the company commander \colorbox{red!7}{was} \colorbox{red!9}{wounded}. under his command\colorbox{red!7}{,} each of the platoons assaulted \colorbox{red!6}{a} different cluster of buildings to which \colorbox{green!100}{they} had been assigned during training \colorbox{red!5}{on} the \colorbox{green!88}{replica} village at hastings. the west side boys' \colorbox{red!8}{ammunition} store was found and secured and, once the rest of the \colorbox{red!5}{buildings} had been \colorbox{red!4}{cleared}, the paras took \colorbox{red!8}{up} defensive \colorbox{red!9}{positions} to \colorbox{red!6}{block} any potential counter@-@ \colorbox{green!83}{attack} and patrols went into \colorbox{red!5}{the} immediate jungle in search of any west side boys \colorbox{red!4}{hiding} in the \colorbox{red!6}{bushes}\colorbox{red!5}{.} \colorbox{red!7}{the} village was completely secure by \colorbox{red!7}{08}: \colorbox{red!6}{00} and \colorbox{red!5}{the} paras \colorbox{red!7}{secured} the approaches \colorbox{red!16}{with} \colorbox{red!70}{\textcolor{blue}{clay}}\colorbox{red!100}{more}\\\\
\colorbox{red!63}{)}, increased her metacentric \colorbox{red!19}{height} to 6@.@ 3 feet( 1\colorbox{green!100}{@}.@ 9 m) at deep load, and all of the changes to her equipment increased her \colorbox{red!18}{crew} to \colorbox{red!17}{a} total of 1@,@ \colorbox{red!19}{188}. \colorbox{red!17}{despite} the \colorbox{red!16}{bulge}s she was able to reach a speed \colorbox{red!20}{of} 21@.@ 75 knots\colorbox{red!15}{(} 40@.@ 28 \colorbox{red!16}{km}/ h; 25@.@ 03 mph). \colorbox{red!17}{a} brief \colorbox{red!27}{refit} in early 1927 \colorbox{red!20}{saw} the \colorbox{red!32}{addition} of two more \colorbox{red!30}{four}@-@ \colorbox{red!41}{inch} \colorbox{red!100}{aa} \colorbox{red!100}{\textcolor{blue}{guns}} \colorbox{red!55}{and} \colorbox{red!31}{the} \colorbox{red!23}{removal} \colorbox{red!16}{of} the six@-@ inch guns \colorbox{red!14}{from} \colorbox{red!26}{the} \colorbox{red!19}{shelter} deck\colorbox{red!40}{.} about 1931\colorbox{red!20}{,} \colorbox{red!18}{a} high@-@ angle \colorbox{red!24}{control}\\\\
 became enraged \colorbox{red!7}{at} him, slapping him \colorbox{red!10}{across} the face. he began yelling:" your nerves, hell, you are just a goddamned coward\colorbox{red!7}{.} shut up that goddamned \colorbox{red!7}{crying}. i won' t have \colorbox{green!31}{these} brave men who have been shot at seeing this yellow bastard sitting here crying." patton then \colorbox{red!15}{reportedly} \colorbox{red!9}{slapped} bennett again, knocking his helmet liner off, and \colorbox{red!20}{ordered} \colorbox{green!82}{the} \colorbox{green!100}{receiving} \colorbox{red!31}{officer}\colorbox{red!21}{,} \colorbox{red!100}{\textcolor{blue}{major}} \colorbox{red!100}{charles} \colorbox{red!62}{b}\colorbox{red!21}{.} et\colorbox{red!14}{ter}\colorbox{red!17}{,} \colorbox{red!6}{not} \colorbox{red!10}{to} admit \colorbox{red!7}{him}. \colorbox{red!6}{patton} \colorbox{red!7}{then} \colorbox{red!12}{threatened} bennett,\colorbox{red!6}{"} you' re going back \colorbox{red!6}{to} the front lines and you \colorbox{red!8}{may} get shot and killed, but you' \colorbox{red!7}{re} going to fight. if \colorbox{red!10}{you} \colorbox{green!34}{don}' t, i'\\\\
 \colorbox{red!36}{secondary} guns, two of which were disabled. the ammunition stores \colorbox{red!12}{for} these two guns were set on \colorbox{red!10}{fire} and the magazines had to be flooded to prevent an explosion. the ship nevertheless \colorbox{red!15}{remained} combat \colorbox{red!10}{effective}, as her primary \colorbox{red!12}{battery} remained in operation\colorbox{red!10}{,} as \colorbox{red!11}{did} most of her secondary guns; ko\colorbox{red!16}{nig} \colorbox{green!39}{could} also steam \colorbox{red!14}{at} close to her maximum speed\colorbox{red!12}{.} other areas of the ship had to be counter@-@ \colorbox{red!14}{flooded} to maintain stability; 1@,@ 600 tons of water entered \colorbox{red!15}{the} ship\colorbox{red!13}{,} either \colorbox{green!67}{as} \colorbox{red!13}{a} \colorbox{green!100}{result} \colorbox{red!35}{of} \colorbox{red!100}{\textcolor{blue}{battle}} \colorbox{red!100}{damage} \colorbox{red!15}{or} counter\colorbox{red!12}{@}-@ flooding \colorbox{green!34}{efforts}. \colorbox{red!17}{the} \colorbox{red!18}{flooding} \colorbox{red!9}{rendered} the \colorbox{red!18}{battleship} sufficiently low in the water \colorbox{red!13}{to} prevent the ship from \colorbox{red!16}{being} able\\\\
 \colorbox{red!34}{in} 1924 and \colorbox{red!14}{rice} institute, houston, \colorbox{red!17}{texas} in 1928\colorbox{red!15}{.} he dropped out of graduate school after one year and decided to hitchhike to san francisco. \colorbox{red!21}{the} lack of work meant hunger, so he \colorbox{green!32}{chose} to join the united states army' s 11th cavalry regiment as a private on july 30, 1930, \colorbox{red!13}{serving} in monterey, california. after a year in the horse cavalry, par\colorbox{red!21}{rish} became an \colorbox{green!58}{aviation} cadet in june 1931 and subsequently qua lified as an enlisted pilot\colorbox{red!25}{.} \colorbox{red!16}{he} completed \colorbox{green!42}{flight} \colorbox{red!16}{training} in 1932 and \colorbox{red!21}{was} assigned to \colorbox{green!100}{the} 13th \colorbox{red!41}{attack} \colorbox{red!31}{squadron} \colorbox{red!100}{at} \colorbox{red!100}{\textcolor{blue}{fort}} \colorbox{red!52}{cr}\colorbox{red!87}{ock}\colorbox{red!87}{ett}\colorbox{red!31}{,} \colorbox{red!50}{near} \colorbox{red!16}{galveston}\colorbox{red!31}{,} \colorbox{red!38}{texas}\colorbox{red!48}{.} one year \colorbox{red!20}{later} in \colorbox{red!23}{september} 1933 parrish\\\\
 during the battle\colorbox{green!97}{,} murray was awarded the victoria cross. \colorbox{red!4}{soon} after his victoria cross action, he was promoted to major and earned a bar to \colorbox{red!8}{his} distinguished service order during an attack on the hindenburg line near bullec\colorbox{green!72}{our}t. promoted to lieutenant colonel in early \colorbox{red!6}{1918}, \colorbox{red!10}{he} \colorbox{red!20}{assumed} \colorbox{red!8}{command} \colorbox{red!10}{of} the \colorbox{red!50}{4th} \colorbox{red!17}{machine} \colorbox{red!100}{gun} \colorbox{red!100}{\textcolor{blue}{battalion}}\colorbox{red!35}{,} \colorbox{red!30}{where} he would \colorbox{red!10}{remain} until \colorbox{red!6}{the} end \colorbox{red!5}{of} \colorbox{red!5}{the} war\colorbox{red!15}{.} \colorbox{red!11}{returning} to australia in 1920, murray eventually \colorbox{red!10}{settled} in \colorbox{red!7}{queensland}, where he purchased the \colorbox{green!100}{grazing} \colorbox{red!10}{farm} \colorbox{red!7}{that} would be his home \colorbox{red!6}{for} the remainder of his life\colorbox{red!5}{.} re@-@ \colorbox{green!70}{enlist}ing for service \colorbox{red!6}{in} the second world war, he was appo inted as commanding officer\\\\
 \colorbox{red!70}{10} officers and 315 \colorbox{red!12}{enlisted} men, plus an additional four officers and 19 enlisted men if serving as a flotilla \colorbox{red!15}{flagship}\colorbox{red!14}{.}== construction and career=\colorbox{green!100}{=} the \colorbox{red!12}{ship} \colorbox{red!11}{was} ordered on 7 july 1934 and laid \colorbox{red!11}{down} at deutsche werke, kiel, on 2 january 1935 as yard number< unk>. she \colorbox{red!11}{was} launched on 30 \colorbox{red!11}{november} 1935 and completed on 8 \colorbox{red!12}{april} 1937\colorbox{red!10}{.} \colorbox{red!10}{she} was \colorbox{red!18}{named} after max \colorbox{red!13}{schultz} who \colorbox{red!29}{commanded} the torpedo boat< unk\colorbox{red!13}{>} \colorbox{red!27}{and} \colorbox{green!54}{was} \colorbox{red!51}{killed} \colorbox{red!100}{in} \colorbox{red!100}{\textcolor{blue}{action}} \colorbox{red!49}{in} \colorbox{red!24}{january} \colorbox{red!57}{1917}\colorbox{green!65}{.} \colorbox{red!17}{ko}r\colorbox{red!12}{vet}ten\colorbox{red!16}{ka}pitan martin\colorbox{red!15}{<} unk> was appointed as her first captain. max schultz was assigned to the 1st destroyer division on 26\\\\
 \colorbox{red!41}{the} command of otto von diederichs. the squadron participated in the fall maneuvers in \colorbox{red!26}{1894}, which simulated a two@\colorbox{green!44}{-}@ front war against france and \colorbox{red!15}{russia}; deutschland' s squadron acted as the russian fleet during the exercises. between 1894 and 1897, deutschland was \colorbox{red!16}{rebuilt} in the imperial dockyard in wilhelmshaven. the \colorbox{red!16}{ship} was \colorbox{red!23}{converted} \colorbox{red!21}{into} \colorbox{green!100}{an} armored cruiser\colorbox{red!26}{;} her \colorbox{red!100}{heavy} \colorbox{red!100}{\textcolor{blue}{guns}} \colorbox{red!57}{were} \colorbox{red!68}{removed} \colorbox{red!29}{and} \colorbox{red!39}{replaced} with lighter \colorbox{red!32}{weapons}, \colorbox{red!16}{including} eight 15 cm( \colorbox{red!18}{5}@.@ 9 in\colorbox{red!17}{)} and eight 8@.\colorbox{green!100}{@} 8 \colorbox{red!18}{cm}( 3@\colorbox{green!42}{.}@ 5 in\colorbox{red!24}{)} \colorbox{red!28}{guns}\colorbox{red!21}{.} \colorbox{red!14}{her} entire rigging equipment \colorbox{red!22}{was} removed and two heavy military \colorbox{green!89}{mast}s \colorbox{red!22}{were} installed\\\\
 called on many times to maintain order \colorbox{red!20}{in} times of disaster and to keep peace during periods of political unrest\colorbox{red!17}{.} oklahoma governor john c. walton used division \colorbox{green!62}{troops} \colorbox{green!37}{to} prevent the \colorbox{red!22}{state} legislature \colorbox{red!14}{from} \colorbox{red!17}{meeting} when they were preparing to impeach him in \colorbox{red!14}{1923}. governor william h. murray called out the guard several times during the depression to close banks, distribute food \colorbox{red!12}{and} once to force the state of texas to keep open \colorbox{red!17}{a} free bridge over the red river which texas intended to collect toll\colorbox{red!16}{s} for, even after federal \colorbox{green!68}{courts} ordered the bridge not be \colorbox{red!17}{opened}\colorbox{red!41}{.} \colorbox{red!31}{the} \colorbox{red!27}{division} \colorbox{red!25}{would} \colorbox{red!23}{go} \colorbox{red!42}{on} \colorbox{red!22}{to} \colorbox{green!100}{see} \colorbox{red!100}{\textcolor{blue}{combat}} \colorbox{red!100}{in} \colorbox{red!32}{world} \colorbox{red!46}{war} \colorbox{red!16}{ii} as \colorbox{red!19}{one} of \colorbox{red!24}{four} national \colorbox{red!16}{guard} \colorbox{red!29}{divisions} active during\\\\
\newline \newline
 {\bf Transformer factor 170 in layer 10 with saliency map \newline Explaination: topic: music production }  \newline \newline
 2nd street tunnel and part of downtown los \colorbox{red!7}{angeles} spread \colorbox{green!65}{out} over a 48\colorbox{green!65}{@}-\colorbox{green!71}{@} hour period. kesha explained the idea behind the \colorbox{green!100}{video} as well \colorbox{red!8}{as} the experience during \colorbox{green!75}{an} interview with \colorbox{red!14}{mtv} \colorbox{red!8}{news}; she said that the video was different from \colorbox{red!10}{her} other videos, noting \colorbox{red!9}{that} it was \colorbox{red!11}{going} to show a sexier side of herself. \colorbox{red!8}{the} \colorbox{red!49}{music} \colorbox{red!42}{video} \colorbox{red!100}{for}\colorbox{red!100}{\textcolor{blue}{"}} \colorbox{red!18}{we} \colorbox{red!30}{r} \colorbox{red!26}{who} \colorbox{red!11}{we} \colorbox{red!19}{r}\colorbox{red!77}{"} \colorbox{red!35}{is} \colorbox{red!10}{presented} as an underground \colorbox{red!10}{party}. \colorbox{red!9}{the} \colorbox{red!14}{video} \colorbox{red!14}{starts} off with futuristic flashing lights. kesha, seen in a ponytail wearing gray and black makeup, chains\colorbox{red!9}{,} ripped stockings, and a sparkly one@-@ piece leotard made of shards of broken\\\\
 \colorbox{red!43}{and} several european territories), her" endless love" \colorbox{red!11}{duet} with luther \colorbox{red!9}{van}dro\colorbox{red!10}{ss}\colorbox{red!15}{(} number@-@ one in new zealand) \colorbox{red!12}{and}\colorbox{red!18}{"} \colorbox{red!16}{against} \colorbox{red!13}{all} odds\colorbox{red!18}{"} \colorbox{red!65}{featuring} \colorbox{red!100}{west}\colorbox{red!100}{\textcolor{blue}{life}}\colorbox{red!19}{(} \colorbox{red!18}{number}\colorbox{green!100}{@}-@ \colorbox{red!11}{one} in \colorbox{red!11}{the} \colorbox{red!12}{united} \colorbox{red!12}{kingdom})\colorbox{red!12}{.}" \colorbox{red!10}{thank} god i found you" was also omitted \colorbox{green!58}{from} the japanese \colorbox{red!12}{track} listing, and replaced with" all i want for \colorbox{green!73}{christmas} is you\colorbox{red!12}{"}. \colorbox{green!49}{for} the album artwork, carey launched a social media campaign on april 12, \colorbox{red!9}{2015}, whereby fans had to share a link to her website \colorbox{green!57}{in} order to reveal the cover which was concealed by a curtain\colorbox{red!15}{.} using the hashtag"< unk>",\\\\
 \colorbox{red!34}{single}," we belong together". he contained to add" but still\colorbox{red!14}{,} if mimi ’ s going to mine from her own extensive back catalog of \colorbox{red!7}{ballads}\colorbox{red!10}{,} those \colorbox{red!6}{are} the primo melodies to \colorbox{red!10}{go} for." a reviewer for \colorbox{red!10}{dj} booth \colorbox{green!100}{thought} that minaj" ruined" the song\colorbox{red!14}{.}\colorbox{green!78}{=}== music video=== the \colorbox{red!8}{accompanying} music \colorbox{red!19}{video} \colorbox{red!37}{for} \colorbox{red!28}{the} \colorbox{red!54}{remix} \colorbox{red!100}{of}\colorbox{red!100}{\textcolor{blue}{"}} \colorbox{red!20}{up} \colorbox{red!35}{out} \colorbox{red!26}{my} \colorbox{red!18}{f} ace\colorbox{red!72}{"} \colorbox{red!50}{was} \colorbox{red!13}{directed} \colorbox{red!10}{by} \colorbox{red!9}{carey}' s husband, \colorbox{red!14}{nick} cannon. minaj spoke about filming a video with carey and \colorbox{red!9}{how} she \colorbox{red!8}{did} not believe that the video would ever be released:" i didn ’ t even tell anyone \colorbox{red!9}{i} shot a video with\\\\
 \colorbox{red!10}{the} producer, few days after he \colorbox{red!10}{had} finished the \colorbox{red!15}{composition}\colorbox{red!8}{,} \colorbox{red!25}{madonna} \colorbox{red!10}{completed} writing \colorbox{red!17}{the} \colorbox{red!48}{lyrics} \colorbox{red!100}{of}\colorbox{red!100}{\textcolor{blue}{"}} \colorbox{red!32}{i} \colorbox{red!25}{don}\colorbox{red!26}{'} \colorbox{red!23}{t} \colorbox{green!53}{give} a\colorbox{red!65}{"}. solve\colorbox{red!14}{ig} understood \colorbox{red!13}{that} the \colorbox{red!11}{lyrics} were \colorbox{green!36}{probable} \colorbox{red!10}{references} towards madonna' s life and \colorbox{green!93}{thus} received coverage in the press. \colorbox{red!9}{however}, \colorbox{red!13}{he} was not aware of the inner meaning behind the lyrics. with billboard magazine\colorbox{red!7}{,} the producer further explained: at first i thought we were going to work on one song; that was the original \colorbox{green!63}{plan}. let' s try to work on one song and take it from there-- not spend too much \colorbox{red!11}{time} thinking \colorbox{green!100}{about} the l egend, and do something that \colorbox{red!10}{just} makes sense\colorbox{red!9}{.}\\\\
 \colorbox{red!19}{provided} an additional \colorbox{red!13}{and} assistant \colorbox{red!16}{engineering}. all the instruments were provided by eriksen and hermansen while \colorbox{green!100}{dean} sang the background vocals. in may \colorbox{green!64}{2011}, in the mix review, an analyzing commercial productions, mike senior \colorbox{green!83}{of} sound on sound \colorbox{red!14}{revisited} the original mixing \colorbox{green!95}{of} \colorbox{red!17}{the} \colorbox{red!15}{song}. according to him, before he started \colorbox{red!13}{the} mix, \colorbox{red!13}{senior} played the song a couple of times before releasing what thing about it" bugged" him. \colorbox{green!91}{working} it out, he noted that the harmony \colorbox{green!66}{of} the \colorbox{red!23}{mix} is undermine\colorbox{red!15}{d} by the \colorbox{red!17}{kick} drum\colorbox{red!41}{.}\colorbox{red!100}{\textcolor{blue}{"}} \colorbox{red!15}{what}\colorbox{red!31}{'} \colorbox{red!19}{s} \colorbox{red!29}{my} name\colorbox{red!28}{?}\colorbox{red!100}{"} \colorbox{red!86}{contains} \colorbox{red!17}{basic} \colorbox{red!34}{harmonies} that \colorbox{red!22}{are} a bar of f minor\colorbox{red!20}{,} a bar of a major\\\\
 \colorbox{red!44}{practiced} \colorbox{red!13}{in} \colorbox{red!13}{their} backyards and \colorbox{red!15}{at}< \colorbox{red!17}{un}\colorbox{red!16}{k}\colorbox{green!40}{>} salon, owned by \colorbox{red!18}{knowles}' s mother, tina\colorbox{red!20}{.} the group would test routines \colorbox{red!18}{in} the salon, when it was on montrose boulevard in houston, \colorbox{red!19}{and} sometimes would collect tips from the customers. their try out would be critiqued by the people \colorbox{red!16}{inside}. during their school days, girl' s ty\colorbox{green!35}{me} performed at local gigs. when summer came, mathew knowles established a" boot camp" to train them in dance and \colorbox{red!19}{vocal} \colorbox{red!16}{lessons}. after rigorous training, they \colorbox{red!18}{began} \colorbox{red!17}{performing} \colorbox{red!17}{as} \colorbox{red!27}{opening} \colorbox{red!39}{acts} \colorbox{red!65}{for} established \colorbox{red!66}{r}\colorbox{red!82}{\&} \colorbox{red!100}{b} \colorbox{red!100}{\textcolor{blue}{groups}} \colorbox{green!100}{of} \colorbox{red!15}{that} time \colorbox{red!21}{such} as swv, dr\colorbox{red!14}{u} hill \colorbox{red!16}{and} immature. tina\\\\
 \colorbox{red!11}{day} reception at the greek embassy. \colorbox{red!13}{upon} \colorbox{red!8}{return} to greece, she was \colorbox{green!60}{greeted} at the airport by fans along \colorbox{green!90}{with} \colorbox{red!24}{the} \colorbox{red!67}{music} \colorbox{red!28}{video} \colorbox{red!100}{of}\colorbox{red!100}{\textcolor{blue}{"}} \colorbox{red!27}{my} \colorbox{red!41}{number} \colorbox{red!20}{one}\colorbox{red!83}{"} \colorbox{red!35}{playing} \colorbox{red!32}{on} \colorbox{red!13}{the} \colorbox{red!12}{video} monitors. while in \colorbox{red!9}{greece}, she \colorbox{red!13}{attended} the \colorbox{red!8}{opening} \colorbox{red!8}{ceremony} \colorbox{red!11}{of} the european final four for the volleyball champions league in< unk>, \colorbox{red!10}{where} her song was \colorbox{green!100}{played} \colorbox{red!9}{as} she appeared on stage with cheerleaders\colorbox{red!9}{.} on \colorbox{green!89}{march} 29, paparizou arrived in valletta, malta \colorbox{red!9}{where} she signed autographs, appeared on television stations, and gave interviews to the local media. following malta, she traveled to serbia and \colorbox{red!8}{montenegro} where she gave additional interviews before moving on to and\\\\
 \colorbox{red!74}{and} \colorbox{red!15}{her} low hip@\colorbox{red!14}{-}@ \colorbox{red!14}{grind} during' rude boy' were the smash hits of her body language\colorbox{red!15}{.}" deborah linton of city life wrote that rihanna\colorbox{red!15}{"} even \colorbox{red!14}{manages} to make a psychiatric couch \colorbox{red!12}{look} sexy". linton called the show' s stage sets \colorbox{red!19}{impressive} and imaginative. rick massimo of the providence journal wrote that \colorbox{red!20}{rihanna}" looked like a neon@-@ sign rendition of herself during' rehab', rarely addressed the audience, and didn\colorbox{red!17}{'} t rise above flat cliche in that until the very end of the \colorbox{red!17}{show}\colorbox{red!19}{"}\colorbox{red!41}{.}\colorbox{red!22}{"} rehab" \colorbox{red!35}{and} \colorbox{red!100}{rihanna}\colorbox{red!26}{'} \colorbox{red!70}{s} \colorbox{red!99}{2009} \colorbox{red!100}{\textcolor{blue}{single}}\colorbox{red!52}{"} \colorbox{green!100}{russian} \colorbox{red!20}{ro}ule\colorbox{green!50}{tte}\colorbox{red!18}{"} \colorbox{red!41}{were} \colorbox{red!17}{excluded} \colorbox{red!23}{from} \colorbox{red!15}{the} set\\\\
 \colorbox{red!16}{only} a \colorbox{red!20}{few} hours. he said\colorbox{red!11}{:}" there were a lot of \colorbox{red!13}{tracks}, but i \colorbox{red!16}{just} enjoyed it, to be honest. i knew how i wanted it \colorbox{red!17}{to} sound, and it was pretty much the last song we cut; a \colorbox{red!13}{lot} of the mixing was nailed in the production as well, \colorbox{red!13}{which} helped. dream did a great job producing this track\colorbox{red!12}{.}" \colorbox{red!19}{the} bar one \colorbox{red!30}{guitar} \colorbox{red!15}{track} \colorbox{red!100}{of}\colorbox{red!100}{\textcolor{blue}{"}} school\colorbox{red!45}{in}\colorbox{red!34}{'} \colorbox{red!13}{life}\colorbox{red!91}{"} \colorbox{red!22}{was} entirely programmed. similarly, \colorbox{red!11}{the} \colorbox{red!27}{live} drum \colorbox{red!13}{section} in the \colorbox{red!13}{hook} was actually done with programmed drums\colorbox{red!11}{.} \colorbox{red!12}{once} the mixing was \colorbox{green!100}{over}, swivel' \colorbox{red!20}{s} impression were \colorbox{red!13}{as} follows\colorbox{red!10}{:}\colorbox{green!90}{[}' schoolin' life] absolutely\\\\
 \colorbox{red!25}{tour} began on \colorbox{red!10}{march} 1, 2000 \colorbox{red!10}{at} the house \colorbox{red!9}{of} blues in los angeles\colorbox{red!10}{,} while other venues included paris olympia, trump taj mahal, brixton academy, the montreux jazz festival, and the essence jazz festival in new orleans. \colorbox{red!10}{by} july, the tour' \colorbox{green!65}{s} first half had sold out in each city. the tour \colorbox{red!13}{lasted} nearly \colorbox{green!67}{eight} months, \colorbox{red!11}{w}hil e performances went \colorbox{red!9}{for} up to three hours a night\colorbox{red!7}{.} the voodoo tour was taken internationally, with \colorbox{red!10}{one} of the most notable performances being the free \colorbox{green!100}{jazz} festival \colorbox{red!7}{in} brazil\colorbox{red!10}{.} \colorbox{red!34}{the} \colorbox{red!46}{music} \colorbox{red!68}{video} \colorbox{red!100}{for}\colorbox{red!100}{\textcolor{blue}{"}} \colorbox{red!20}{untitled}\colorbox{red!19}{(} how \colorbox{red!12}{does} \colorbox{red!12}{it} \colorbox{red!13}{feel}\colorbox{red!17}{)}\colorbox{red!40}{"} portrayed d' angelo as \colorbox{red!10}{a} \colorbox{red!10}{sex} symbol\\\\
 ho\colorbox{green!69}{bson} noted that \colorbox{red!18}{rihanna}" rejects the victim \colorbox{red!10}{stance}\colorbox{red!17}{"} \colorbox{red!13}{in} \colorbox{red!21}{the} \colorbox{red!97}{video} \colorbox{red!100}{for}\colorbox{red!100}{\textcolor{blue}{"}} man \colorbox{red!28}{down}\colorbox{red!39}{"}, \colorbox{green!100}{and} elucidated that she played the role of a rape survivor who shot \colorbox{red!10}{her} attacker. she attributed the \colorbox{green!34}{location} of shooting the video \colorbox{red!8}{in} jamaica as significant, due to how the image of a gun proliferated \colorbox{red!9}{during} 1990s jamaican dance \colorbox{red!9}{hall}\colorbox{red!9}{'} s to" express \colorbox{green!90}{female} rage". \colorbox{red!12}{the} prologue \colorbox{red!8}{depicts} rihanna \colorbox{green!91}{as} a" \colorbox{green!92}{dark}\colorbox{red!9}{@}-@ hooded" femme fatale whereby the narrative explains her motives for murder and provokes the spectator to sympathi\colorbox{red!9}{ze} with her because she danced in \colorbox{red!7}{a} \colorbox{green!47}{provocative} manner with a \colorbox{green!80}{man} in a \colorbox{red!9}{club}\colorbox{red!9}{,} which\\\\
 \colorbox{red!20}{had} a deep impact \colorbox{red!12}{on} delonge in that he spent a \colorbox{red!14}{night} \colorbox{red!12}{up} crying for him when he wrote \colorbox{red!23}{the} \colorbox{red!56}{track}\colorbox{red!19}{.}\colorbox{red!100}{\textcolor{blue}{"}} a \colorbox{red!31}{little}\colorbox{red!20}{'} \colorbox{red!35}{s} \colorbox{red!23}{enough}\colorbox{red!100}{"} \colorbox{red!54}{was} \colorbox{red!37}{inspired} \colorbox{red!26}{by} a religious concept in \colorbox{red!10}{which} a god came to bring positive change on earth when it faces terrorism, war or famine.\colorbox{green!100}{"} the war\colorbox{red!12}{"}\colorbox{red!19}{,} an \colorbox{red!22}{anthem} about the iraq war \colorbox{green!82}{and} \colorbox{green!73}{its} death toll\colorbox{red!14}{,} is succeeded by\colorbox{red!11}{"} it hurts", \colorbox{red!12}{a} \colorbox{green!69}{track} about a friend of delonge with a \colorbox{red!12}{cheating} girlfriend." it' s \colorbox{red!10}{a} terrible situation where my friend \colorbox{red!6}{is} being crushed from the inside out by all the manipulative stuff she' s doing and this song' s\\\\
 \colorbox{red!43}{just} \colorbox{red!16}{took} \colorbox{green!81}{that} dress out of the storage \colorbox{red!12}{–} it has a 27@-@ foot train and it was just all hand@-@ beaded and stuff and so i figured we might as \colorbox{red!10}{well} get a use out of it.'=== synopsis=\colorbox{green!95}{=}= the video \colorbox{red!17}{features} \colorbox{red!17}{carey} \colorbox{red!13}{ready}ing for her wedding, and follows her to the altar, as \colorbox{red!13}{well} as her escape from the reception. many of the \colorbox{green!100}{actors} \colorbox{red!13}{featured} \colorbox{green!72}{in} \colorbox{red!17}{carey}' s" it' s like \colorbox{red!14}{that}" \colorbox{red!36}{video} \colorbox{red!18}{were} \colorbox{green!62}{in} that \colorbox{red!59}{of}\colorbox{red!100}{"} \colorbox{red!35}{we} \colorbox{red!29}{belong} \colorbox{red!34}{together}\colorbox{red!100}{\textcolor{blue}{"}}\colorbox{red!31}{,} \colorbox{red!33}{which} \colorbox{red!18}{was} shot \colorbox{red!11}{as} a continuation \colorbox{red!13}{from} the" it' s like that" \colorbox{red!15}{video}. it begins with\\\\
 \colorbox{red!46}{3} in dutch@-@ speaking flanders and \colorbox{red!6}{number} 2 in french@-@ speaking wallonia\colorbox{red!8}{.} it was certified gold by \colorbox{red!7}{the} \colorbox{red!6}{belgian} \colorbox{red!7}{entertainment} association( bea) for selling more than 15@,@ 000 \colorbox{green!48}{copies}. although the song spent only 1 week on the italian singles chart\colorbox{red!10}{(} at number 8\colorbox{red!7}{)}, \colorbox{red!7}{it} was certified platinum by the federazione industria musicale italiana( fimi) in \colorbox{red!6}{2014} for selling more than 30@,@ 000 \colorbox{green!40}{copies}.== music \colorbox{red!6}{video}===== background and synopsis=== anthony \colorbox{green!94}{man}\colorbox{red!6}{dler} \colorbox{green!100}{directed} \colorbox{red!19}{the} \colorbox{red!21}{music} \colorbox{red!50}{video} \colorbox{red!100}{for}\colorbox{red!100}{\textcolor{blue}{"}} \colorbox{red!19}{man} \colorbox{red!32}{down}\colorbox{red!58}{"} \colorbox{red!12}{in} \colorbox{red!19}{april} \colorbox{red!9}{2011} \colorbox{red!8}{on} \colorbox{red!15}{a} \colorbox{red!14}{beach} in\\\\
 \colorbox{red!43}{at} numbers 18 and \colorbox{green!41}{43} in \colorbox{red!10}{the} united states, and experienced moderate \colorbox{green!76}{success} \colorbox{red!8}{worldwide}\colorbox{red!12}{.} unlike her previous records, spears did \colorbox{red!16}{not} heavily promote \colorbox{red!11}{blackout}; \colorbox{red!10}{her} only televised \colorbox{red!11}{appearance} for blackout was a \colorbox{green!100}{universally}\colorbox{red!12}{@}-@ \colorbox{red!11}{pan}\colorbox{red!21}{ned} \colorbox{red!75}{performance} \colorbox{red!100}{of}\colorbox{red!100}{\textcolor{blue}{"}} \colorbox{red!36}{gi}\colorbox{red!61}{mme} \colorbox{red!53}{more}\colorbox{red!83}{"} \colorbox{red!37}{at} \colorbox{red!32}{the} 2007 \colorbox{red!12}{mtv} video music awards.== background and development== in \colorbox{green!40}{november} 2003, while promoting her fourth studio album in the zone, spears told entertainment weekly \colorbox{red!10}{that} she was already writing songs for her next album and \colorbox{red!8}{was} also hoping to start \colorbox{red!9}{her} own record label in 2004. henrik jonback confirmed that he had \colorbox{green!60}{written} songs with her during the european leg \colorbox{red!16}{of} the onyx hotel tour,"\\\\
 \colorbox{red!52}{of} albums also had increased sales due to \colorbox{green!74}{discount}ing and publicity generated by \colorbox{red!19}{the} single and her performance. billboard estimated that her top@-@ 10 digital sales collectively increased over 1@,@ 700 percent\colorbox{red!22}{.} \colorbox{red!19}{madonna}' s bestsell\colorbox{red!17}{ing} album was \colorbox{red!18}{the} 2009 greatest@-@ \colorbox{red!17}{hits} collection, celebration, which sold 16@,@ 000 \colorbox{green!78}{copies}( up 1@,@ 341 percent) and reentered the billboard \colorbox{red!14}{200} album chart. the following week \colorbox{red!16}{celebration} fell 105 spots \colorbox{green!69}{on} the \colorbox{red!14}{chart} to number 157, with sales falling to \colorbox{green!72}{4}@\colorbox{green!100}{,}@ \colorbox{red!26}{000} \colorbox{red!24}{copies}\colorbox{red!100}{.}\colorbox{red!100}{\textcolor{blue}{"}} \colorbox{red!84}{give} \colorbox{red!71}{me} \colorbox{red!32}{all} your lu\colorbox{red!23}{vin}\colorbox{red!26}{'}\colorbox{red!42}{"} \colorbox{red!59}{fell} to \colorbox{red!31}{number} \colorbox{red!17}{39} \colorbox{red!45}{on} \colorbox{red!20}{the} hot\\\\
 \colorbox{red!26}{opened} the performance with" yeah 3x" \colorbox{green!42}{and} was dressed in a white formal suit, accompanied by" full@-@ skirted dancers". brown was eventually joined onstage by tuxedo@-@ clad dancers and began dancing to \colorbox{red!20}{the} 1993 wu\colorbox{red!17}{@}-@ tang clan \colorbox{red!19}{single}" protect ya neck". his dance routine \colorbox{red!16}{then} \colorbox{red!16}{moved} into \colorbox{red!19}{1991}, \colorbox{red!17}{where} \colorbox{red!12}{he} \colorbox{red!46}{danced} \colorbox{red!44}{to} \colorbox{red!63}{nirvana}\colorbox{red!41}{'} \colorbox{red!100}{s}\colorbox{red!100}{\textcolor{blue}{"}} \colorbox{red!51}{smells} \colorbox{red!36}{like} \colorbox{red!21}{teen} \colorbox{green!100}{spirit}\colorbox{red!64}{"}\colorbox{red!10}{.} \colorbox{red!13}{brown}' s performance then came back to the future\colorbox{red!15}{,} where he began to \colorbox{red!11}{sing}" beautiful people\colorbox{green!33}{"}. \colorbox{red!11}{while} performing the song, he was \colorbox{green!56}{suspended} in the air, and then lowered to another stage \colorbox{red!14}{where} he \colorbox{red!10}{continued} to\\\\
 \colorbox{red!38}{register} that she didn' t know she had." from the moment she \colorbox{red!12}{was} signed in the film\colorbox{red!9}{,} \colorbox{red!17}{madonna} \colorbox{red!13}{had} \colorbox{red!11}{expressed} interest in \colorbox{red!23}{recording} a \colorbox{red!32}{dance} \colorbox{red!69}{version} \colorbox{red!100}{of}\colorbox{red!100}{\textcolor{blue}{"}} \colorbox{red!37}{don}\colorbox{red!17}{'} \colorbox{red!14}{t} \colorbox{red!14}{cry} \colorbox{red!10}{for} \colorbox{red!33}{me} \colorbox{green!99}{argentina}\colorbox{red!51}{"}\colorbox{red!12}{.} according to her \colorbox{green!100}{public}ist liz rosenberg," since she didn' t write the music and \colorbox{green!47}{lyrics}, she wanted \colorbox{red!13}{her} signature on that song... i think on her mind, the \colorbox{green!43}{best} \colorbox{green!68}{way} to do it was go in the studio and \colorbox{green!59}{work} up a remix". for this, \colorbox{red!12}{in} august 1996\colorbox{red!11}{,} while still mixing the film' s soundtrack, madonna hired remixers pablo \colorbox{red!12}{flores} and javier garza. according to \colorbox{red!11}{flores}, the singer\\\\
 \colorbox{red!5}{d} accumulated until then but that was instead an ideal marriage of \colorbox{red!6}{production} and \colorbox{red!6}{performance}.\colorbox{red!7}{"} instead, the red lights on the \colorbox{green!100}{stage} played up the" ominous" tone \colorbox{green!76}{of} the song as it gradually increased its tempo to the point whereby \colorbox{red!8}{the} end of the song was on the verge of sounding like an \colorbox{green!60}{inca}ntation\colorbox{red!20}{.} for \colorbox{red!8}{the} \colorbox{green!87}{diamonds} world \colorbox{red!10}{tour}\colorbox{red!12}{,} \colorbox{red!56}{rihanna} \colorbox{red!100}{performed}\colorbox{red!100}{\textcolor{blue}{"}} \colorbox{red!19}{man} \colorbox{red!30}{down}\colorbox{red!72}{"} \colorbox{red!15}{in} \colorbox{red!14}{a} \colorbox{green!78}{caribbean}@-@ theme \colorbox{red!17}{section} of the \colorbox{red!8}{show}, \colorbox{red!8}{which} also included" you da one"," no \colorbox{red!8}{love} allowed"," what' s my name?\colorbox{red!9}{"} \colorbox{red!7}{and}\colorbox{green!64}{"} rude boy"\colorbox{red!5}{.} james lachno of the telegraph highlight the caribbean@-@ themed\\\\
 \colorbox{red!22}{edge} of several realities: the \colorbox{green!70}{film}, the dream it inspires, the waking world it illuminates". the \colorbox{green!99}{music} in" i just can' t stop loving you", a duet \colorbox{green!100}{with} si\colorbox{green!60}{eda}h garrett, consisted mainly of finger snaps and timpani." \colorbox{red!10}{just} good \colorbox{red!8}{friends}", a duet \colorbox{red!10}{with} stevie wonder, was viewed by \colorbox{red!12}{critics} as sounding good at the beginning of \colorbox{green!85}{the} song, ending with a" chin\colorbox{red!11}{@}-@ bobbing cheerfulness\colorbox{green!79}{"}.\colorbox{green!75}{"} \colorbox{red!14}{the} way you \colorbox{red!13}{make} me feel"' \colorbox{red!10}{s} music consisted of \colorbox{red!15}{blues} harmonies. \colorbox{red!38}{the} \colorbox{red!52}{lyrics} \colorbox{red!100}{of}\colorbox{red!100}{\textcolor{blue}{"}} \colorbox{red!9}{another} \colorbox{red!21}{part} \colorbox{red!12}{of} \colorbox{red!40}{me}\colorbox{red!64}{"} \colorbox{red!33}{deal} \colorbox{red!10}{with} being united\colorbox{red!12}{,} as" \colorbox{red!10}{we}\\\\
 \colorbox{red!11}{not} manufactured\colorbox{green!50}{.} \colorbox{red!9}{no} one paid these kids.\colorbox{red!10}{"}=== \colorbox{red!9}{live} \colorbox{red!10}{performances}=== \colorbox{red!25}{one} direction \colorbox{red!100}{performed}\colorbox{red!100}{\textcolor{blue}{"}} \colorbox{red!43}{what} \colorbox{red!21}{makes} \colorbox{red!51}{you} \colorbox{red!12}{beautiful}\colorbox{red!83}{"} \colorbox{red!43}{on} red or black? \colorbox{red!14}{on} 10 september \colorbox{red!16}{2011}\colorbox{red!10}{.} \colorbox{red!9}{the} \colorbox{green!98}{performance} started with hosts ant\colorbox{red!8}{\&} dec announcing that the band \colorbox{red!10}{was} supposedly running late \colorbox{red!9}{for} their appearance, and cut to a video of one \colorbox{green!100}{direction} \colorbox{red!10}{boarding} a london tube carriage full of fans, as \colorbox{red!11}{the} studio version of the song began playing\colorbox{red!10}{.} each fan on the tube was given a numbered ticket. \colorbox{red!7}{the} band and fans disembarked the tube and made their way to the television studio, where \colorbox{red!8}{the} remainder of the song was sung \colorbox{green!52}{live}. after the song,\\\\
\newline \newline 
\noindent {\bf This is the end of visualization of high-level transformer factor. Click [\ref{sec:hyper}] to go back. }
\end{document}